\documentclass{article}

\usepackage{microtype}
\usepackage{graphicx}
\usepackage{subcaption}
\usepackage{booktabs} % for professional tables
\usepackage{multirow}
\usepackage{float}
\usepackage{placeins}

\usepackage{makecell}% added by sanka for data-efficiency results

\usepackage{hyperref}

\usepackage[accepted]{icml2026}

\usepackage{amsmath}
\usepackage{amssymb}
\usepackage{mathtools}
\usepackage{amsthm}
\usepackage{enumitem}

\usepackage[capitalize,noabbrev]{cleveref}

\theoremstyle{plain}

\theoremstyle{definition}

\theoremstyle{remark}

\usepackage[textsize=tiny]{todonotes}

\icmltitlerunning{Confusion-Aware Transfer Teacher Curriculum Learning Framework: Disentangling Scoring and Pacing Effects }

\begin{document}

\twocolumn[
  \icmltitle{Confusion-Aware Transfer Teacher Curriculum Learning Framework: Disentangling Scoring and Pacing Effects } %: A Framework for Disentangling Scoring and Pacing Effects% 



  \icmlsetsymbol{equal}{*}

  \begin{icmlauthorlist}
    \icmlauthor{Savini Kommalage}{yyy}
    \icmlauthor{Sanka Mohottala}{comp}
    \icmlauthor{Asiri Gawesha}{sch}
    \icmlauthor{Dulara Madhusanka}{yyy}
    \icmlauthor{Menan Velayuthan}{zzz}
    \icmlauthor{Dharshana Kasthurirathna}{yyy}
    \icmlauthor{Mahima Milinda Alwis Weerasinghe}{yyy}
    \icmlauthor{Charith Abhayaratne}{xxx}
  \end{icmlauthorlist}
% need to double check the spelling of names and all

  \icmlaffiliation{yyy}{Faculty of Computing, Sri Lanka Institute of Information Technology, Sri Lanka}
  \icmlaffiliation{comp}{Faculty of Engineering, University of Sri Jayewardenepura, Sri Lanka}
  \icmlaffiliation{sch}{Faculty of Engineering, Sri Lanka Institute of Information Technology, Sri Lanka}
  \icmlaffiliation{xxx}{University of Sheffield, United Kingdom}
  \icmlaffiliation{zzz}{Utrecht University, The Netherlands}

  \icmlcorrespondingauthor{Savini Kommalage}{it24100641@my.sliit.lk}
  \icmlcorrespondingauthor{Sanka Mohottala}{sanka@sjp.ac.lk}

  \icmlkeywords{Machine Learning, ICML}

  \vskip 0.3in
]



\printAffiliationsAndNotice{}  % no special notice (required even if empty)

\begin{abstract}

Curriculum learning couples two design choices, how samples are scored by difficulty and how harder samples are paced into training, making it difficult to attribute observed gains to either component. We disentangle these factors with two evaluation protocols: stage-wise test subsets that validate scoring functions independently of curriculum training, and a baseline that applies the same pacing schedule to randomly ordered data. Within the Transfer Teacher framework (TTF), we use these protocols to evaluate a confusion-aware difficulty score that considers both correct-class confidence and the probability distribution over incorrect classes. On CIFAR-10 with ResNet-18 and VGG-16, the proposed score produces model-interpretable difficulty rankings that align with human intuition. However, at full data, neither curriculum nor anti-curriculum ordering improves accuracy over standard training, indicating that improving the scoring function alone is insufficient to overcome the known failure modes of curriculum learning in TTF. In contrast, We find that confusion-aware curriculum ordering result in consistent data-efficiency benefits, outperforming random ordering by up to 8.7\% points at the 20\% data regime, suggesting the potential of TTF as a data-efficient training method.

\end{abstract}

\section{Introduction}
%introduce CL in a simple way with references
The order in which a model encounters training data shapes the trajectory of optimization. Curriculum learning (CL), introduced in ~\cite{bengio2009curriculum}, formalizes the intuition that models benefit from training on easier examples before progressively incorporating harder ones
. Empirically, CL has been shown to improve convergence and generalization 
across areas such as vision, language and graphs~\cite{wang2021survey,soviany2022curriculum,li2023curriculum}.

%introduces pacing and scoring functions 
In practice, designing a curriculum reduces to two coupled decisions: (i) how to assign a difficulty score to each training sample, and (ii) how to pace the introduction of harder samples ~\cite{bengio2009curriculum}. These components are typically evaluated jointly through final test accuracy, making it difficult to attribute improvements to either scoring or pacing. As a result, curriculum design remains largely empirical.

%need to introduce Transfer teacher direction and standard scoring functions and limitation - lack of confusion information-add references
In the Transfer Teacher 
framework (TTF)~\cite{weinshall2018curriculum}, a 
pre-trained teacher network ranks each sample by its prediction confidence in 
the correct class. Different CL methods in TTF shown improved accuracy~\cite{hacohen2019power, xu2020curriculum} but recent work~\cite{wucurricula} show that TTF only result in marginal accuracy improvements. 

While intuitive, this measure ignores how probability mass is distributed across incorrect classes. Two samples with identical true-class probability may differ substantially: one may reflect random uncertainty, while another reflects a confident but incorrect belief in a specific rival class yet 
receive the same difficulty score. We address this limitation with a \textit{confusion-aware difficulty score} within the TTF.

%connected to previous para - this gives our solution - better to make it a one para and shorter
% We address this limitation with a \textit{confusion-aware difficulty score} within the Transfer Teacher framework. The score
% incorporates the structure of the teacher's residual error distribution,
% measuring how probability mass concentrates among incorrect classes. This
% distinguishes between diffuse uncertainty and class-specific confusion in a
% way that naive confidence does not capture.

%need to show lack of work in scoring function verification AND work that look into how pacing and scoring contribute to convergence and generalization - need references [we can give references for standard verification method - final accuracy and convergence time, CL and anti-CL based verification [no pacing effect cl and  anti-cl measure the effect of scoring function directionality - our permuted accuracy results also relates to this directionality]] 
%then say that to fill this gap we introduce test dataset biniing approach (L1->L5) with human verifiability and 4 experiment evaluation method.

We further introduce two evaluation tools to disentangle curriculum components. First, a stage-wise test subset approach validates the scoring functions independent of curriculum training by verifying that test accuracy decreases monotonically across difficulty bins. Second, a pacing-isolated baseline applies identical pacing to randomly ordered data, isolating the contribution of the scoring function. We incorporate these into a comprehensive evaluation framework.

% pacing functions when continuous difficult to study duty change of data in each epoch - so stairecase function is used - give references like the bengio that uses staircase/step functions to defende that experimental work uses descrete pacing functions AND to study the effect of data-efficiency [results will be negative - ask savini how to get the results - i only need to identify the correct log files then can prepare the table]
Beyond accuracy, TTF has shown improved resilience to label noise and dataset imbalance~\cite{wucurricula, zhou2024curbench}. However, to the best of our knowledge, no prior work has examined whether TTF can serve as a data-efficient training method, reaching comparable accuracy with fewer training data. We find that TTF shows promise along this axis even where it fails to deliver generalization gains. We release code and detailed notes for reproducibility via, \href{https://github.com/BrAINLabs-Inc/confusion-aware-transfer-teacher}{GitHub repository}.
% \footnote{\href{https://github.com/BrAINLabs-Inc/confusion-aware-transfer-teacher}{GitHub link}}

% \footnote{Anonymized GitHub link: \url{https://anonymous.4open.science/r/confusion-aware-transfer-teacher-42E9/}}

% inform th reader that ths only work for classification

Key contributions of this paper are:
\begin{itemize}[itemsep=0pt, parsep=0pt, topsep=0pt]
    % \item We propose a \textbf{confusion-aware difficulty scoring 
    % function} that captures structured inter-class competition beyond 
    % confidence-based measures within the Transfer Teacher framework.
    
    % \item We introduce a novel confusion-aware scoring function and use it to demonstrate that improving the scoring function alone is not sufficient to improve generalization in the Transfer Teacher framework.
    \item We introduce a novel confusion-aware scoring function and use it to demonstrate that improving the scoring function alone is not sufficient to overcome the failure modes of Transfer Teacher framework (TTF).
    \item We propose a generalizable method to validate how well a scoring function captures sample difficulty from the model's perspective, and show that the introduced novel scoring function captures difficulty well and aligns with human intuition.
    \item We design an evaluation approach that studies the effect of scoring and pacing separately, and run extensive experiments on CIFAR-10 with ResNet-18 and VGG-16 that expose the limits of scoring in the TTF.
    % \item Despite these generalization limits, we show that the TTF can function as a data-efficient training method.
    \item We demonstrate that TTF can function as a data-efficient training method despite its generalization limits.
    % \item We release code and detailed notes for reproducibility.\footnote{Anonymized GitHub link: \url{https://anonymous.4open.science/r/confusion-aware-transfer-teacher-42E9/}}
    % \item We introduce a generalizable method to validate how well the scoring function capture the difficulty of a data point from model perspective and show that the novel scoring function capture the difficulty well and that it also correlate with human explainability.
    % \item We design an evaluation approach that study the effect of scoring and pacing functions separately and conduct extensive experiments with CIFAR-10 dataset with ResNet-18 and VGG-16 and show the limitation of scoring function in transfer teacher domain
    % \item despite the generalization limitations, we show that transfer teacher domain is a data-efficient training method 
    % \item Experiments on CIFAR-10 with ResNet-18 and VGG-16  with a novel scoring function.
    % \item Codes and detailed notes are release for reproducibility.
\end{itemize}

\begin{figure*}[!th]
  \centering
  \includegraphics[width=\textwidth]{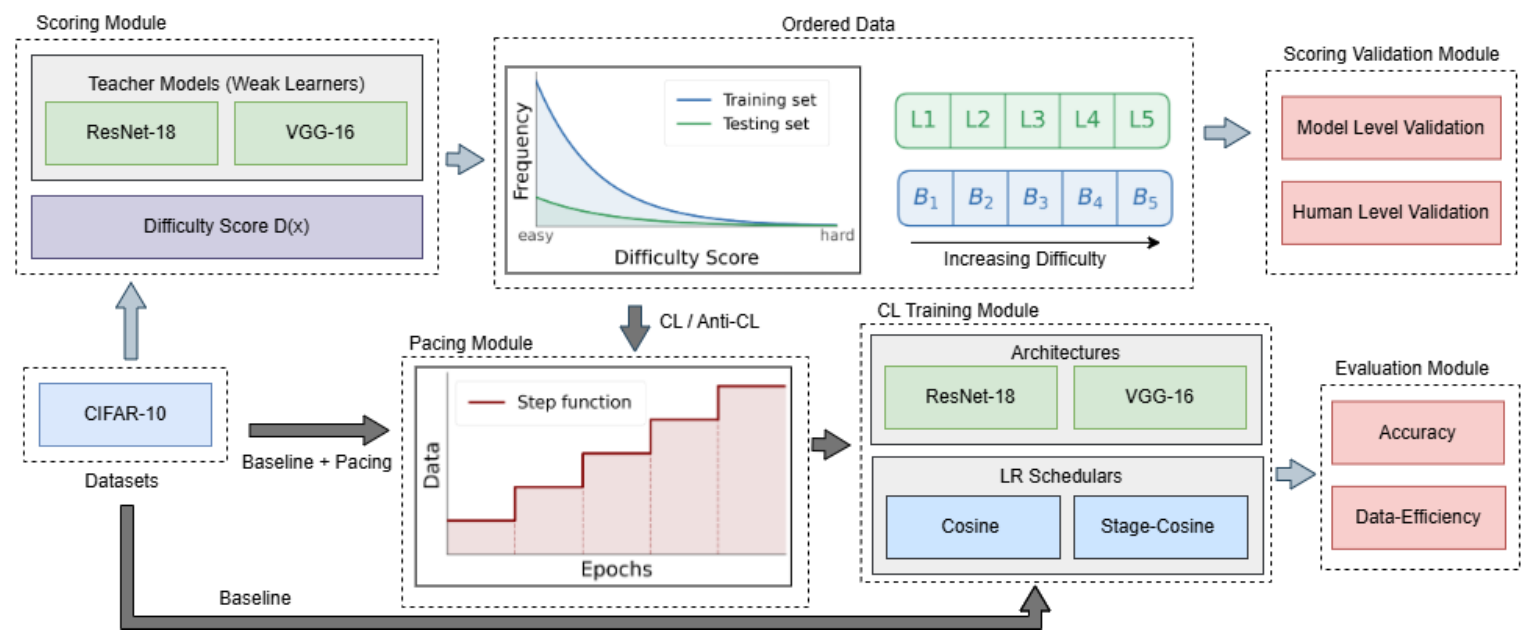}
  \caption{Overview of the confusion-aware transfer teacher curriculum learning framework}
  % \caption{Overview of the proposed curriculum learning pipeline. The teacher model evaluates the dataset and produces softmax outputs, which are passed to the scoring function $D(x)$ to rank samples by difficulty. The ranked dataset is partitioned into easy, medium, and difficult stages, which are introduced progressively via the pacing function to train the student model.}
  \label{fig:pipeline}
\end{figure*}
\section{Methodology}

\subsection{Problem Setup}

Let $\mathcal{D} = \{(x_i, y_i)\}_{i=1}^{N}$ with $y_i \in 
\{1, \dots, C\}$. A teacher model $f_\theta$ produces a softmax 
probability vector $\mathbf{p}(x_i) \in [0,1]^{C}$, where 
$\sum_{k=1}^{C} p_k(x_i) = 1$. We denote $p_{\text{true}} = p_{y_i}$ 
as the probability assigned to the ground-truth class, and 
$m(x_i) = 1 - p_{y_i}(x_i)$ as the total residual probability 
distributed across all incorrect classes. 
% (Figure~\ref{fig:pi}) - perhaps we can do some statistical anaylsis based visualizations on these values

% \subsection{Limitations of Confidence-Based Scoring}

The standard difficulty score in TTF $D_{\text{naive}}(x) = 1 - p_{\text{true}}$ 
uses a single scalar from the output distribution, discarding all 
information about how residual probability mass is distributed across 
incorrect classes. 

An important consequence is that two samples with identical $p_{\text{true}}$ receive the same score regardless of whether the model's errors are randomly spread across $C-1$ classes or concentrated to a specific class (ex: "cat" class may confuse with "dog" class), resulting in qualitatively different failure modes.

% % This has two consequences. First, two samples with 
% % identical $p_{\text{true}}$ receive the same score regardless of whether 
% % the model's errors are randomly spread across $C-1$ classes 
% % or concentrated to a specific rival class, resulting in qualitatively 
% % different failure modes. 

% % Second, because teacher confidence varies 
% % systematically by class, ranking by $p_{\text{true}}$ induces class 
% % imbalance across difficulty stages: visually distinctive classes 
% % dominate easy bins while semantically ambiguous classes saturate hard 
% % bins. This conflates class-level 
% % discriminability with sample-level difficulty.
% (See Appendix~\ref{app:class_distributions}). 

\subsection{Confusion-Aware Difficulty Score}
% Let $m = 1 - p_{\text{true}}$. We define a conditional distribution 
% over incorrect classes:
% %
% \begin{equation}
%     \tilde{p}_k = \frac{p_k}{m}, \quad k \neq y_i, 
%     \qquad \sum_{k \neq y} \tilde{p}_k = 1
% \end{equation}
% %
% This is the distribution over wrong classes \emph{given} that the 
% model errs. We define the \textbf{confusion variance} as the 
% population variance of this distribution:
% %
% \begin{equation}
%     \text{ConfVar}(x) = \frac{1}{C-1} \sum_{k \neq y} 
%     \left(\tilde{p}_k - \frac{1}{C-1}\right)^2
% \end{equation}
% %
% The final difficulty score is:
% %
% \begin{equation}
%     \boxed{D(x) = (1 - p_{\text{true}}) \cdot \text{ConfVar}(x)}
% \end{equation}
% %
Let $m(x_i) = 1 - p_{y_i}(x_i)$, where $p_{y_i}(x_i)$ is the model's
predicted probability for the true class $y_i$ given input $x_i$.
We define a conditional distribution over incorrect classes:
\begin{equation}
    \tilde{p}_k(x_i) = \frac{p_k(x_i)}{m(x_i)}, \quad k \neq y_i,
    \qquad \sum_{k \neq y_i} \tilde{p}_k(x_i) = 1
\end{equation}
This is the distribution over wrong classes \emph{given} that the
model errs on $x_i$. We define the \textbf{confusion variance} as the
population variance of this distribution:
\begin{equation}
    \mathrm{ConfVar}(x_i) = \frac{1}{C-1} \sum_{k \neq y_i}
    \left(\tilde{p}_k(x_i) - \frac{1}{C-1}\right)^2
\end{equation}
The final difficulty score is:
\begin{equation}
    D(x_i) = \bigl(1 - p_{y_i}(x_i)\bigr) \cdot \mathrm{ConfVar}(x_i)
\end{equation}

The two factors are interpretable: $(1-p_{\text{true}})$ captures 
\emph{how wrong} the model is; $\text{ConfVar}(x)$ captures 
\emph{how structured} that wrongness is. Difficulty score is high 
only when the model is both uncertain about the correct class 
\emph{and} confused in a class-specific way. If the confusion is spread across classes, then even if model is uncertain about the class, it may give a low difficulty score.

\subsection{Teacher Model}

We train ResNet-18 and VGG-16 teacher models on 100\% of the 
CIFAR-10 training set (LR 0.1, cosine annealing, batch size 128, 
momentum 0.9, weight decay $5\!\times\!10^{-4}$, dropout 0.2, 
no augmentation). To overcome the overconfidence in teacher model and to retain the class-wise confusion information in $\mathbf{p}(x_i)$, a weak learner is obtained by limiting the training epochs to 30.  Teacher training 
details and ablation results (10\%, 20\%, 100\% 
data) are provided in Appendix~\ref{app:teacher_ablations}.

\subsection{Transfer Teacher Training and Evaluation}
\label{transfer_teacher_evluation}
All experiments are conducted with CIFAR-10 dataset while ResNet-18 and VGG-16 are used as the architectures in experiments. To validate the scoring function measurement of difficulty, we partition the test set into five difficulty bins (L1--L5) using teacher scores and evaluate a student model 
trained \emph{without} any curriculum on each bin separately. A 
valid scoring function must produce a strictly monotone accuracy 
gradient from L1 (easiest) to L5 (hardest). 

As the pacing function, we use a step function based scheduler~\cite{bengio2009curriculum,wang2021survey}. After ranking the CIFAR-10 dataset according to the $D(x)$, it was partitioned into five equal stages of 10000 samples ($B_i$ where $i\in \{1, \dots, 5\}$). Figure~\ref{fig:pipeline} show an overview of this TTF pipeline. Training begins on the easiest stage ($B_1$) and adds one stage every 
20 epochs over 100 total epochs. We compare two learning rate 
schedules: Cosine Annealing~\cite{cazenave2021cosine}, a single global schedule over 
100 epochs and Stage-Wise Cosine Annealing which restarts at 
each stage boundary, allowing the model to re-adapt as harder samples 
are introduced (See Appendix~\ref{fig:lr_schedules}).

% \subsection{Pacing-Isolated Baseline}

To isolate the contribution of the scoring function, we introduce an evaluation protocol where we use the step function scheduler with randomly ordered samples (referred as "Baseline + Pacing" in Table~\ref{tab:results}). Performance difference between this and Curriculum and Anti-Curriculum evaluation protocols can be attributable primarily to the scoring function.

\subsection{Data-Efficiency in TTF}
% Since the Baseline+Pacing, Curriculum (CL) and Anti-Curriculum (Anti-CL) protocols introduce data in the same manner, we can compare the model performance at the end of each stage and they corrospond to a subset of the training dataset. For instance, at the end of $B_1$ (20 epochs), performance across these 3 protocols corrospond to 20\% of the training dataset in which Baseline+Pacing uses random 20\% while CL uses 20\% easiest data and at the end of $B_2$ (40 epochs), they have used 40\% data in the same way so on and so on. Thus this performance evaluation can be used to study if the CL/Anti-CL method has the potential to work as data-efficient training methods. 
Because Baseline+Pacing, Curriculum (CL), and Anti-Curriculum (Anti-CL) share the same pacing schedule, their stage-end checkpoints correspond to matched data budgets. After $B_1$ (20 epochs), each has seen 20\% of the data, random for Baseline+Pacing, easiest for CL, and hardest for Anti-CL, the fractions grow identically through $B_5$. Comparing protocols at each stage isolates the effect of ordering from the effect of data volume, indicating whether CL or Anti-CL offers data-efficiency gains.

\begin{figure}[ht]
    \centering
    \includegraphics[width=0.99\linewidth, height=6cm, keepaspectratio]{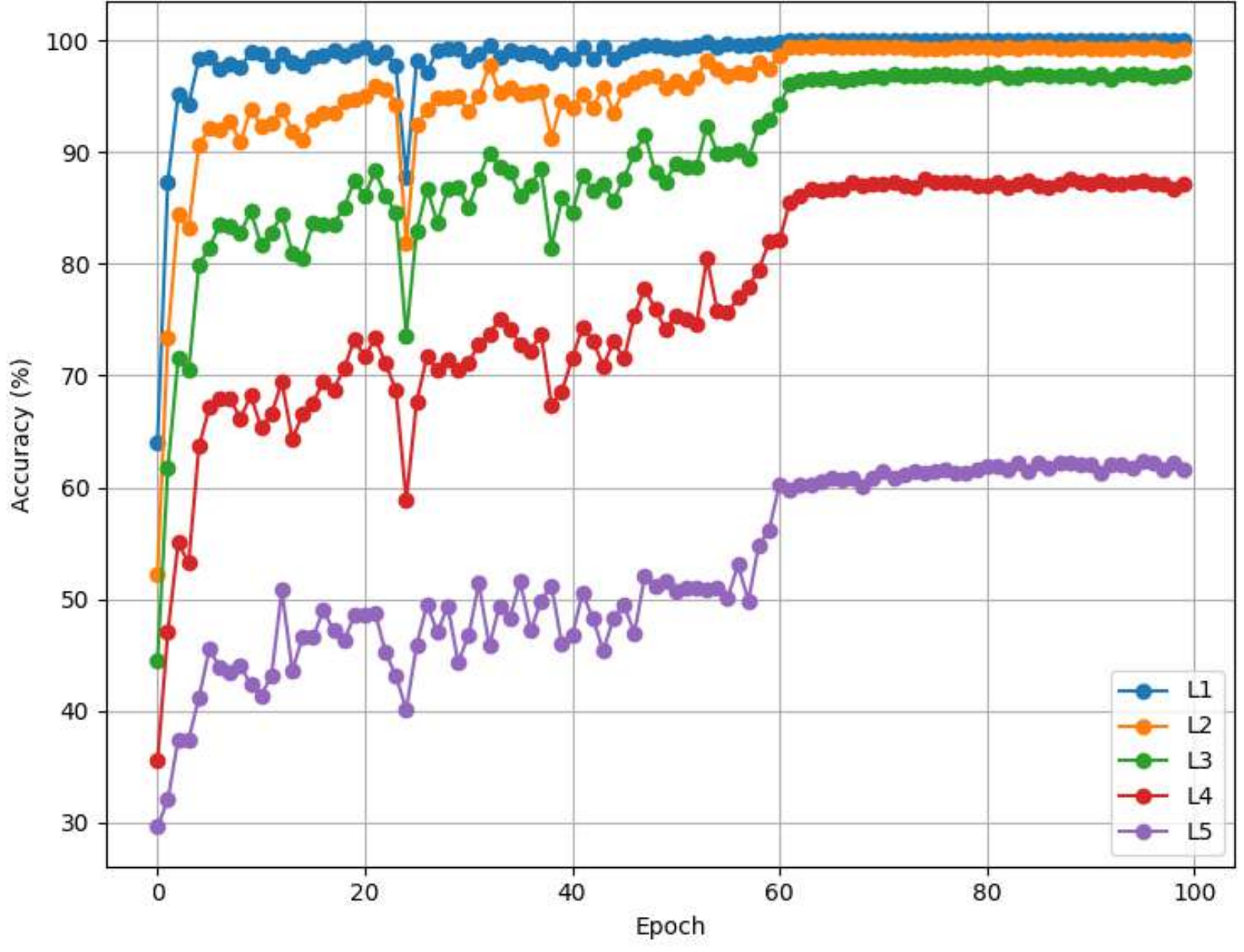}
    \caption{Baseline (Non-CL) performance with only the test data divided into fixed difficulty bins L1--L5.}
    \label{fig:stage_100}
\end{figure}
% \begin{table*}
% \caption{Comparison of training strategies trained on 100\% of the CIFAR-10 dataset with ResNet-18, evaluated across test subsets $L_1$–$L_5$ of increasing difficulty. Aggregate test accuracy is reported in the final column.}
% \label{tab:stgnn-comparison}
% \centering
% \begin{tabular}{lcccccc}
% \toprule
% \multirow{2}{*}{\makecell{Training \\ Strategy}} & \multicolumn{5}{c}{Test Subset} & \multicolumn{1}{c}{\multirow{2}{*}{\makecell{Aggregate \\ Test Accuracy}}} \\
% \cmidrule(lr){2-6}
%  & $L_{1}$ & $L_{2}$ & $L_{3}$ & $L_{4}$ & $L_{5}$ & \\
% \midrule
% Baseline         & 100.00 & 99.25 & 97.10 & 87.20 & 61.55 & 89.02 \\
% Pacing baseline  & 99.90  & 98.50 & 92.55 & 77.65 & 52.90 & 84.30 \\
% Curriculum       & 100.00 & 99.50 & 96.75 & 83.65 & 50.65 & 86.11 \\
% Anti-curriculum  & 99.15  & 94.60 & 86.60 & 70.60 & 48.30 & 79.85 \\
% \bottomrule
% \end{tabular}
% \end{table*}

\section{Results and Discussion}

\subsection{Validation of Scoring Function}

\begin{figure}[h]
    \centering
    \includegraphics[width=\columnwidth]{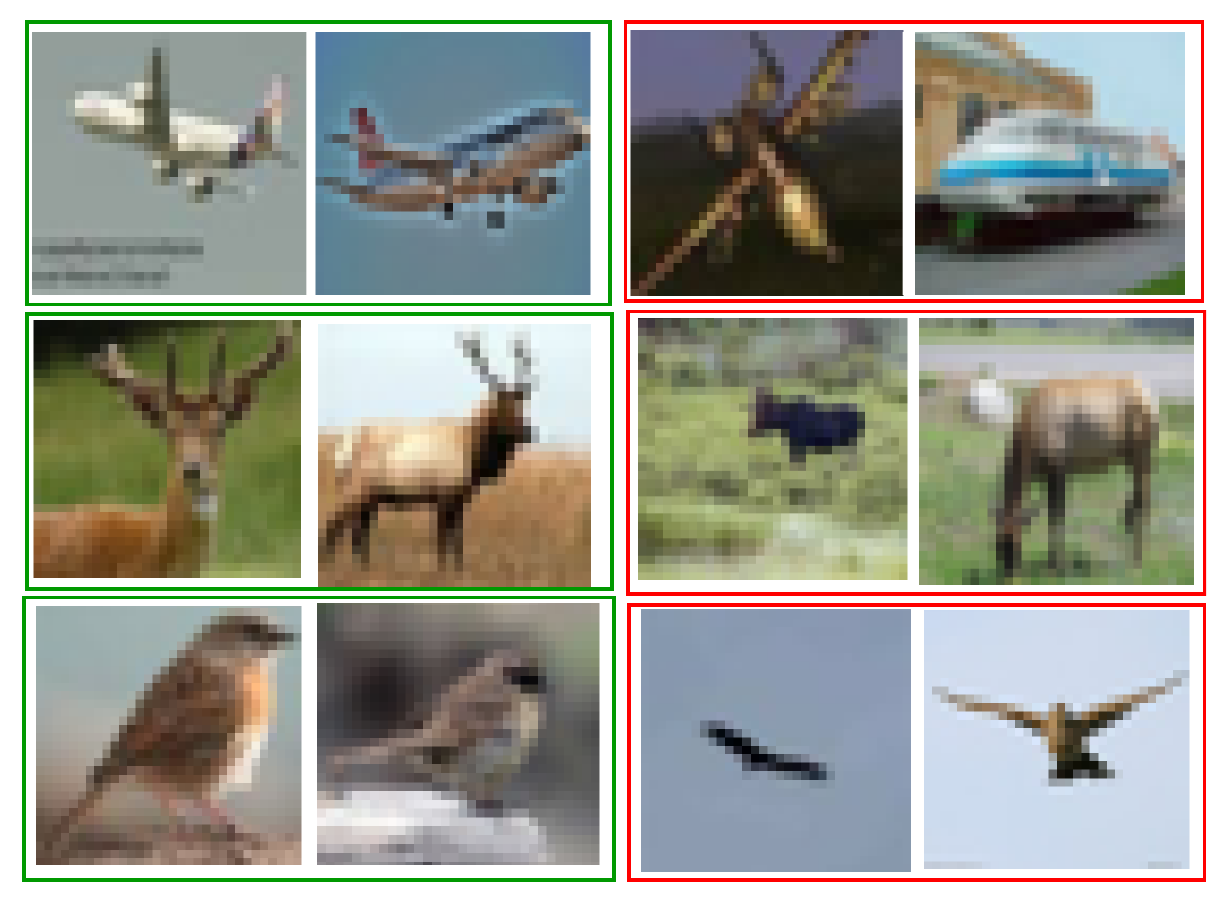}
    \caption{Qualitative comparison of easy (left, green) vs. difficult 
    (right, red) samples across classes \textit{airplane}, \textit{deer}, 
    and \textit{bird}.}
    % \caption{Qualitative comparison of easy (left, green) vs. difficult 
    % (right, red) samples across classes \textit{airplane}, \textit{deer}, 
    % and \textit{bird}. Easy samples show near-zero 
    % difficulty, while hard samples exhibit semantic confusion 
    % (high ConfVar $\approx$ 0.097).}
    \label{fig:qualitative_grid}
\end{figure}

Figure~\ref{fig:stage_100} reports stage-wise test accuracy 
across all 100 training epochs. The 100\% teacher produces a clear 
monotone gradient that holds throughout training, validating the scoring function independently of curriculum training and confirming that $D(x)$ is a good measure of sample difficulty from the model's perspective.
% teacher expressiveness is a necessary condition for valid difficulty estimation.
In contrast, a teacher trained on only 10\% of the data produces a near-flat gradient, showing that a reliable difficulty signal needs a teacher trained on the full dataset. Figure~\ref{fig:qualitative_grid} supports this with high $D(x)$ samples showing clear semantic confusion, where hard samples exhibit structured semantic confusion like deer mistaken for bird, airplane for automobile which matches real difficulty rather than random noise. Detailed results on difficulty score analysis appear in Appendix~\ref{app:scoring}.

%-------original------
% Figure~\ref{fig:stage_100} reports stage-wise test accuracy 
% across all 100 training epochs. The 100\% teacher produces a clear 
% monotone gradient that holds throughout training ( 
% L1 $64.00\%$ $\to$ L2 $52.25\%$ $\to$ L3 $44.45\%$ $\to$ L4 
% $35.50\%$ $\to$ L5 $29.60\%$), validating the scoring function independently of curriculum training and confirming that $D(x)$ is a good measure of sample difficulty from the model's perspective.
% % teacher expressiveness is a necessary condition for valid difficulty estimation.
% In contrast, a teacher trained on only 10\% of the data produces a near-flat gradient ( L1 $44.75\%$ $\to$ 
% L2 $42.25\%$ $\to$ L3 $42.95\%$ $\to$ L4 $42.65\%$ $\to$ L5 
% $43.40\%$, range $42.25\%$--$44.75\%$), showing that a reliable difficulty signal needs a teacher trained on the full dataset. Figure~\ref{fig:qualitative_grid} supports this with high $D(x)$ samples showing clear semantic confusion, where hard samples exhibit structured semantic confusion like deer mistaken for bird, airplane for automobile which matches real difficulty rather than random noise. Detailed results on difficulty score analysis appear in Appendix~\ref{app:scoring}.

\begin{figure}[h]
    \centering
    \includegraphics[width=\columnwidth]{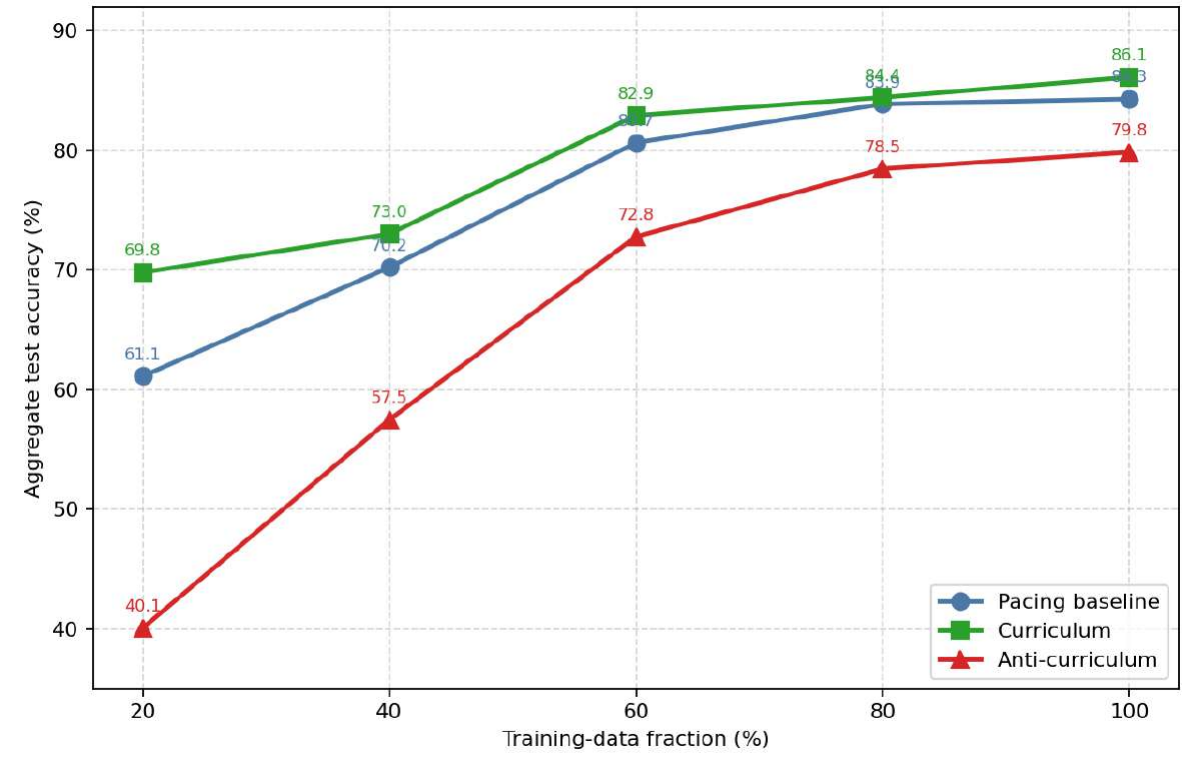}
    \caption{Data efficiency of curriculum-based training strategies (aggregate test accuracy $=$ full test set accuracy).}
    \label{fig:data_eff_1}
\end{figure}

\subsection{Performance Evaluation}

\begin{table}[t]
  \caption{Test accuracy (\%) on CIFAR-10 under four training protocols
  and two learning rate schedulers.}
  \label{tab:results}
  \centering
  \begin{tabular}{llcc}
    \toprule
    Model &  Protocol & Cosine & Stage-Cosine \\
    \midrule
    \multirow{4}{*}{ResNet-18}
      & Baseline          & 89.20 & 88.30 \\
      & Baseline + Pacing & 84.44 &  87.30 \\
      & Curriculum        & 86.13 & 87.53\\
      & Anti-Curriculum   & 79.94 & 86.43 \\
    \midrule
    \multirow{4}{*}{VGG-16}
      & Baseline          & 89.57 & 89.48 \\
      & Baseline + Pacing & 88.05 & 89.01 \\
      & Curriculum        & 87.86 & 88.75 \\
      & Anti-Curriculum   & 87.18 & 88.64 \\
    \bottomrule
  \end{tabular}
\end{table}

Table~\ref{tab:results} reports CIFAR-10 test accuracy under the introduced evaluation framework, which decouples scoring (Curriculum, Anti-Curriculum) from pacing (Baseline + Pacing) against the non-CL Baseline. 
% \paragraph{Finding 1: Improved scoring does not raise accuracy over the Baseline.}
With tuned hyperparameters, neither Curriculum nor Anti-Curriculum exceeds the Baseline on either backbone or learning rate scheduler (LRS). 
% The largest gap appears on ResNet-18 with cosine annealing (Baseline 89.20\% vs.\ Curriculum 86.13\%, $\Delta=3.07$); VGG-16 shows a smaller but consistent gap (89.57\% vs.\ 87.86\%, $\Delta=1.71$). 
Combined with the model-aligned difficulty score validation, this directly corroborate our claim that within the Transfer Teacher framework, an improved scoring function is not sufficient to improve accuracy over standard training.
% \paragraph{Finding 2: Pacing imposes a cost; scoring's recovery is architecture-dependent.}

Baseline + Pacing underperforms the Baseline in all four configurations (e.g., 84.44\% vs.\ 89.20\% on ResNet-18, cosine), consistent with reduced data exposure during early discrete stages. On ResNet-18, Curriculum recovers part of this gap under both LRS (86.13\% to 84.44\%, cosine; 87.53\% to 87.30\%, stage-cosine), attributing the improvement to confusion-aware ordering rather than pacing. On VGG-16 the recovery vanishes with curriculum trails Baseline + Pacing under both LRS (87.86\% vs.\ 88.05\%; 88.75\% vs.\ 89.01\%), and the overall spread is narrower (87.18--89.57\% on VGG-16 vs.\ 79.94--89.20\% on ResNet-18). We hypothesis that this is a result of robustness of VGG-16 to training-order perturbations. 
% The separated design surfaces this dependence — a scoring-only evaluation would have conflated it with pacing.
% \paragraph{Finding 3: Stage-cosine annealing helps every paced protocol but harms the Baseline.}

Stage-cosine improves accuracy over cosine in all six paced configurations. The largest gain is Anti-Curriculum on ResNet-18 (79.94\% to 86.43\%, +6.49pp), the smallest is Curriculum on VGG-16 (+0.89pp). In contrast, the non-CL Baseline degrades slightly under stage-cosine on both backbones (89.20\% to 88.30\%; 89.57\% to 89.48\%). Re-adapting the learning rate at the start of each stage  appears highly effective, indicating that curriculum-aware learning rate scheduling could be the potential solution to TTF generalization limitations.
Visualizations of these experiment results are provided in Appendix~\ref{app:AppendixB}.

\subsection{Data-Efficiency Evaluation}

Figures~\ref{fig:data_eff_1} and~\ref{fig:data_eff_2} report 
accuracy at progressive training subset sizes. Curriculum ordering 
consistently outperforms Baseline + Pacing across all data regimes, 
with the largest gap at 20\% data (69.81\% vs.\ 61.13\%). These results suggest that CL has the potential to be used as a data-efficient training method, especially when limited by computing resources. Detailed results are provided in 
Appendix~\ref{app:AppendixC}.
\section{Conclusion}

% We presented a confusion-aware difficulty scoring function and two evaluation 
% tools for disentangling the contributions of scoring and pacing in curriculum 
% learning. Our stage-accuracy gradient test provides a principled way to 
% validate difficulty rankings independently of curriculum training, while the 
% pacing-isolated baseline isolates the scoring function's effect on performance. 
% Experiments on CIFAR-10 with ResNet-18 and VGG-16 show that the scoring 
% function accounts for a measurable portion of curriculum gains, and that 
% curriculum ordering provides consistent data efficiency benefits --- 
% particularly when training on smaller ranked subsets, where curriculum 
% outperforms random staging by up to 8.7 percentage points at the 20\% data 
% regime. These results suggest that principled difficulty scoring matters most 
% when data is constrained, and that the broader evaluation framework introduced 
% here can help distinguish genuine curriculum effects from artifacts of staged 
% training schedules.

We introduced a confusion-aware difficulty score and two evaluation tools, a stage-wise accuracy evaluation and a pacing-isolated baseline that seperates scoring from pacing in CL. On CIFAR-10, score function produces an increasing difficulty that aligns with human intuition, but accuracy with CL/Anti-CL does not improve, indicating that within the TTF, an improved scoring function alone is not enough for accuracy gains. In data-efficiency experiment setups, confusion-aware ordering outperforms random staging by up to 8.7\% points at 20\% of dataset, positioning TTF as a data-efficient training strategy despite its suboptimal performance with full dataset. The consistent improvement from stage-cosine scheduling positions designing of curriculum-aware training pipeline as a future research direction to improve the accuracy of TTF over standard training.


% Authors are kindly asked to make their submissions as accessible as possible
% for everyone including people with disabilities and sensory or neurological
% differences. Tips of how to achieve this and what to pay attention to will be
% provided on the conference website \url{http://icml.cc/}.

% \section*{Software and Data}

% If a paper is accepted, we strongly encourage the publication of software and
% data with the camera-ready version of the paper whenever appropriate. This can
% be done by including a URL in the camera-ready copy. However, \textbf{do not}
% include URLs that reveal your institution or identity in your submission for
% review. Instead, provide an anonymous URL or upload the material as
% ``Supplementary Material'' into the OpenReview reviewing system. Note that
% reviewers are not required to look at this material when writing their review.

% % Acknowledgements should only appear in the accepted version.
% \section*{Acknowledgements}

% \textbf{Do not} include acknowledgements in the initial version of the paper

% If a paper is accepted, the final camera-ready version can (and usually should)
% include acknowledgements.  Such acknowledgements should be placed at the end of
% the section, in an unnumbered section that does not count towards the paper
% page limit. Typically, this will include thanks to reviewers who gave useful
% comments, to colleagues who contributed to the ideas, and to funding agencies
% and corporate sponsors that provided financial support.

\section*{Acknowledgements}

This research was supported by a research grant funded by the Sri Lanka Institute of Information Technology, Sri Lanka (Grant No.~PVC(R\&I)/RG/2025/12). The computational resources used in this work were provided through equipment funded by the Accelerating Higher Education Expansion and Development (AHEAD) Operation of the Ministry of Higher Education of Sri Lanka, funded by the World Bank (\url{https://ahead.lk/result-area-3/}).

\section*{Impact Statement}

This paper presents work whose goal is to advance the field of Machine
Learning. There are many potential societal consequences of our work, none
which we feel must be specifically highlighted here.

% \section*{Impact Statement}

% Authors are \textbf{required} to include a statement of the potential broader
% impact of their work, including its ethical aspects and future societal
% consequences. This statement should be in an unnumbered section at the end of
% the paper (co-located with Acknowledgements -- the two may appear in either
% order, but both must be before References), and does not count toward the paper
% page limit. In many cases, where the ethical impacts and expected societal
% implications are those that are well established when advancing the field of
% Machine Learning, substantial discussion is not required, and a simple
% statement such as the following will suffice:

% ``This paper presents work whose goal is to advance the field of Machine
% Learning. There are many potential societal consequences of our work, none
% which we feel must be specifically highlighted here.''

% The above statement can be used verbatim in such cases, but we encourage
% authors to think about whether there is content which does warrant further
% discussion, as this statement will be apparent if the paper is later flagged
% for ethics review.

% \newpage
\bibliography{example_paper}
\bibliographystyle{icml2026}

%%%%%%%%%%%%%%%%%%%%%%%%%%%%%%%%%%%%%%%%%%%%%%%%%%%%%%%%%%%%%%%%%%%%%%%%%%%%%%%
%%%%%%%%%%%%%%%%%%%%%%%%%%%%%%%%%%%%%%%%%%%%%%%%%%%%%%%%%%%%%%%%%%%%%%%%%%%%%%%
% APPENDIX
%%%%%%%%%%%%%%%%%%%%%%%%%%%%%%%%%%%%%%%%%%%%%%%%%%%%%%%%%%%%%%%%%%%%%%%%%%%%%%%
%%%%%%%%%%%%%%%%%%%%%%%%%%%%%%%%%%%%%%%%%%%%%%%%%%%%%%%%%%%%%%%%%%%%%%%%%%%%%%%
% \newpage
\clearpage
\appendix
 \onecolumn
\section{Scoring Function Analysis}
\renewcommand{\thefigure}{A\arabic{figure}}
\setcounter{figure}{0}
\renewcommand{\thetable}{A\arabic{table}}
\setcounter{table}{0}
\label{app:scoring}

We empirically justify the composite difficulty score
$D(x) = (1 - p_{\text{true}}) \cdot \mathrm{ConfVar}(x)$
by examining each component's distribution over the 50{,}000-sample CIFAR-10 training set.
%% ----- A.1 -----
\subsection{Teacher Confidence \texorpdfstring{$p_{\text{true}}$}{p\_true}}
\label{app:ptrue}

\begin{figure}[ht]
  \centering
  \begin{subfigure}[]{0.95\linewidth}
    \centering
    \includegraphics[width=0.75\linewidth]{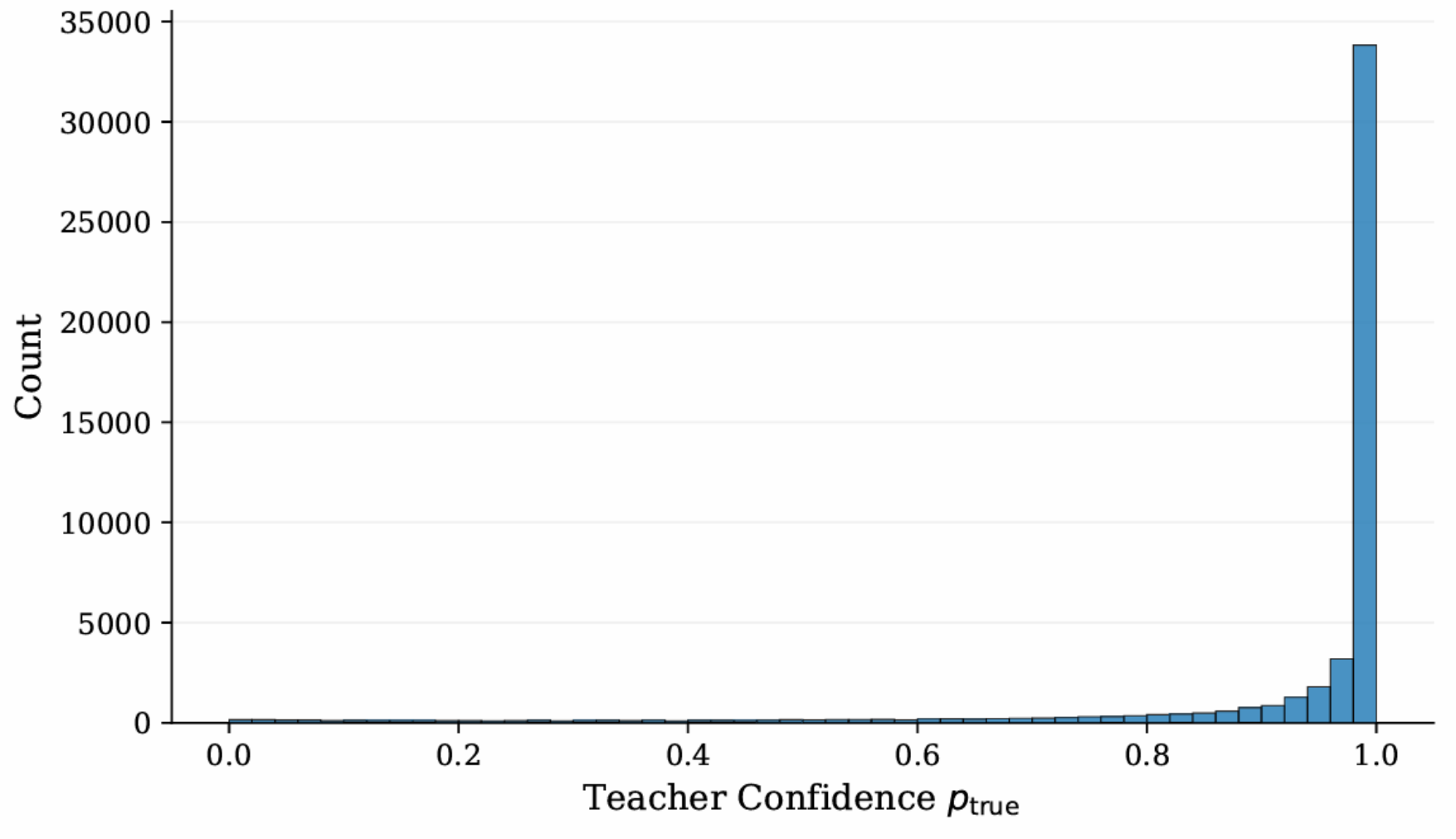}
    \caption{Full training set distribution.}
  \end{subfigure}

  \begin{subfigure}[]{0.95\linewidth}
    \centering
    \includegraphics[width=0.75\linewidth]{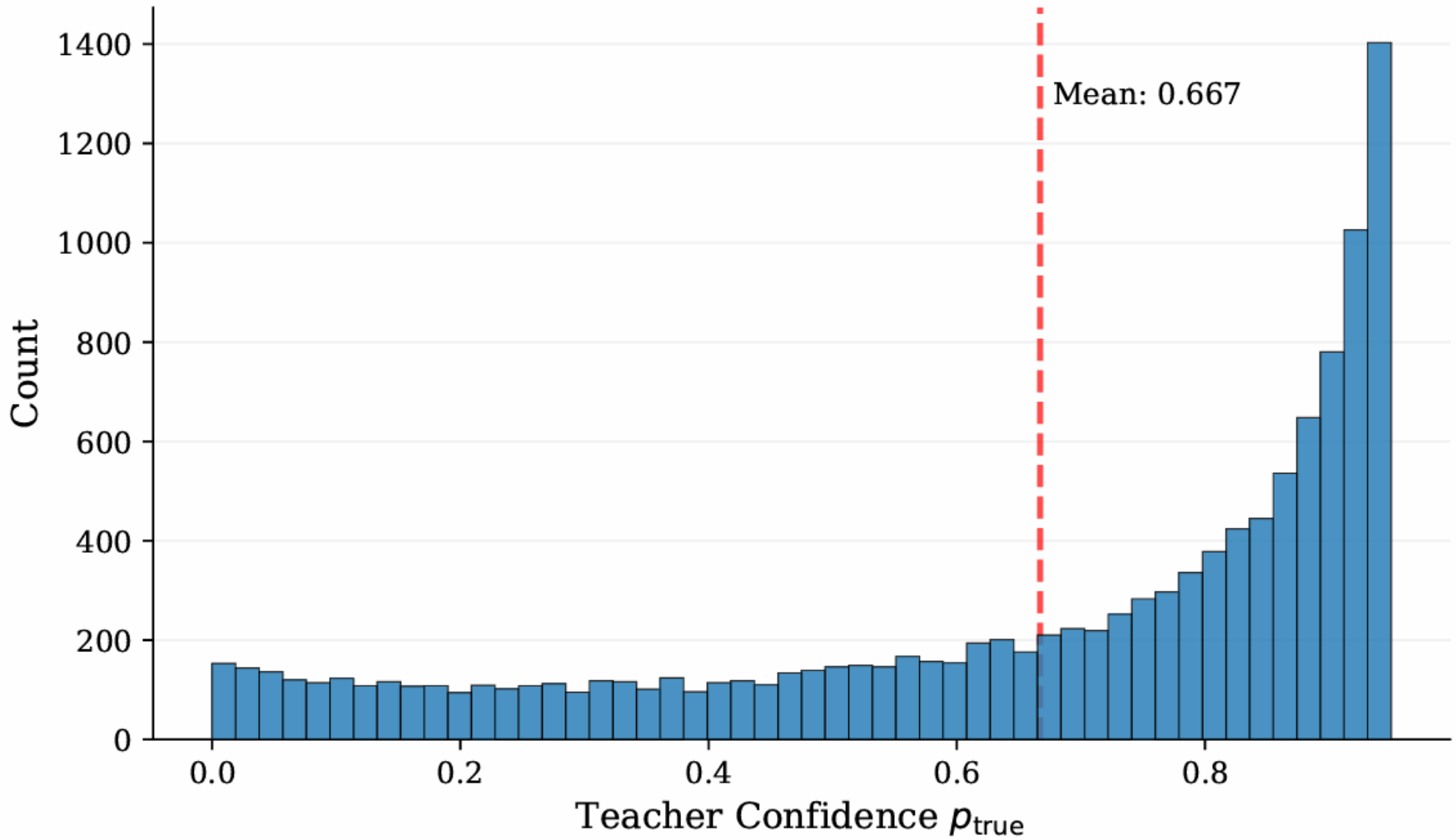}
    \caption{Uncertain subset ($p_{\text{true}} < 0.95$, $N = 11{,}969$ (23.9\% of training set)).}
  \end{subfigure}
  \caption{Distribution of teacher confidence $p_{\text{true}}$.
    \textbf{Top:} The spike at $p_{\text{true}} \approx 1$
    renders naive confidence a coarse difficulty signal.
    \textbf{Bottom:} The uncertain subset reveals a broadly
    spread distribution with mean $0.667$, confirming that genuine ambiguity is
    present but minority.}
  \label{fig:ptrue-split}
\end{figure}

Figure~\ref{fig:ptrue-split} shows the marginal distribution of $p_{\text{true}}$.
The full distribution (top) is dominated by a spike near $1.0$: the teacher
correctly classifies the vast majority of training samples with high confidence,
making $1 - p_{\text{true}}$ a poor discriminator within the easy majority.
The zoomed panel (bottom) restricts to the 11{,}969 uncertain samples
($p_{\text{true}} < 0.95$; 23.9\% of the training set) and reveals a broadly
spread distribution with mean $0.667$, confirming that genuine ambiguity is
present but minority.

\begin{figure}[!t]
  \centering
  \begin{subfigure}{\linewidth}
    \centering
    \includegraphics[width=0.95\linewidth]{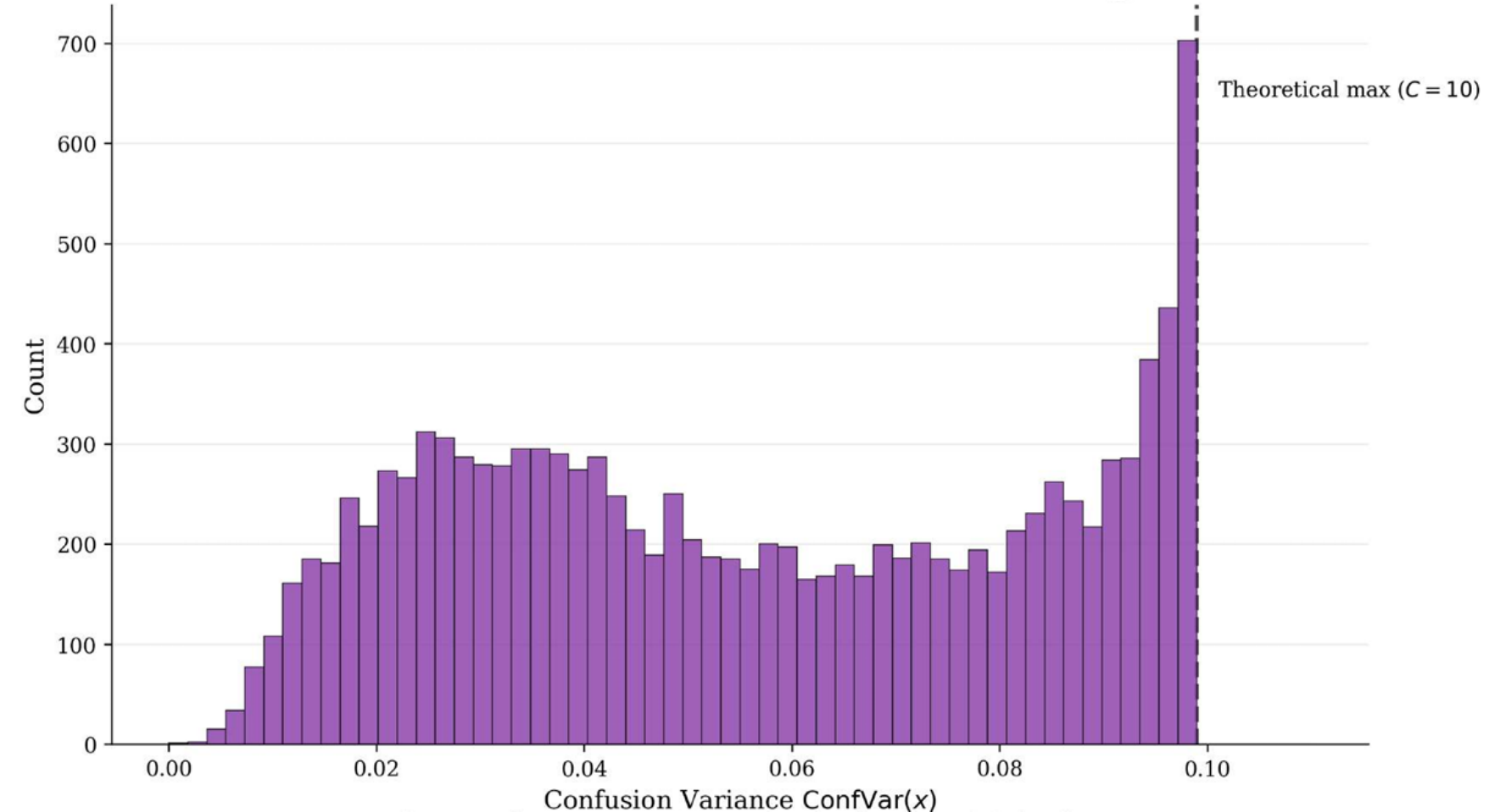}
    \caption{Distribution of $\mathrm{ConfVar}(x)$ over uncertain samples
      ($p_{\text{true}} < 0.95$).
      The broad body provides a well-spread discriminative signal;
      the spike near the theoretical maximum (${\approx}0.099$) captures
      samples with maximally structured, single-class confusion.}
    \label{fig:confvar}
  \end{subfigure}

  \begin{subfigure}{\linewidth}
    \centering
    \includegraphics[width=0.95\linewidth]{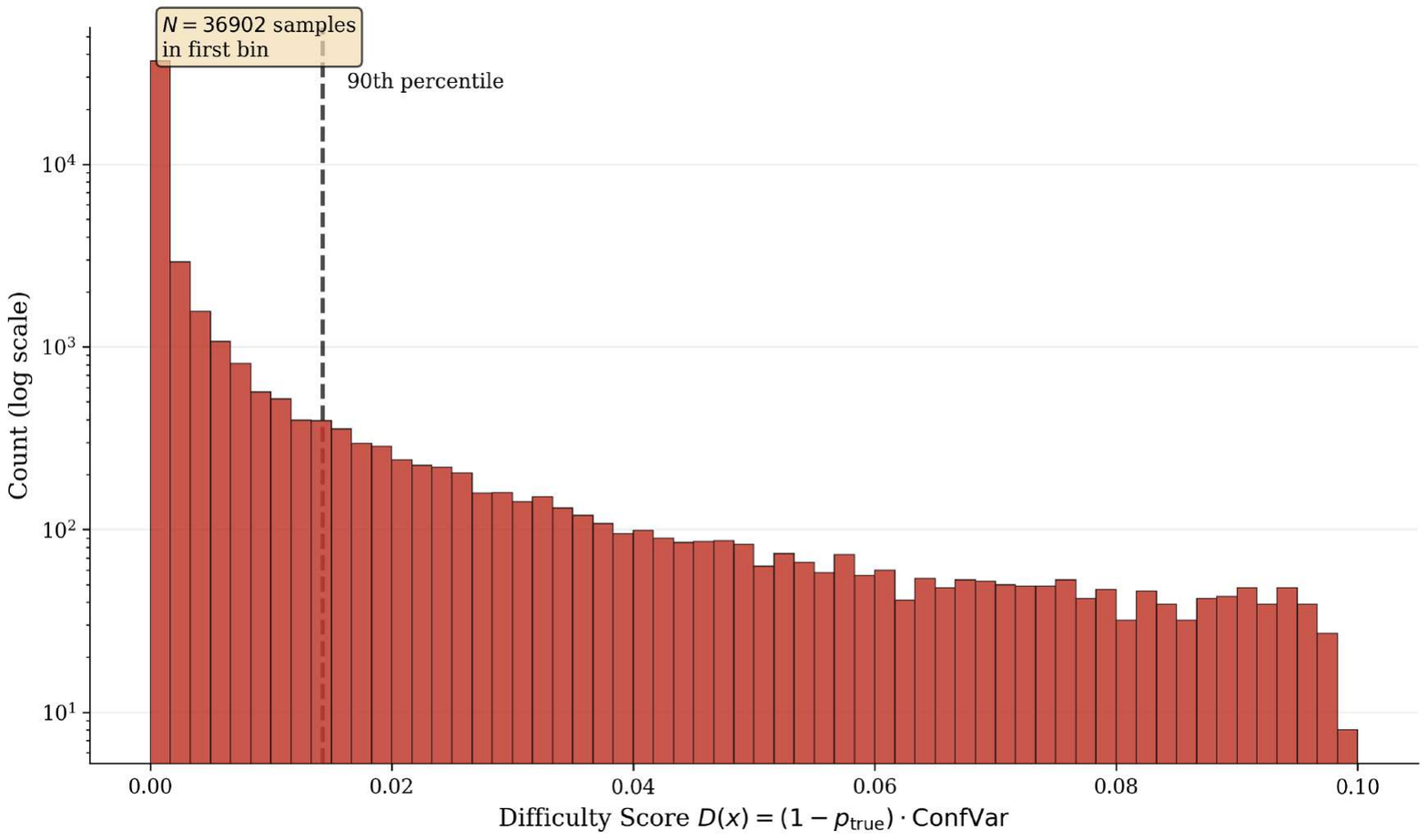}
    \caption{Distribution of $D(x) = (1-p_{\text{true}})\cdot\mathrm{ConfVar}$
      (log-count axis).
      Over 73\% of samples fall in the first bin; the 90th-percentile boundary
      (dashed) lies at $D(x) \approx 0.0015$.
      The smooth power-law tail confirms a coherent hard subset rather than
      isolated outliers.}
    \label{fig:difficulty-log}
  \end{subfigure}

  \caption{Analysis of uncertainty and difficulty metrics.}
  \label{fig:combined}
\end{figure}

%% ----- A.2 -----
\subsection{Confusion Variance \texorpdfstring{$\mathrm{ConfVar}$}{ConfVar}}
\label{app:confvar}

\begin{figure*}[!t]
  \centering
  \includegraphics[width=\textwidth]{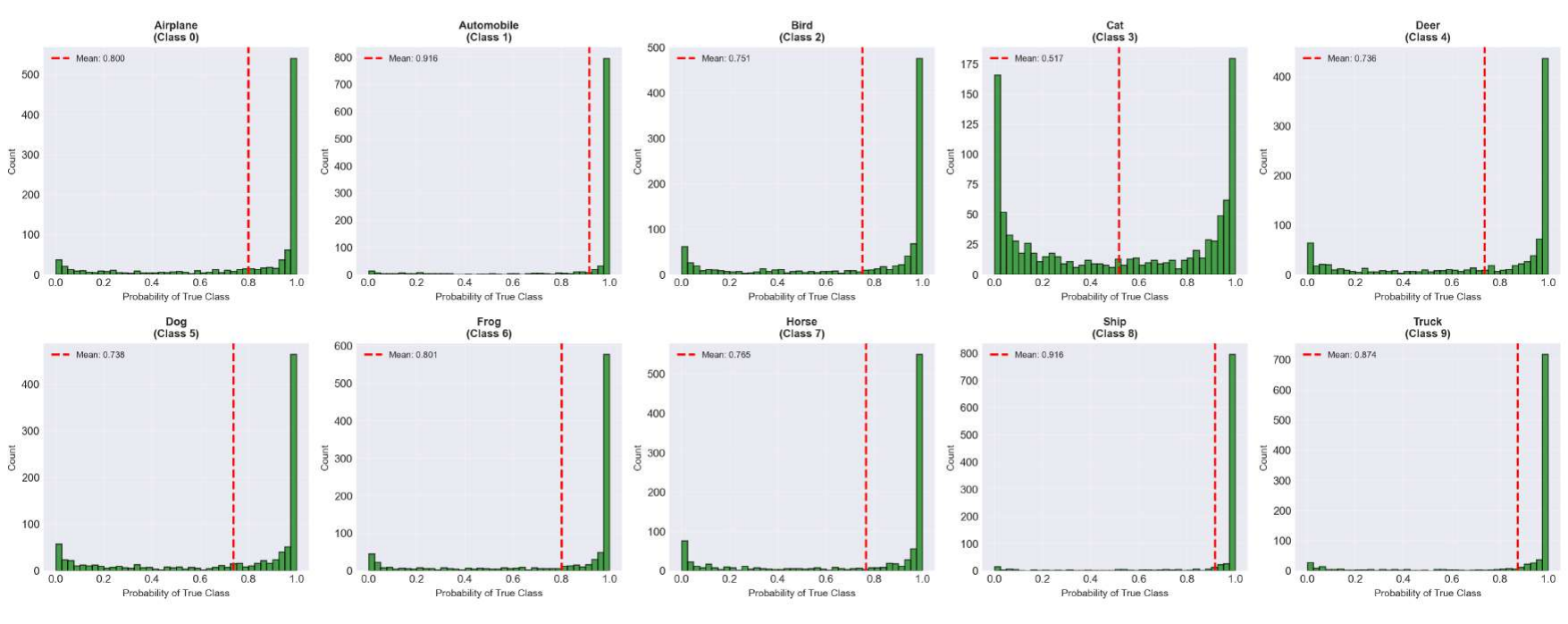}
  \caption{Per-class distribution of teacher confidence $p_{\text{true}}$. 
  Visually distinctive classes show concentrated mass near 
  $p_{\text{true}} \approx 1.0$, while ambiguous classes (\textit{cat}, 
  \textit{dog}) exhibit broader distributions. Red dashed line marks 
  per-class mean.}
  \label{fig:class_ptrue_dist}
\end{figure*}

$\mathrm{ConfVar}(x) = \mathrm{Var}_{j \neq c^*}(q_j)$ measures how
concentrated the teacher's residual probability mass is among competing classes.
Figure~\ref{fig:confvar} shows its distribution over the uncertain subset.
The body is broadly and roughly uniformly spread across $[0, 0.10]$,
providing fine-grained discrimination.
The spike at the theoretical maximum ($(C-2)/(C-1)^2 \approx 0.099$ for $C = 10$)
marks samples where the teacher's confusion is maximally concentrated
on a single competing class --- the clearest signal of structured, semantic confusion
rather than generic uncertainty.
\begin{figure*}[!t]
  \centering
  \includegraphics[width=\textwidth]{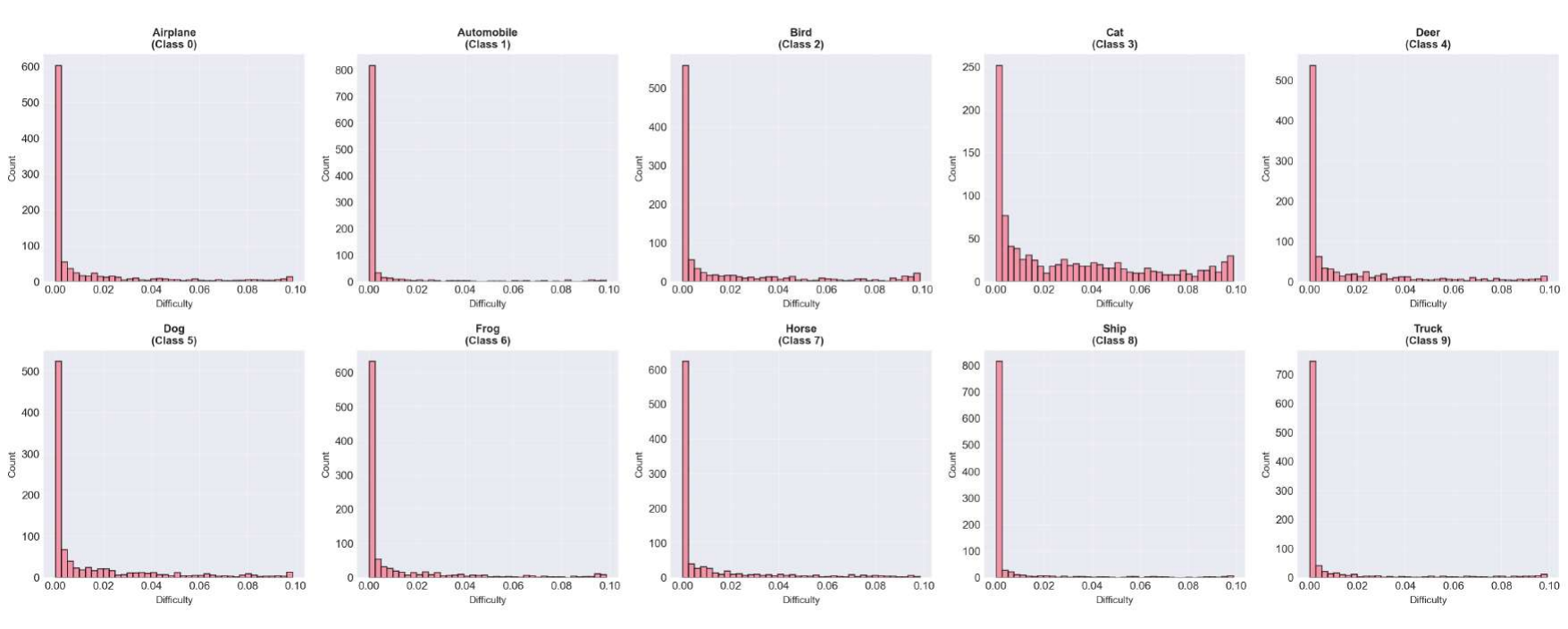}
  \caption{Per-class distribution of confusion-aware difficulty $D(x)$. 
  All classes show similar structure: mass near zero with a long tail. 
  \textit{Cat} exhibits highest mean difficulty, consistent with 
  semantic confusion with \textit{dog}.}
  \label{fig:class_difficulty_dist}
\end{figure*}

%% ----- A.3 -----
\subsection{Difficulty Score \texorpdfstring{$D(x)$}{D(x)}}
\label{app:difficulty}

Figure~\ref{fig:difficulty-log} shows $D(x)$ on a log-count axis.
36{,}902 samples fall in the first bin ($D(x) \approx 0$); the 90th-percentile
boundary lies near $D(x) = 0.0015$.
The remaining 10\% of samples populate a long, smoothly decaying tail
extending to $D(x) \approx 0.10$.
The log scale reveals that this tail is \emph{not} noise --- it decays
continuously rather than cutting off, indicating a coherent hard subset
rather than outliers.
The product formulation enforces that high $D(x)$ requires \emph{both}
low teacher confidence \emph{and} structured competing-class confusion,
filtering mislabeled or corrupted samples that inflate $1 - p_{\text{true}}$
without exhibiting class-specific confusion structure.

\subsection{Class-Level Distribution Analysis}
\label{app:class_distributions}

Figure~\ref{fig:class_ptrue_dist} shows the distribution of $p_{\text{true}}$ 
across CIFAR-10 classes. The per-class histograms reveal systematic 
differences in teacher confidence: visually distinctive classes 
(\textit{airplane}, \textit{automobile}, \textit{ship}, \textit{truck}) 
are dominated by high-confidence predictions (mass concentrated near 
$p_{\text{true}} \approx 1.0$), while visually ambiguous classes 
(\textit{cat}, \textit{dog}, \textit{deer}) exhibit broader 
distributions with substantial probability mass at lower confidence 
values. 
% This class-dependent variation explains why naive confidence-based 
% ranking induces class imbalance across difficulty stages.

Figure~\ref{fig:class_difficulty_dist} shows the distribution of $D(x)$ 
across classes. Unlike $p_{\text{true}}$, the composite score produces 
more balanced per-class distributions: while \textit{cat} (Class 3) 
retains the highest mean difficulty (consistent with cat-dog ambiguity), 
all classes exhibit a shared structure with most samples near zero and 
a long tail capturing genuinely ambiguous instances.

\subsection{Teacher Model Ablations}
\label{app:teacher_ablations}

\begin{figure*}[!t]
    \centering
    \begin{subfigure}[]{0.32\textwidth}
        \centering
        \includegraphics[width=\textwidth]{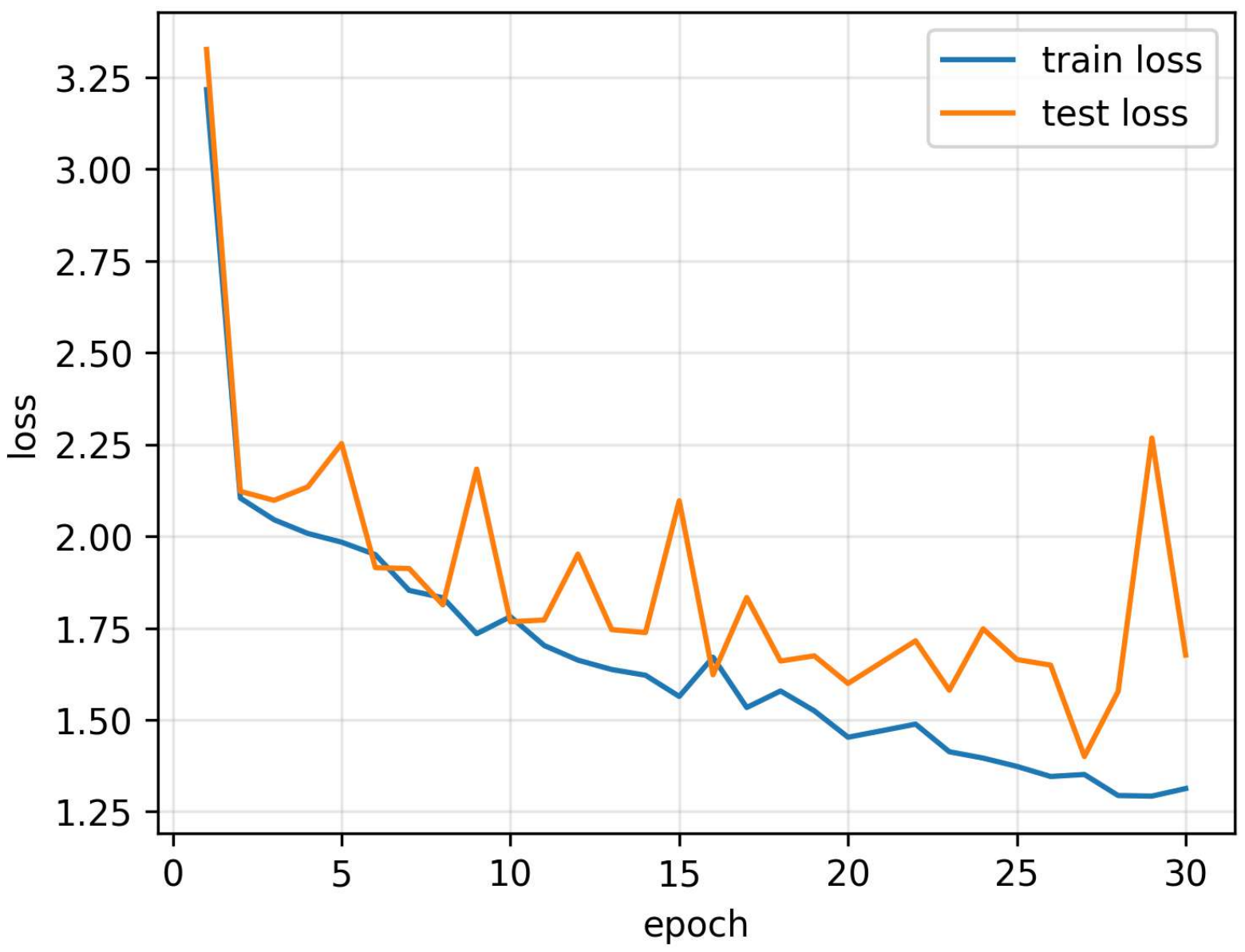}
        \caption{Training and test loss.}
        \label{fig:loss_10pct}
    \end{subfigure}
    \hfill
    \begin{subfigure}[]{0.32\textwidth}
        \centering
        \includegraphics[width=\textwidth]{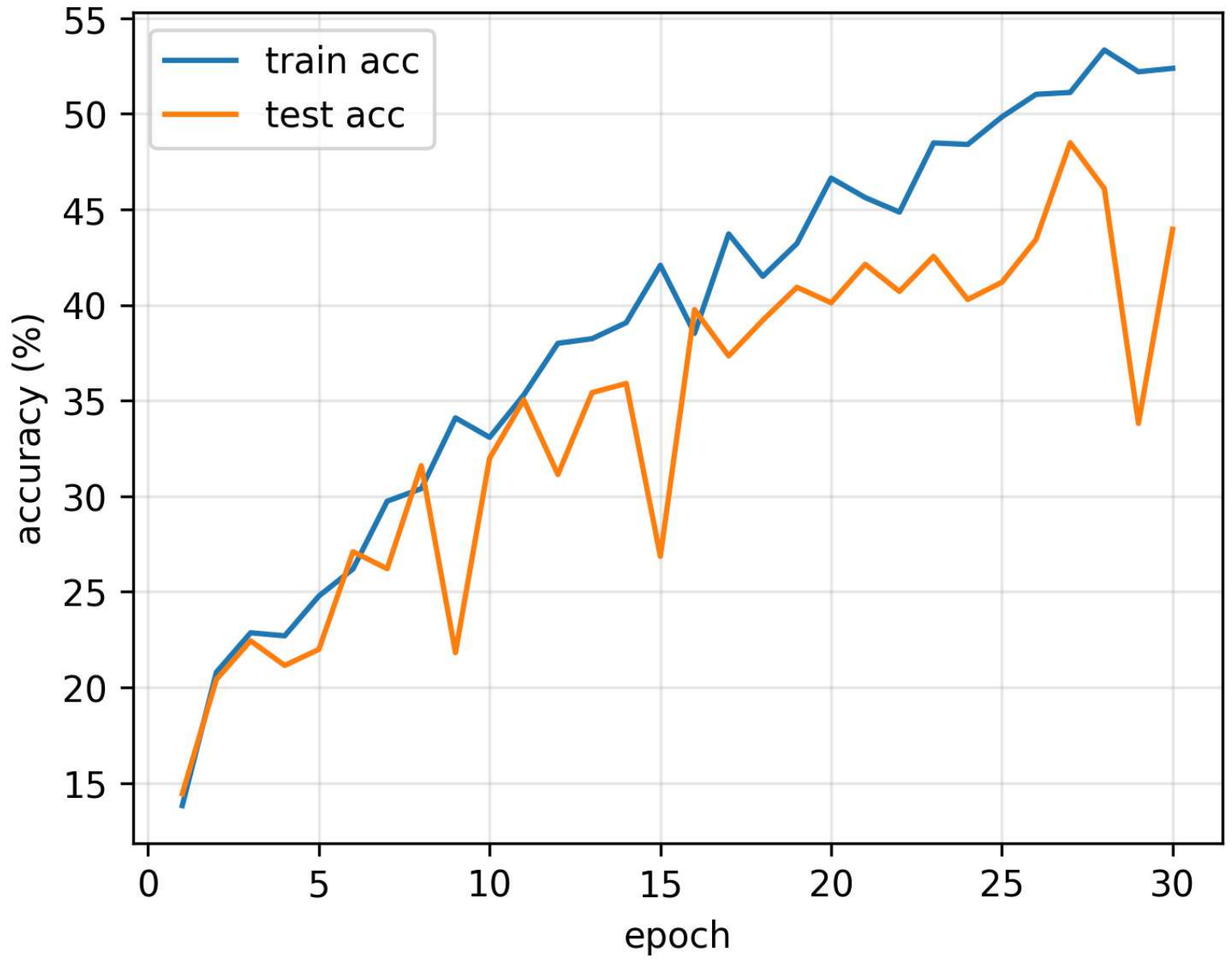}
        \caption{Training and test accuracy.}
        \label{fig:acc_10pct}
    \end{subfigure}
    \hfill
    \begin{subfigure}[]{0.32\textwidth}
        \centering
        \includegraphics[width=\textwidth]{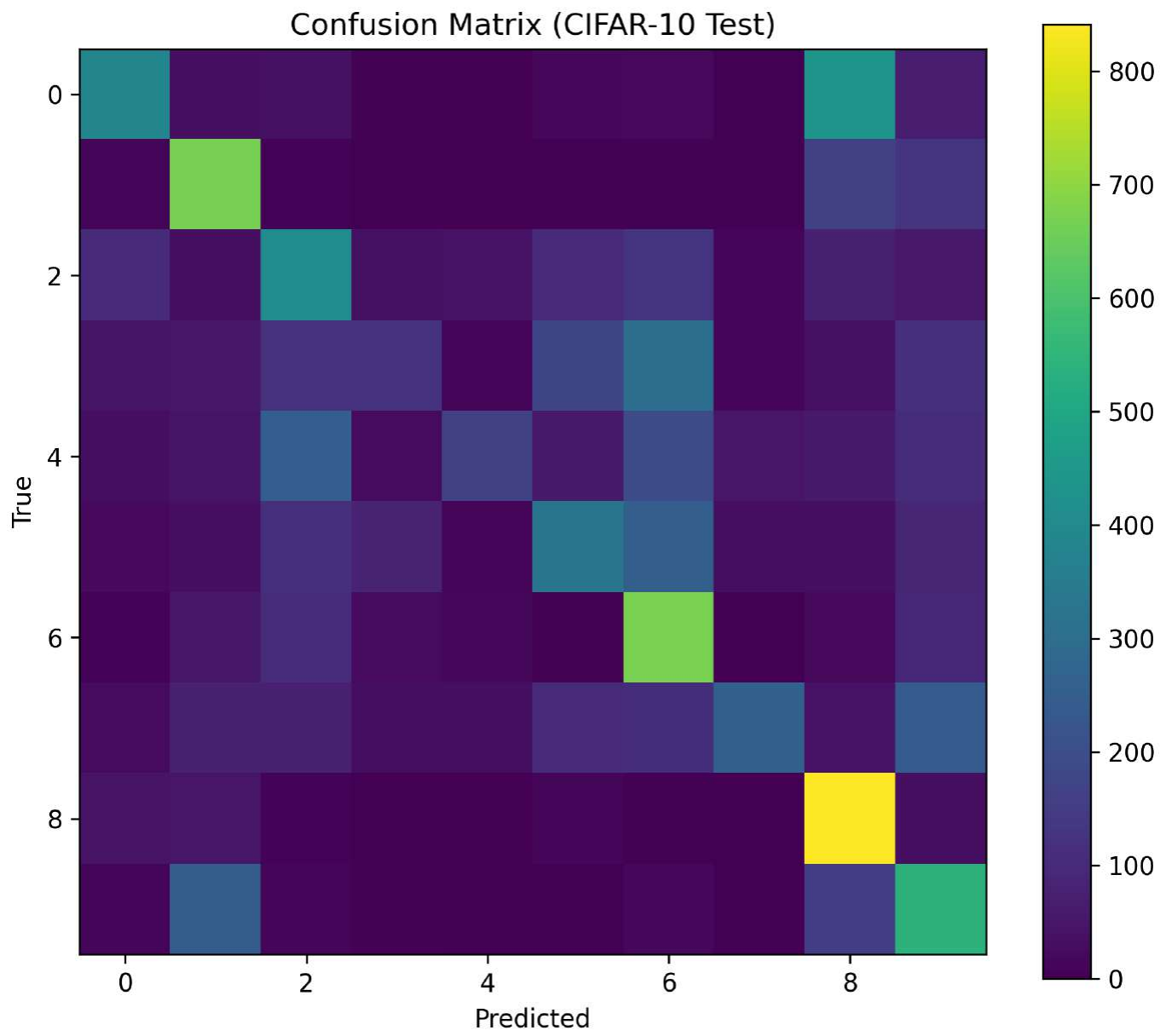}
        \caption{Test-set confusion matrix.}
        \label{fig:conf_10pct}
    \end{subfigure}
    \caption{Teacher ablation for the \textbf{10\% data regime} (Test Acc: $\approx$53\%). The model produces scattered confusion with no semantic structure, meaning its difficulty scores reflect noise rather than genuine sample hardness.}
    \label{fig:ablation_10pct}
\end{figure*}

\begin{figure*}[!t]
    \centering
    \begin{subfigure}[]{0.32\textwidth}
        \centering
        \includegraphics[width=\textwidth]{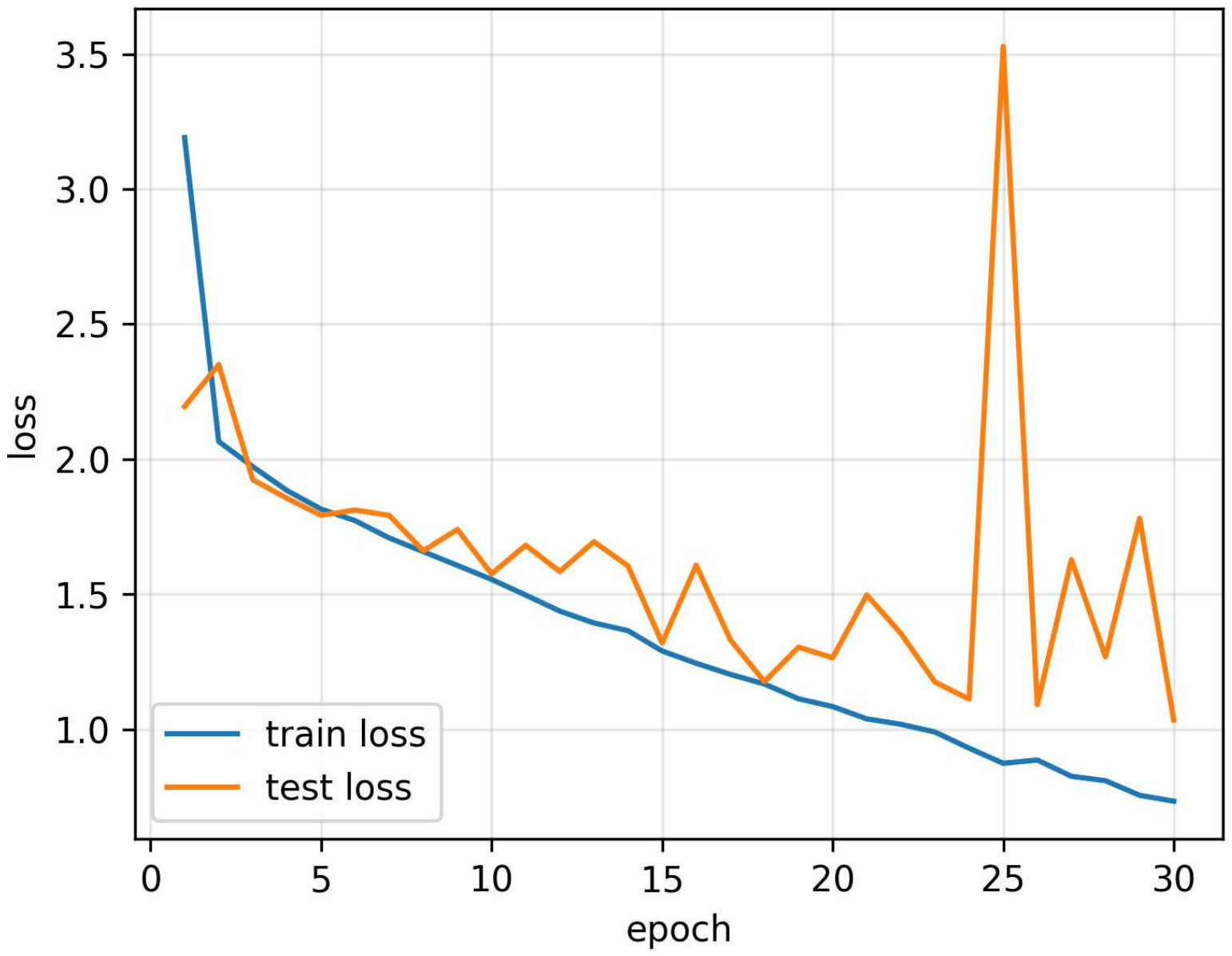}
        \caption{Training and test loss.}
        \label{fig:loss_20pct}
    \end{subfigure}
    \hfill
    \begin{subfigure}[]{0.32\textwidth}
        \centering
        \includegraphics[width=\textwidth]{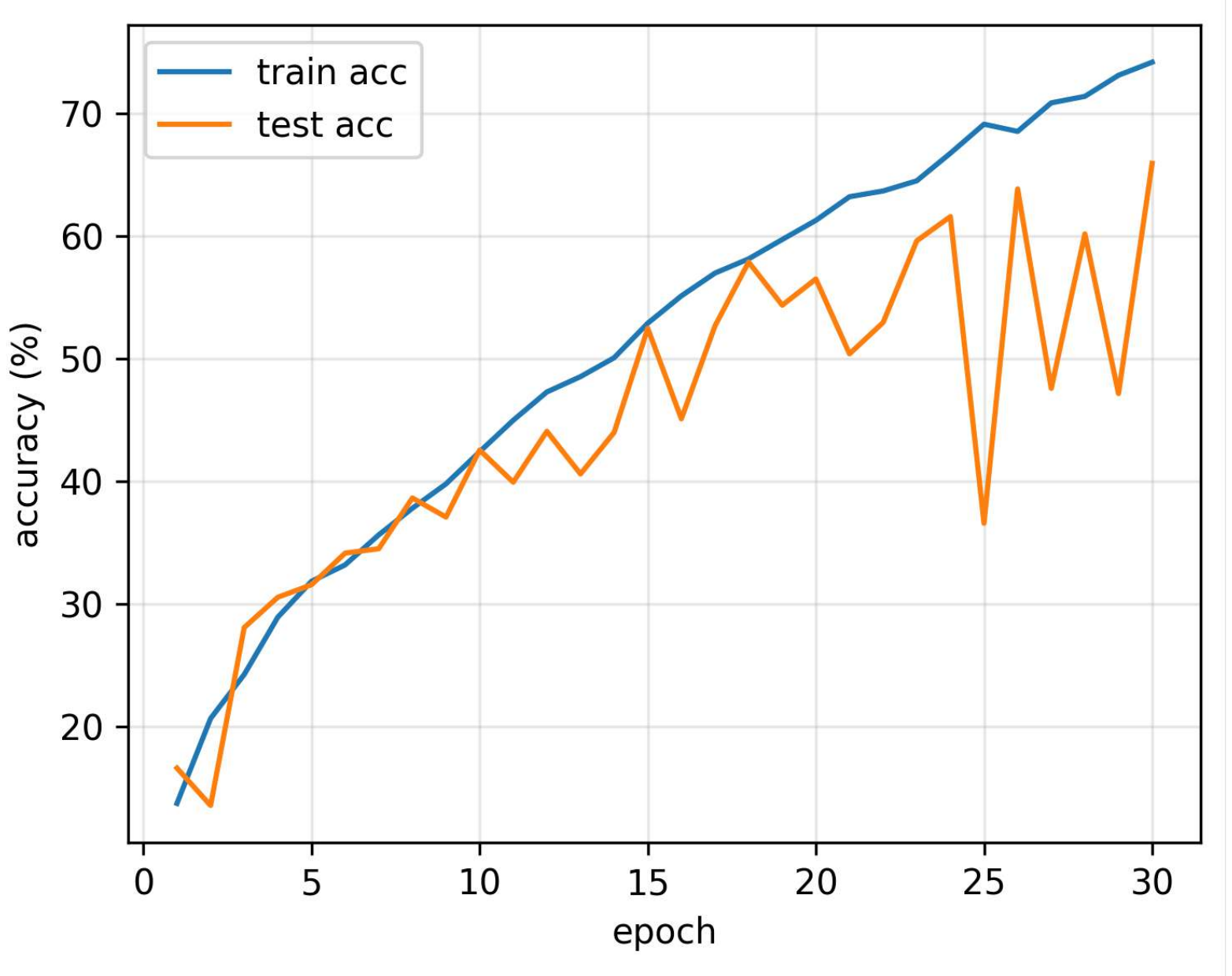}
        \caption{Training and test accuracy.}
        \label{fig:acc_20pct}
    \end{subfigure}
    \hfill
    \begin{subfigure}[]{0.32\textwidth}
        \centering
        \includegraphics[width=\textwidth]{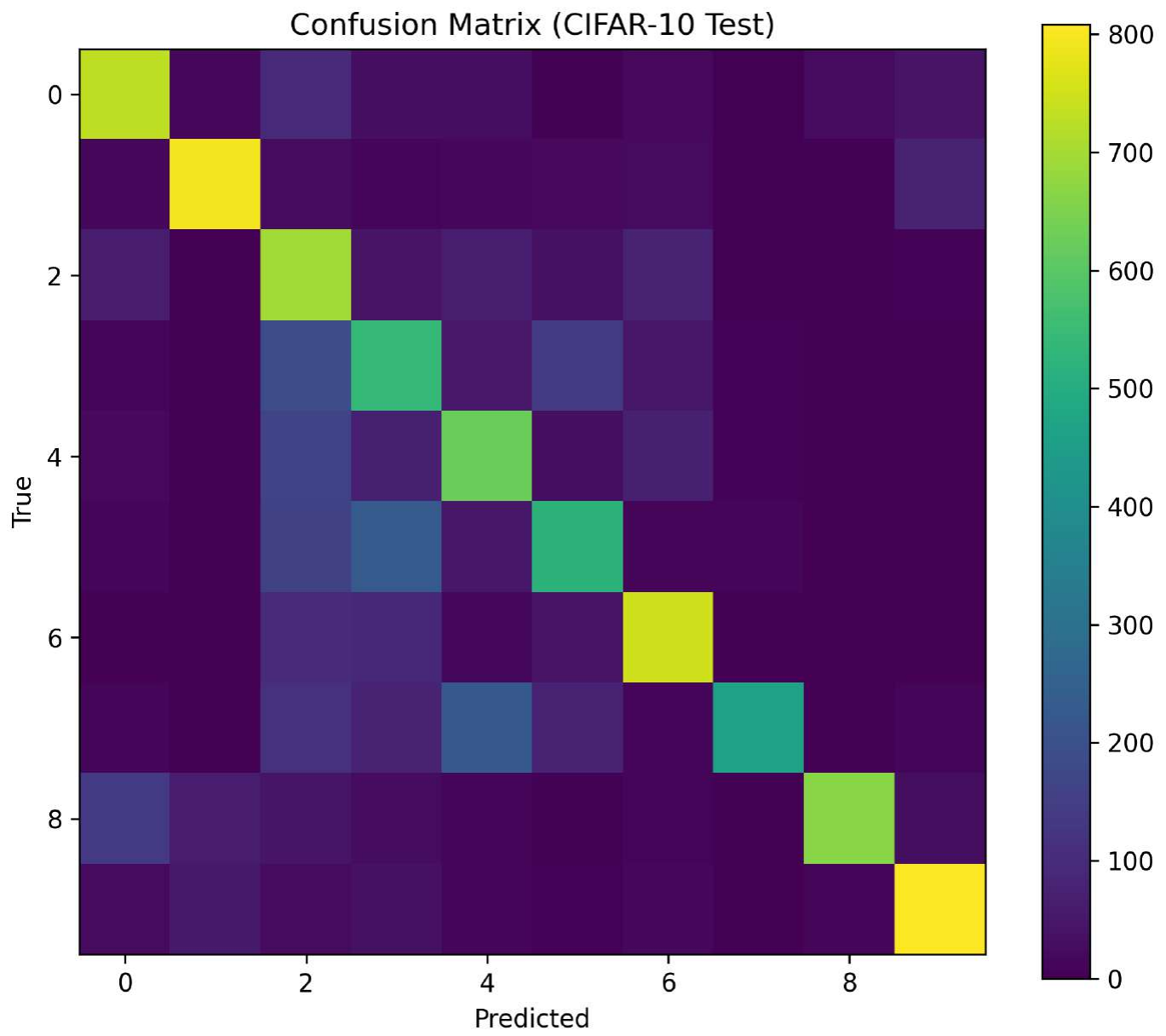}
        \caption{Test-set confusion matrix.}
        \label{fig:conf_20pct}
    \end{subfigure}
    \caption{Teacher ablation for the \textbf{20\% data regime} (Test Acc: $\approx$66\%). The confusion matrix begins to show diffuse inter-class structure, primarily among related visual categories.}
    \label{fig:ablation_20pct}
\end{figure*}

\begin{figure*}[!t]
    \centering
    \begin{subfigure}[]{0.32\textwidth}
        \centering
        \includegraphics[width=\textwidth]{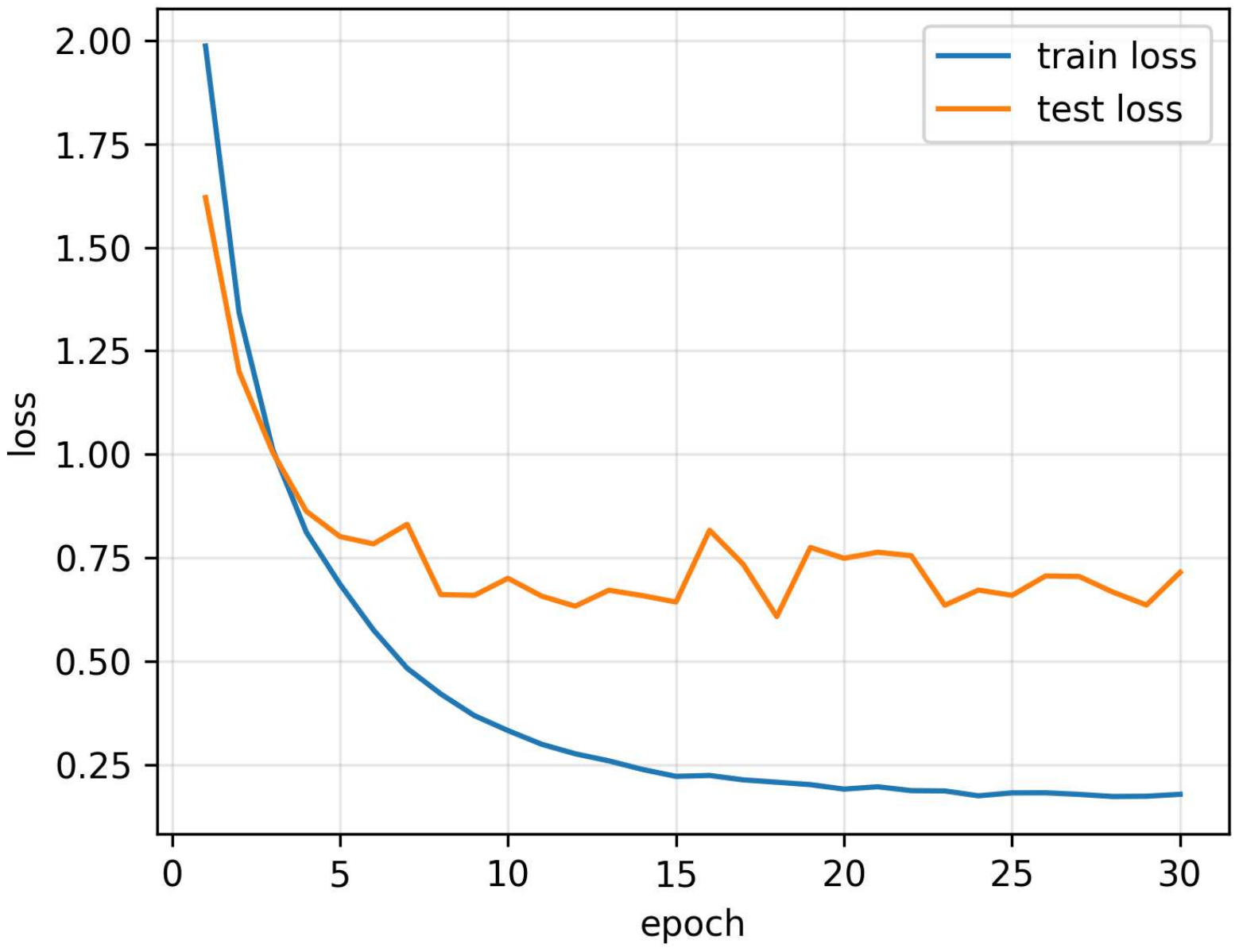}
        \caption{Training and test loss.}
        \label{fig:loss_100pct}
    \end{subfigure}
    \hfill
    \begin{subfigure}[]{0.32\textwidth}
        \centering
        \includegraphics[width=\textwidth]{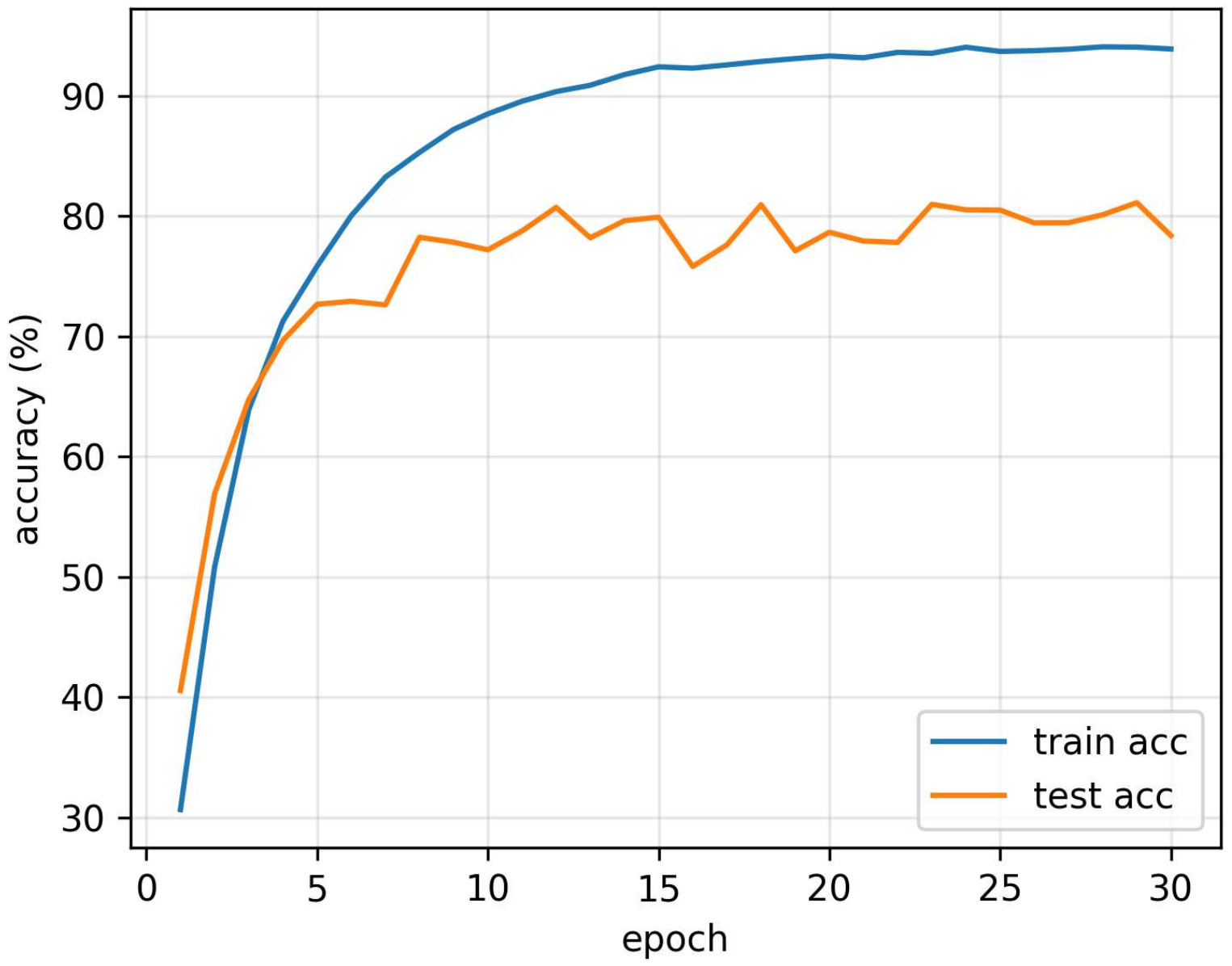}
        \caption{Training and test accuracy.}
        \label{fig:acc_100pct}
    \end{subfigure}
    \hfill
    \begin{subfigure}[]{0.32\textwidth}
        \centering
        \includegraphics[width=\textwidth]{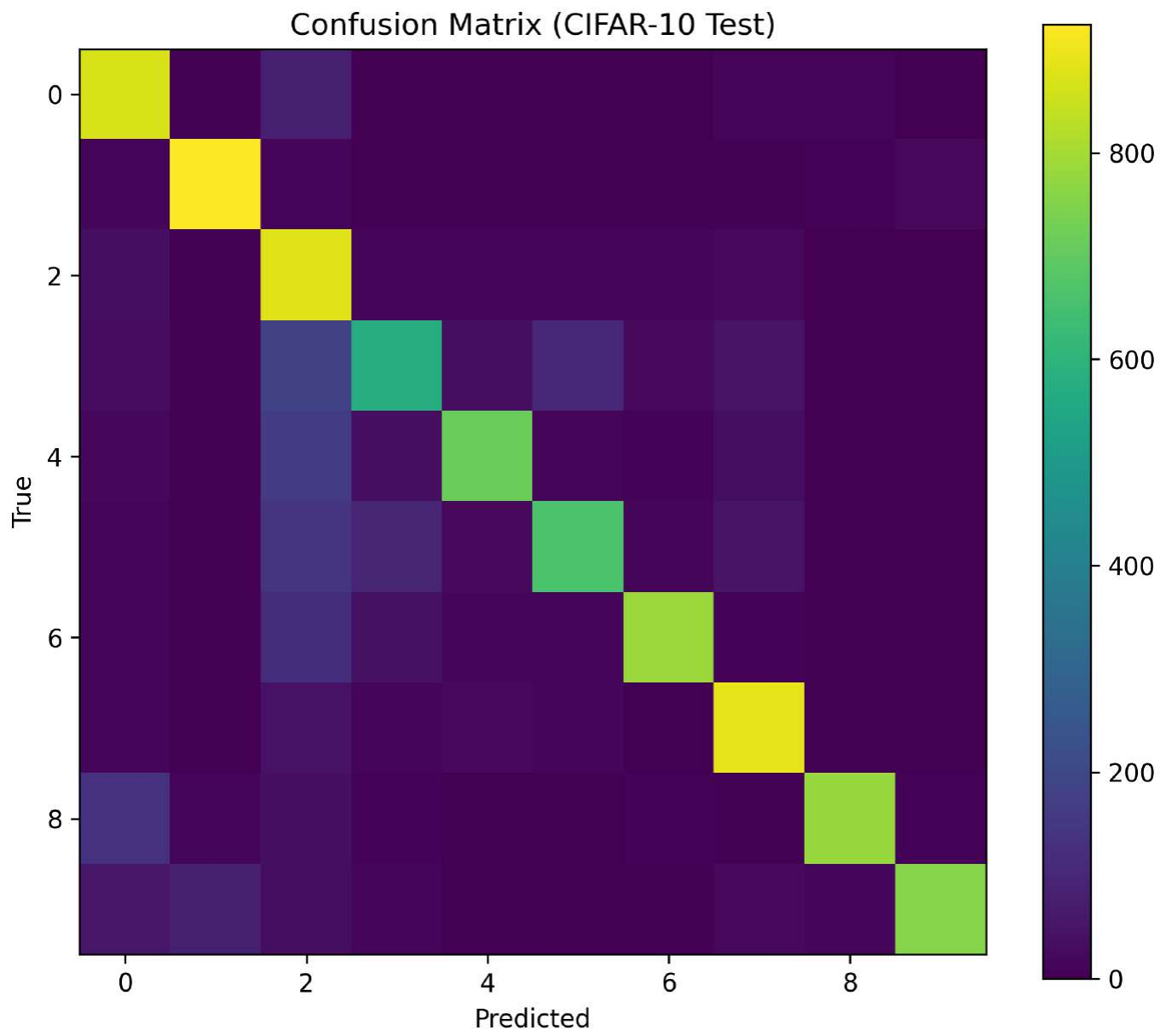}
        \caption{Test-set confusion matrix.}
        \label{fig:conf_100pct}
    \end{subfigure}
    \caption{Teacher ablation for the \textbf{100\% data regime} (Test Acc: 81.0\%). The model produces a strongly structured matrix (e.g., cat $\to$ dog), ensuring that the resulting difficulty scores capture meaningful ambiguity.}
    \label{fig:ablation_100pct}
\end{figure*}

Figures~\ref{fig:ablation_10pct}--\ref{fig:ablation_100pct} show 
training curves and confusion matrices for the ResNet-18 teacher 
across three data regimes. At 10\% (Test Acc: $\approx$53\%), the 
confusion matrix is scattered with no semantic structure, rendering 
difficulty scores unreliable. At 20\% (Test Acc: $\approx$66\%), diffuse 
inter-class structure begins to emerge. At 100\% (Test Acc: 81.0\%), strongly 
structured confusion (e.g., cat~$\to$~dog) confirms the teacher 
retains meaningful uncertainty at hard inter-class boundaries, 
producing a well-spread difficulty gradient suitable for curriculum 
scoring.

\subsection{Per-Class Qualitative Difficulty Visualization}
\label{app:qualitative_viz}
\begin{figure*}[!t]
  \centering
  \begin{subfigure}[]{0.48\textwidth}
    \includegraphics[width=\linewidth]{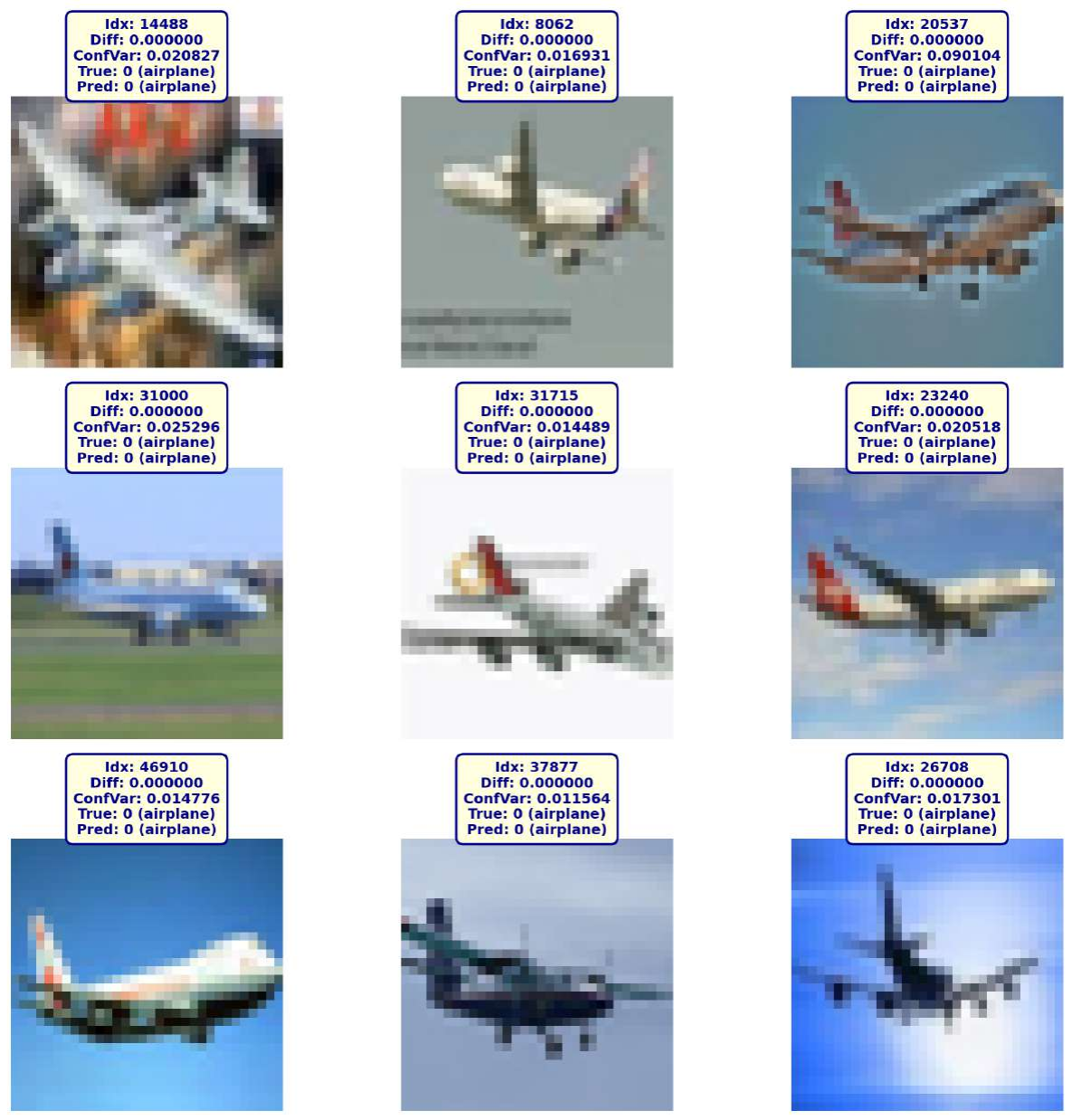}
    \caption{Easiest \textit{airplane} samples.}
  \end{subfigure}
  \hfill
  \begin{subfigure}[]{0.48\textwidth}
    \includegraphics[width=\linewidth]{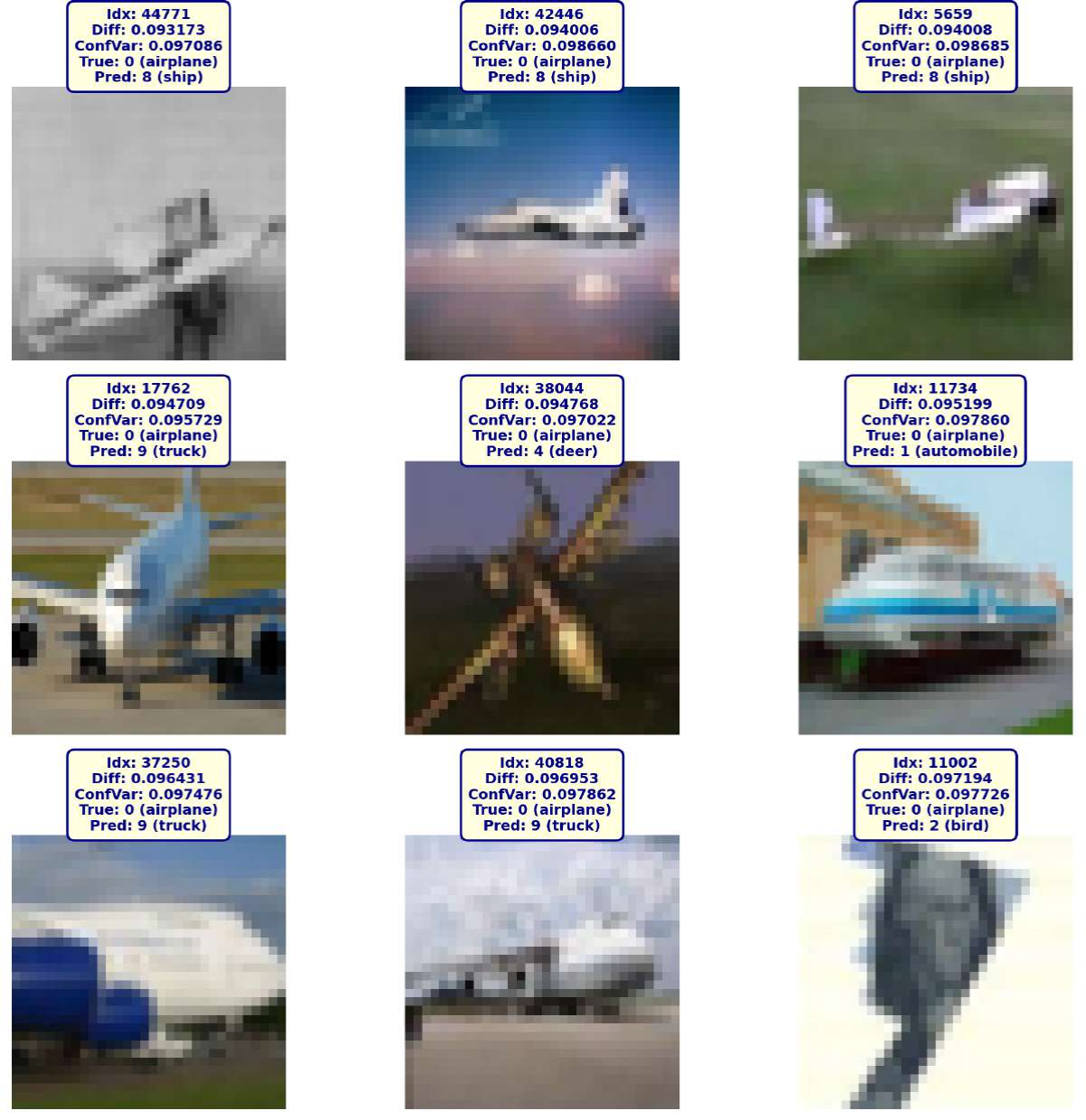}
    \caption{Most difficult \textit{airplane} samples.}
  \end{subfigure}
  \caption{Airplane class.}
  \label{fig:viz_airplane}
\end{figure*}

\begin{figure*}[!t]
  \centering
  \begin{subfigure}[]{0.48\textwidth}
    \includegraphics[width=\linewidth]{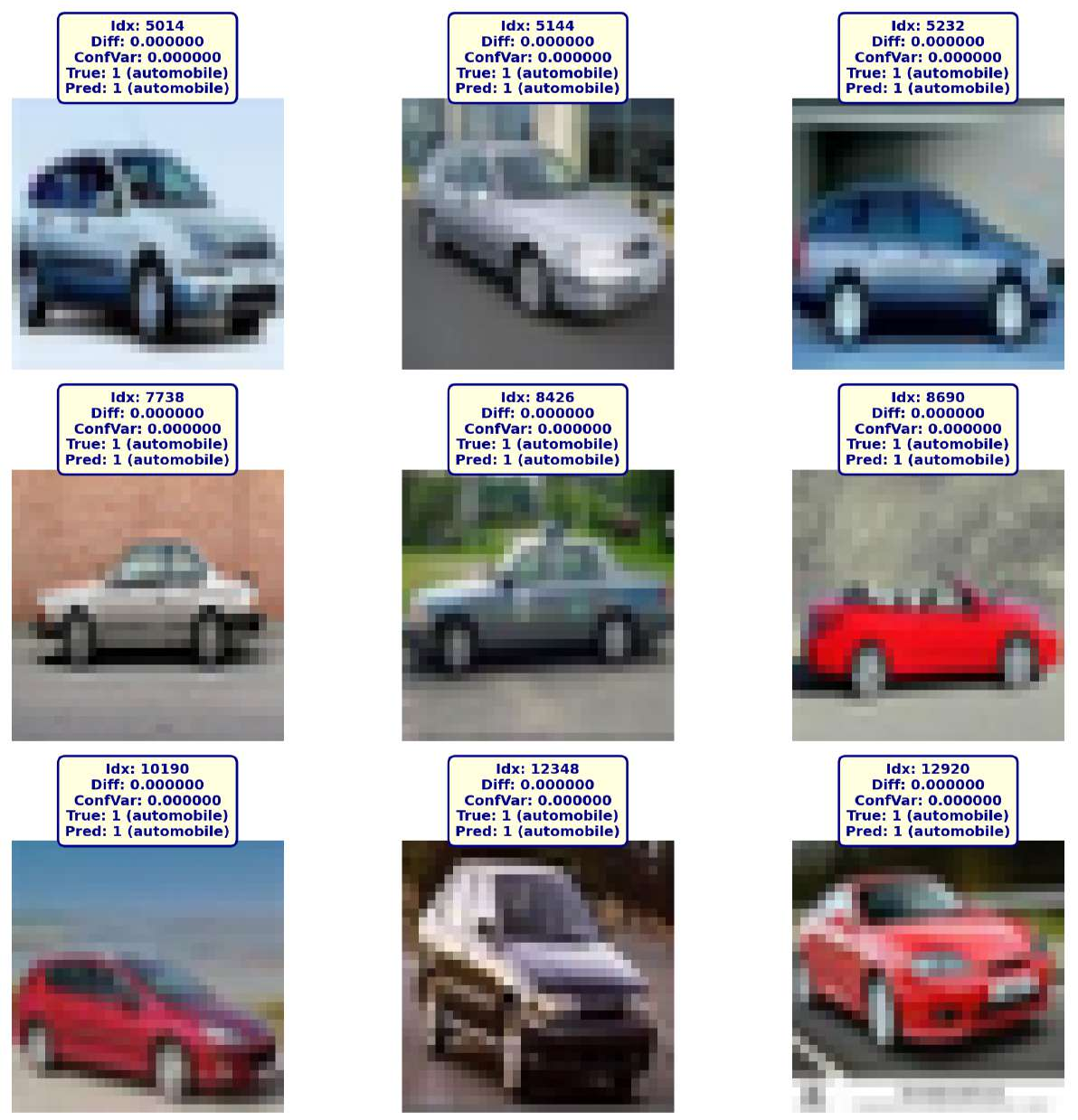}
    \caption{Easiest \textit{automobile} samples.}
  \end{subfigure}
  \hfill
  \begin{subfigure}[]{0.48\textwidth}
    \includegraphics[width=\linewidth]{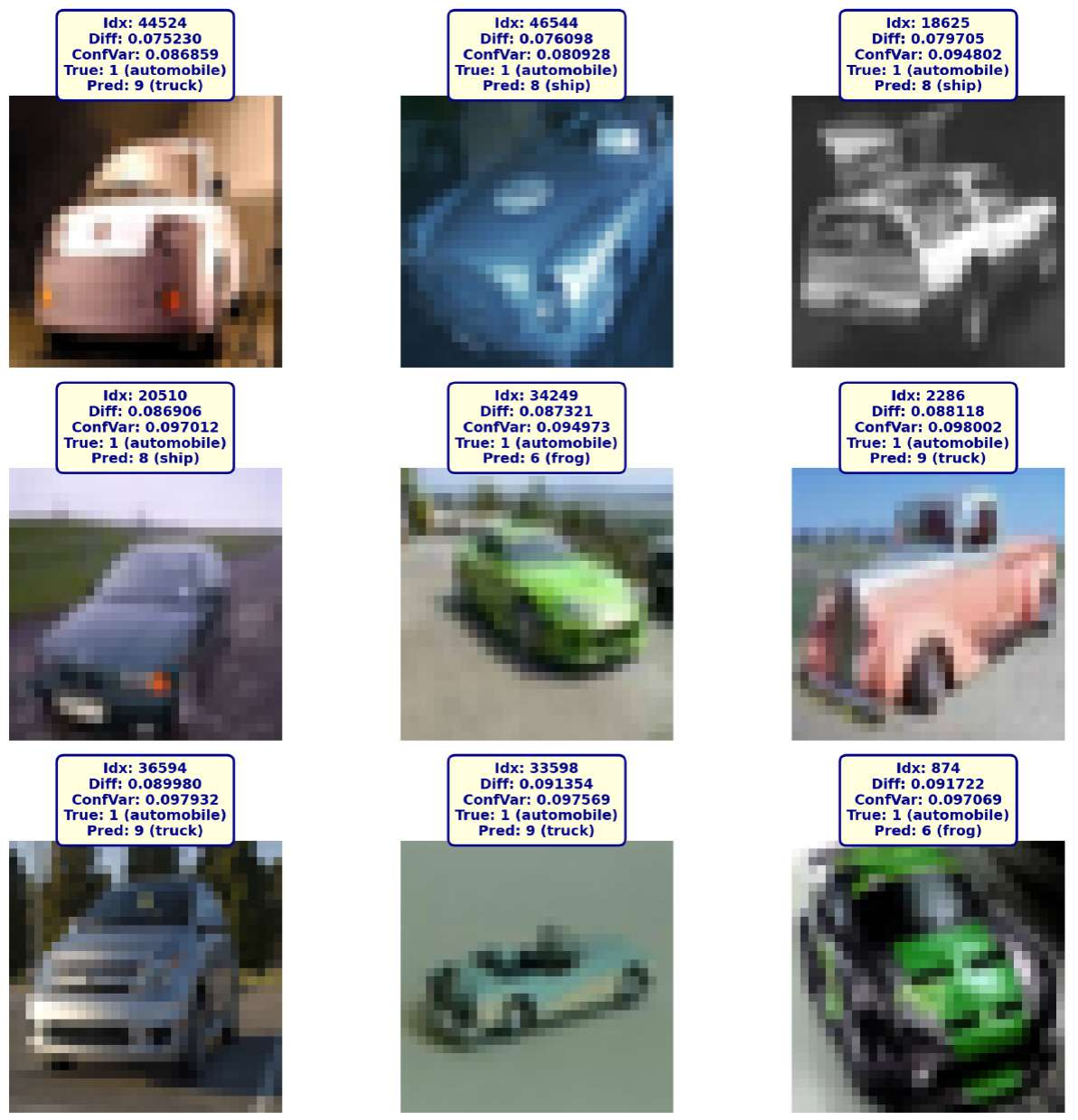}
    \caption{Most difficult \textit{automobile} samples.}
  \end{subfigure}
  \caption{Automobile class.}
  \label{fig:viz_automobile}
\end{figure*}

\begin{figure*}[!t]
  \centering
  \begin{subfigure}[]{0.48\textwidth}
    \includegraphics[width=\linewidth]{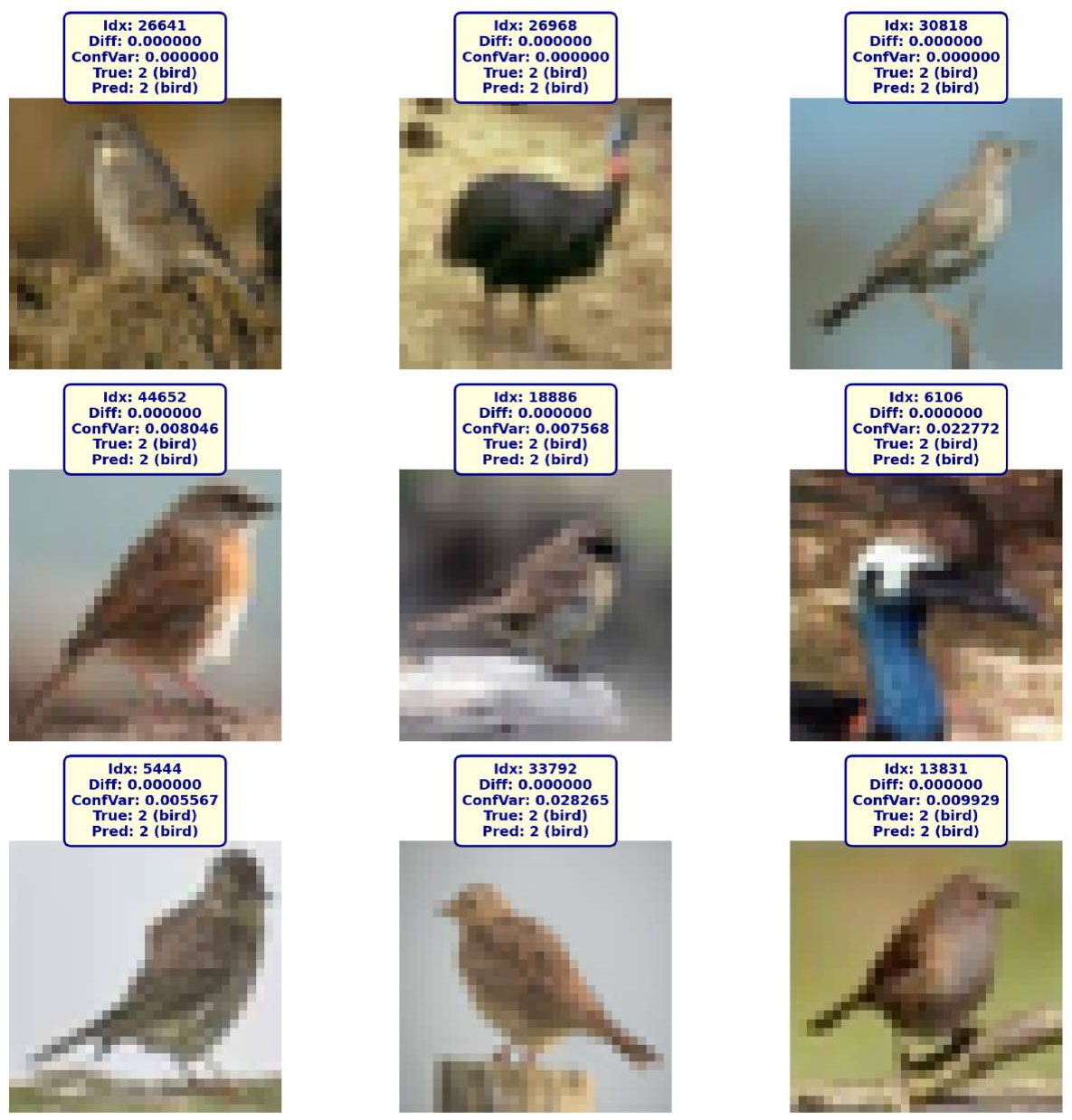}
    \caption{Easiest \textit{bird} samples.}
  \end{subfigure}
  \hfill
  \begin{subfigure}[]{0.48\textwidth}
    \includegraphics[width=\linewidth]{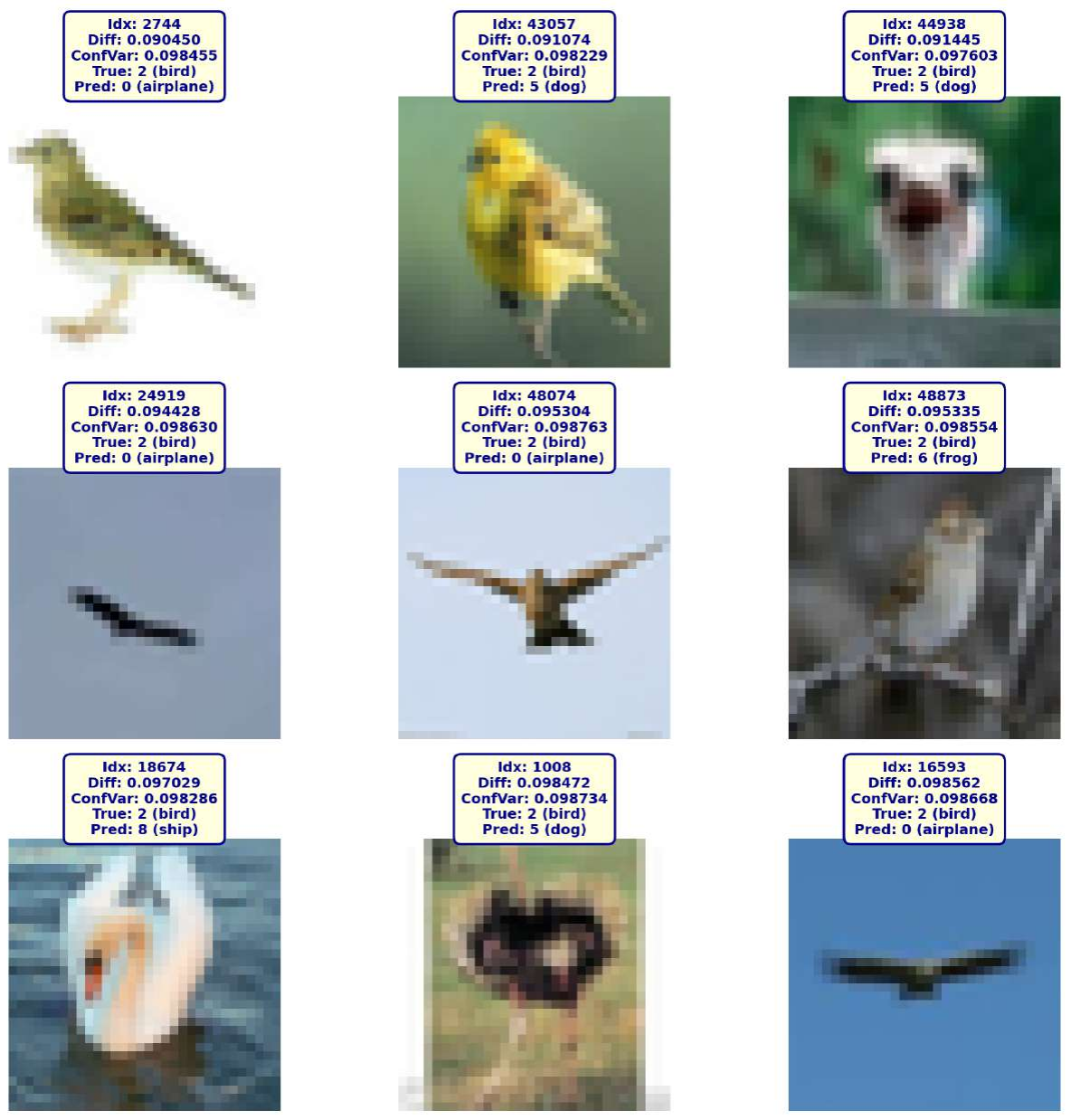}
    \caption{Most difficult \textit{bird} samples.}
  \end{subfigure}
  \caption{Bird class.}
  \label{fig:viz_bird}
\end{figure*}

\begin{figure*}[!t]
  \centering
  \begin{subfigure}[]{0.48\textwidth}
    \includegraphics[width=\linewidth]{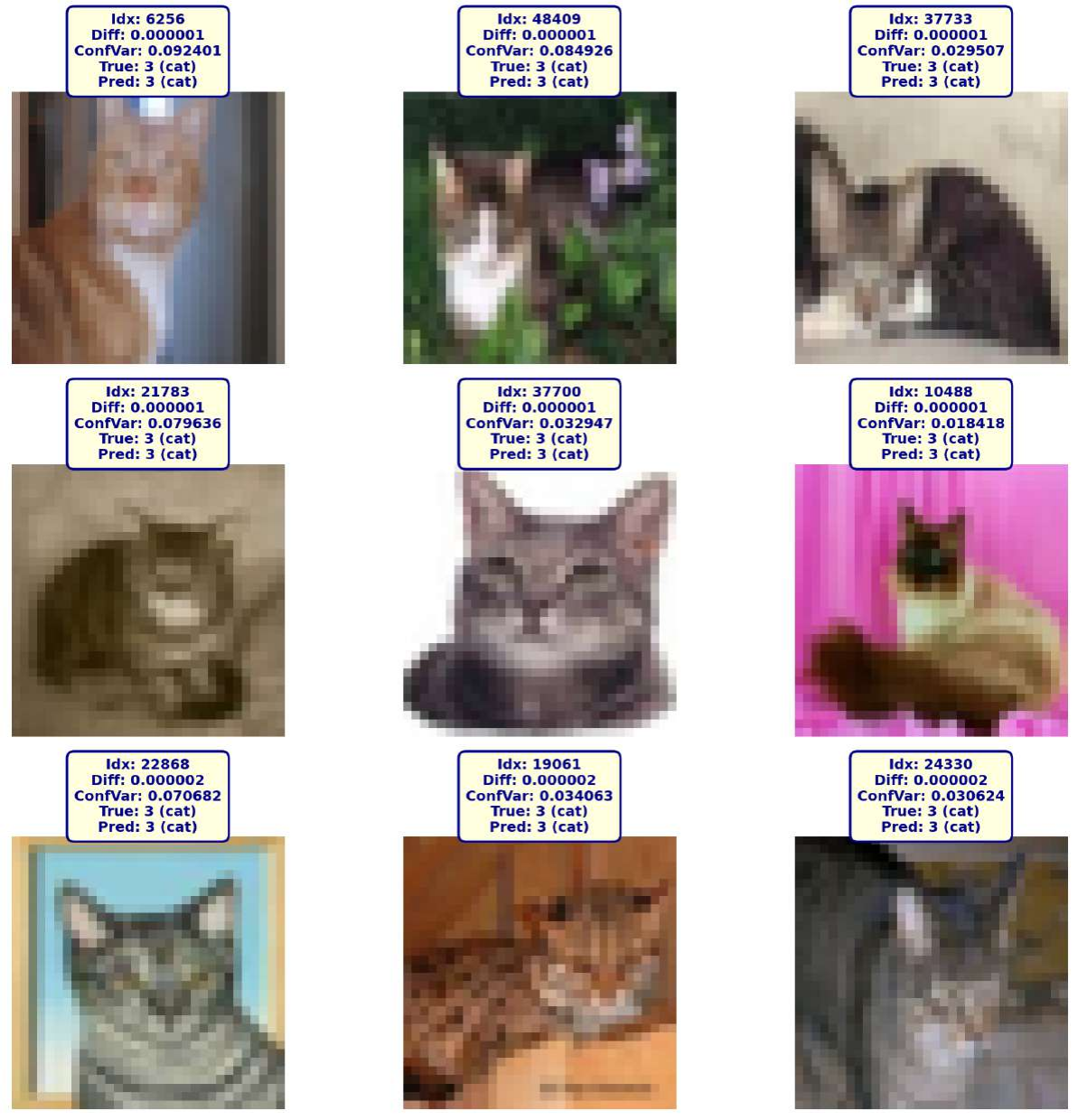}
    \caption{Easiest \textit{cat} samples.}
  \end{subfigure}
  \hfill
  \begin{subfigure}[]{0.48\textwidth}
    \includegraphics[width=\linewidth]{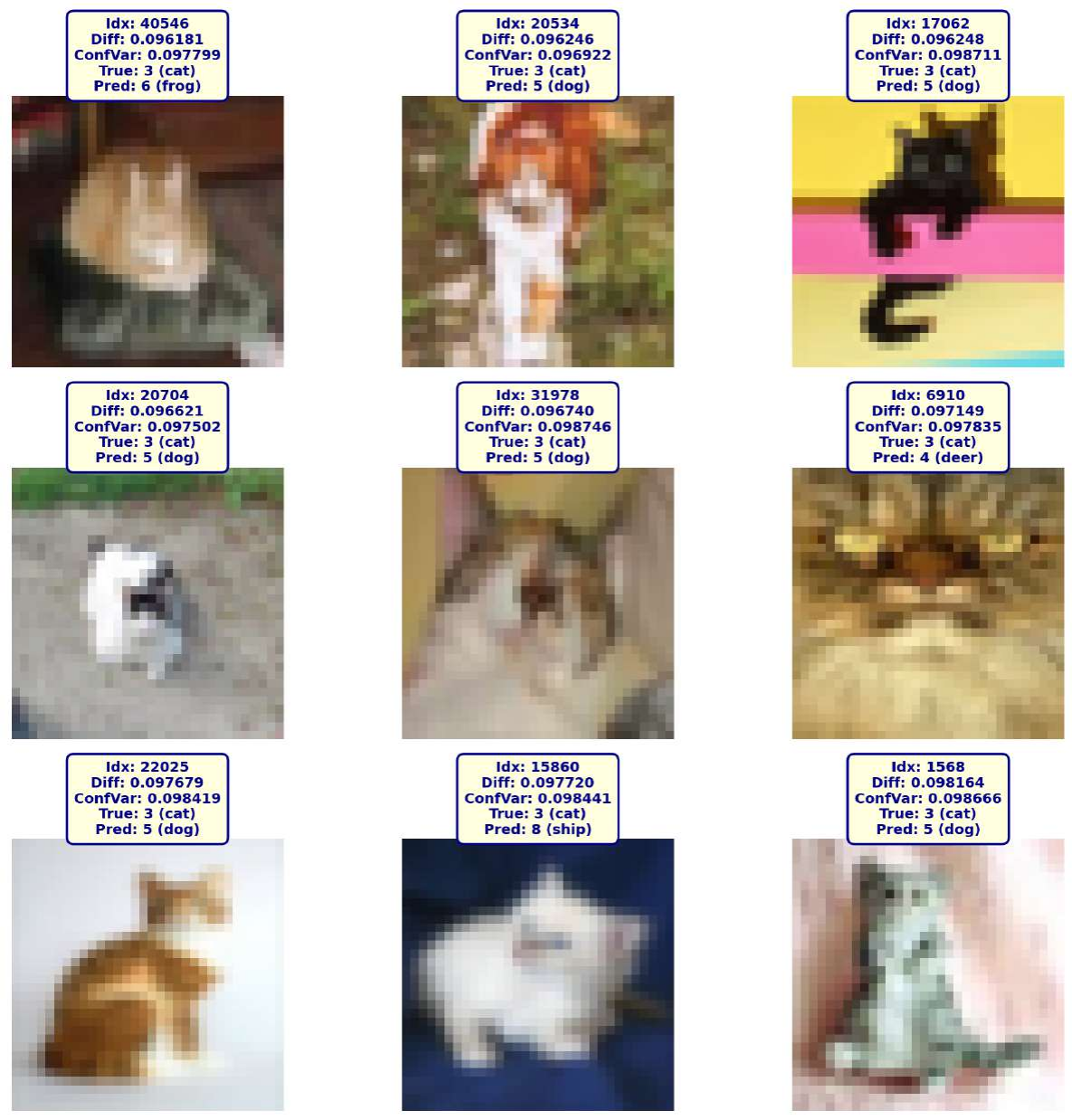}
    \caption{Most difficult \textit{cat} samples.}
  \end{subfigure}
  \caption{Cat class.}
  \label{fig:viz_cat}
\end{figure*}

\begin{figure*}[!t]
  \centering
  \begin{subfigure}[]{0.48\textwidth}
    \includegraphics[width=\linewidth]{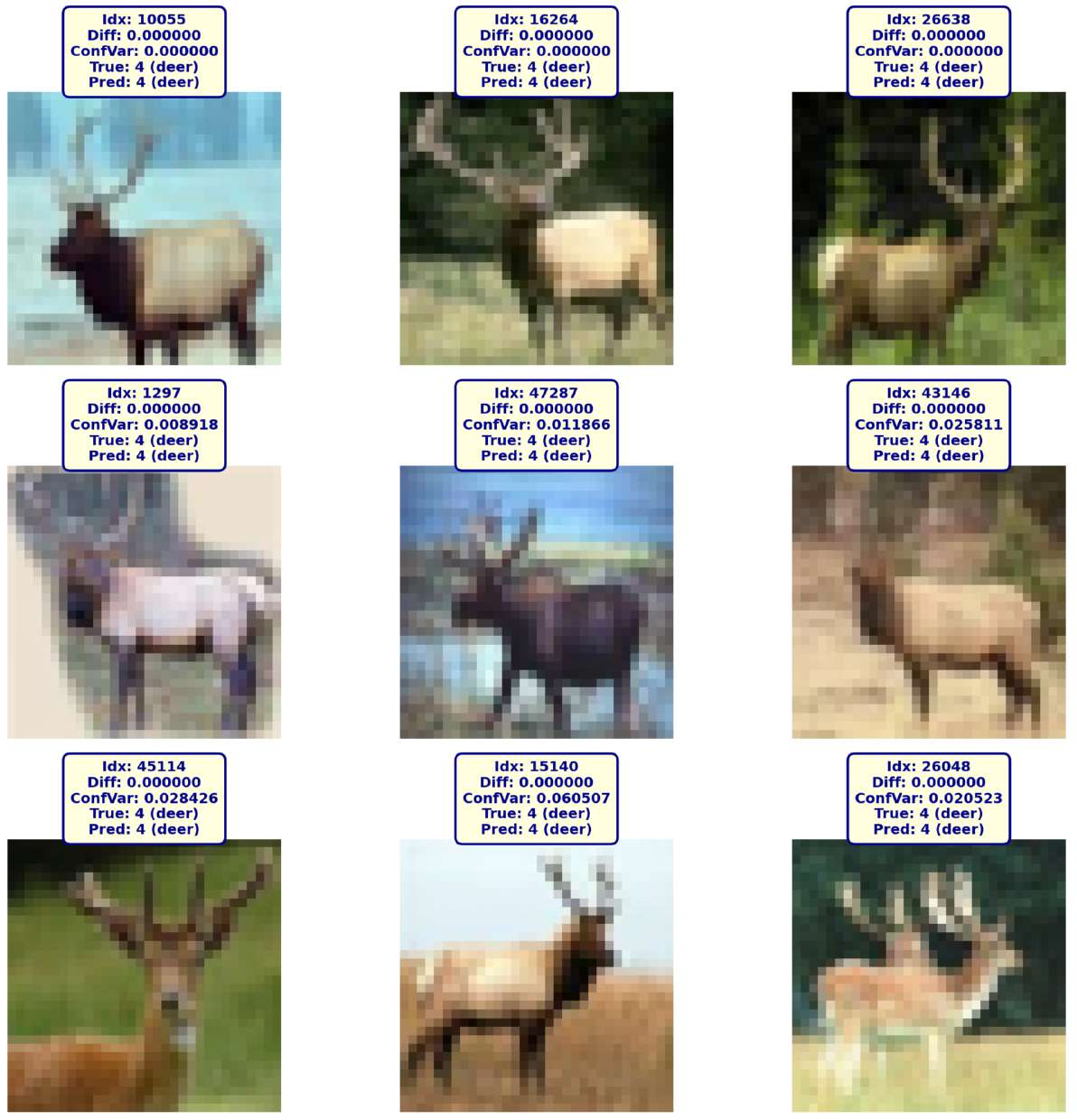}
    \caption{Easiest \textit{deer} samples.}
  \end{subfigure}
  \hfill
  \begin{subfigure}[]{0.48\textwidth}
    \includegraphics[width=\linewidth]{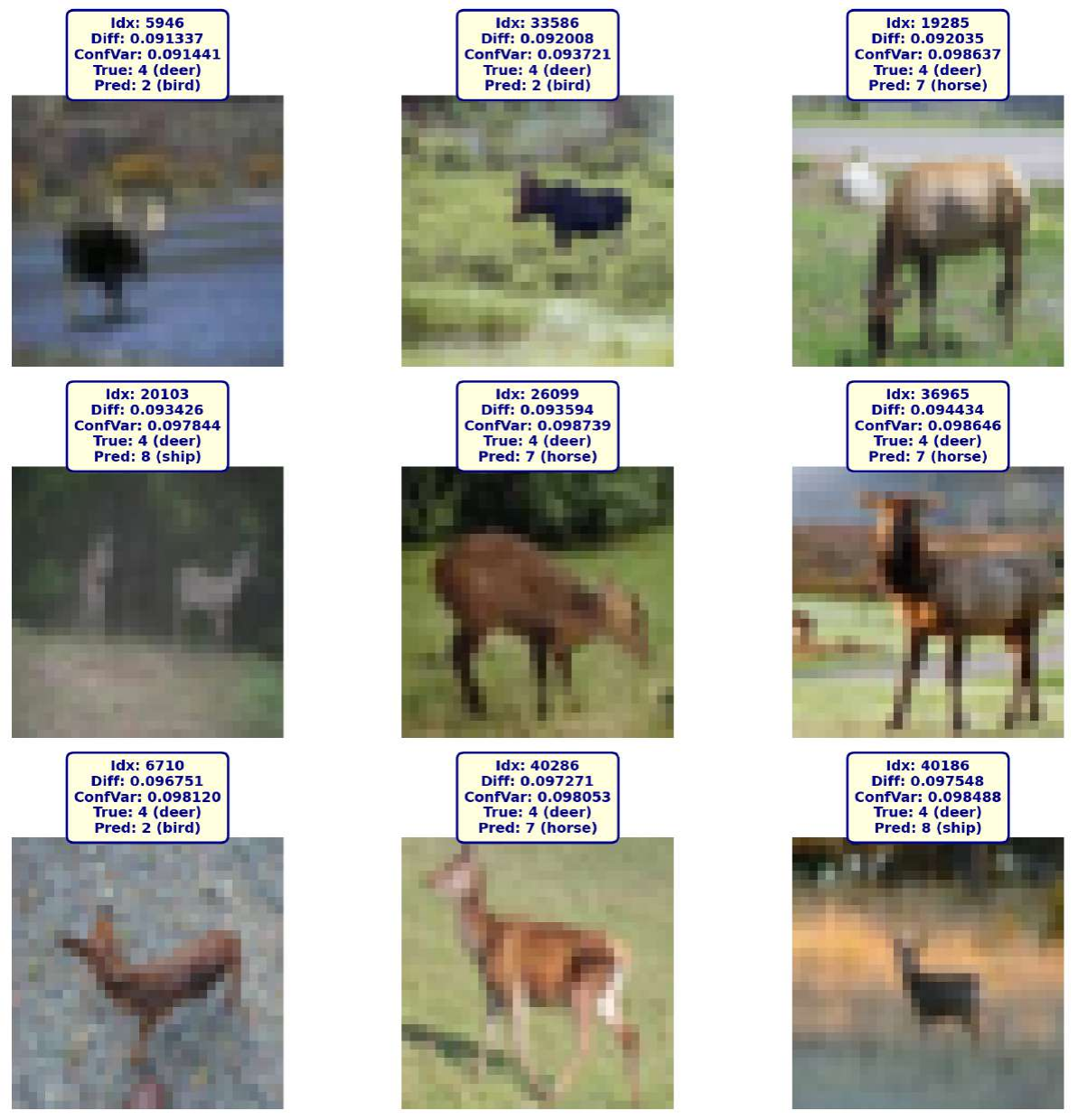}
    \caption{Most difficult \textit{deer} samples.}
  \end{subfigure}
  \caption{Deer class.}
  \label{fig:viz_deer}
\end{figure*}

\begin{figure*}[!t]
  \centering
  \begin{subfigure}[]{0.48\textwidth}
    \includegraphics[width=\linewidth]{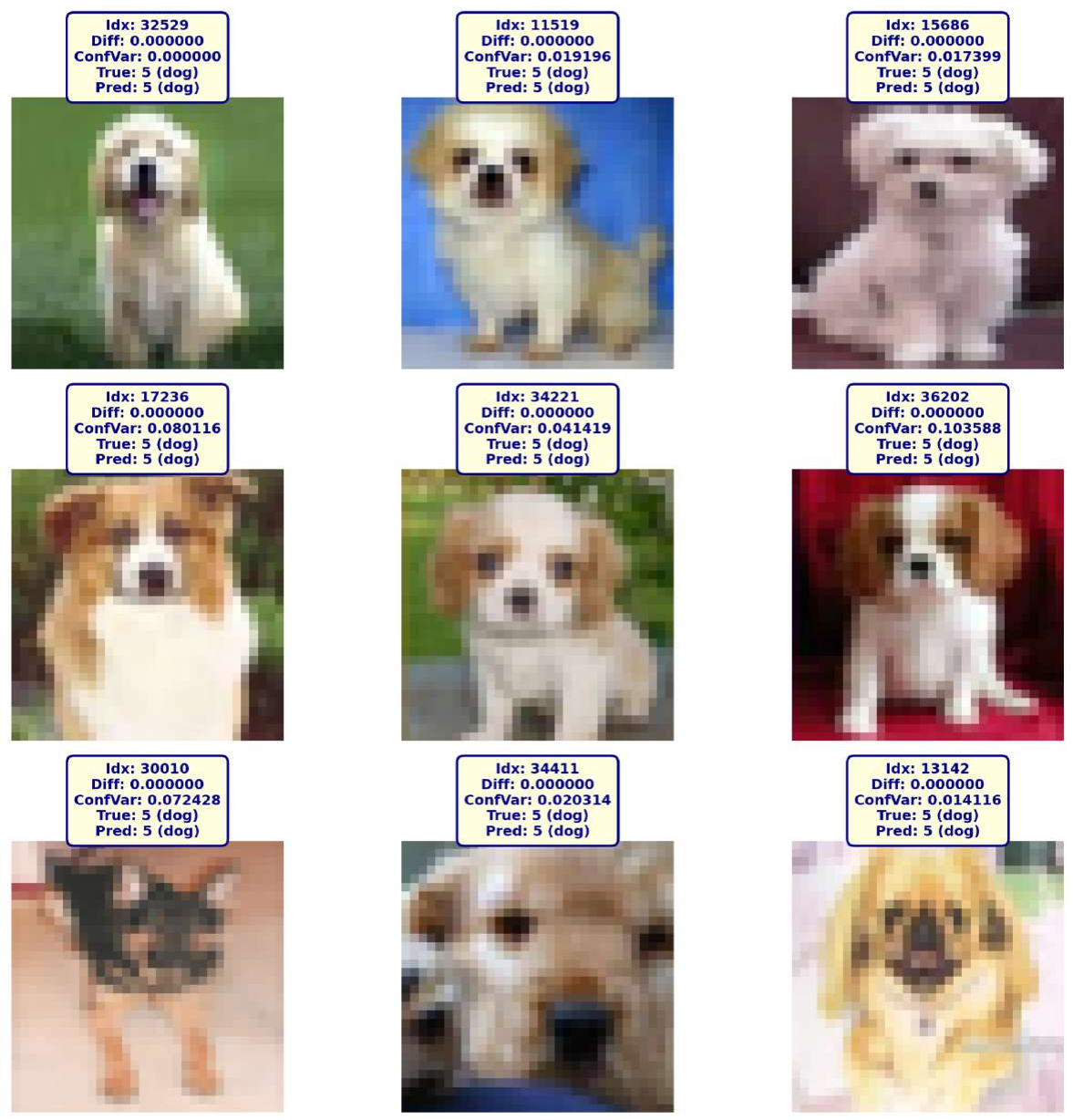}
    \caption{Easiest \textit{dog} samples.}
  \end{subfigure}
  \hfill
  \begin{subfigure}[]{0.48\textwidth}
    \includegraphics[width=\linewidth]{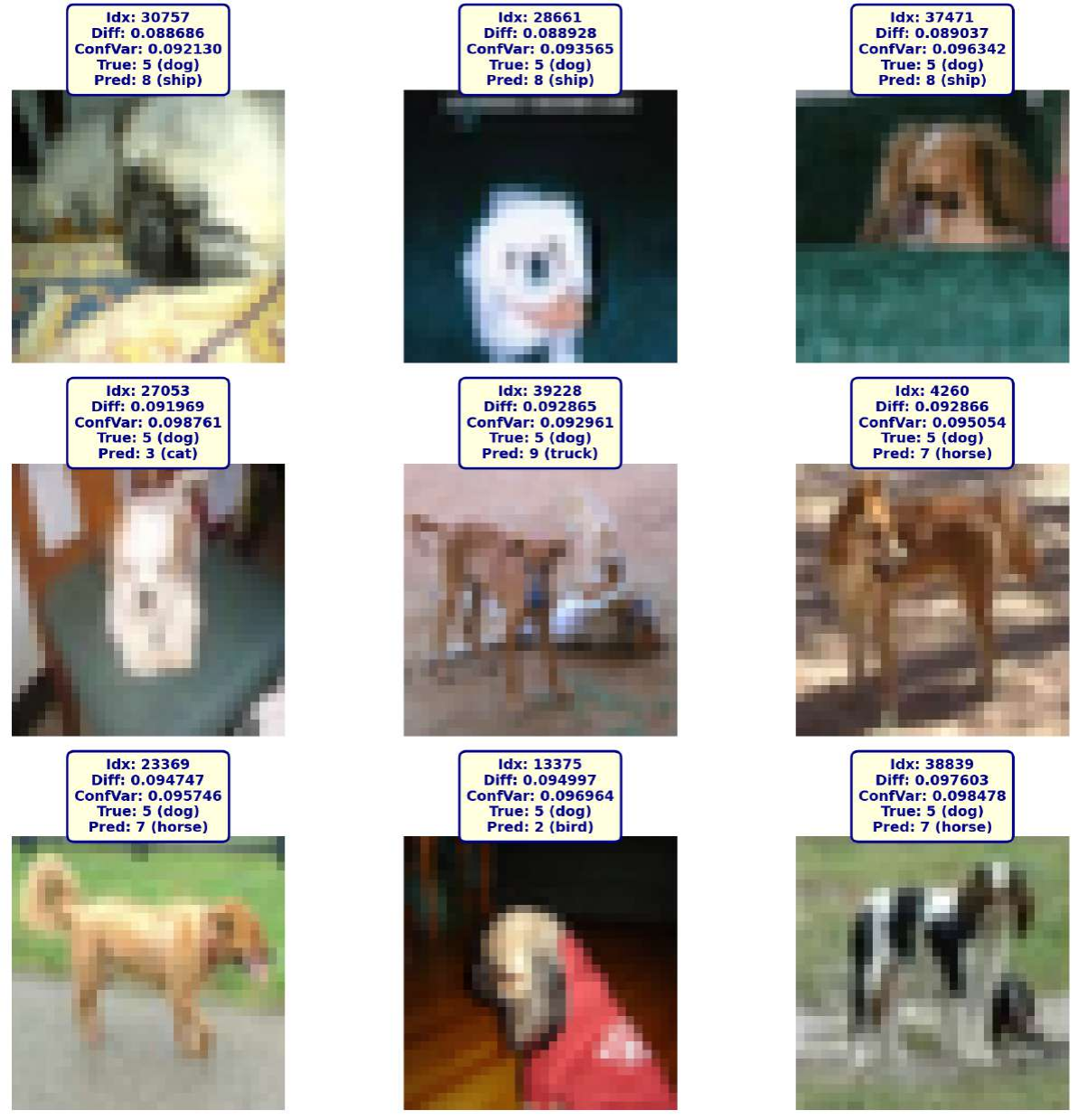}
    \caption{Most difficult \textit{dog} samples.}
  \end{subfigure}
  \caption{Dog class.}
  \label{fig:viz_dog}
\end{figure*}

\begin{figure*}[!t]
  \centering
  \begin{subfigure}[]{0.48\textwidth}
    \includegraphics[width=\linewidth]{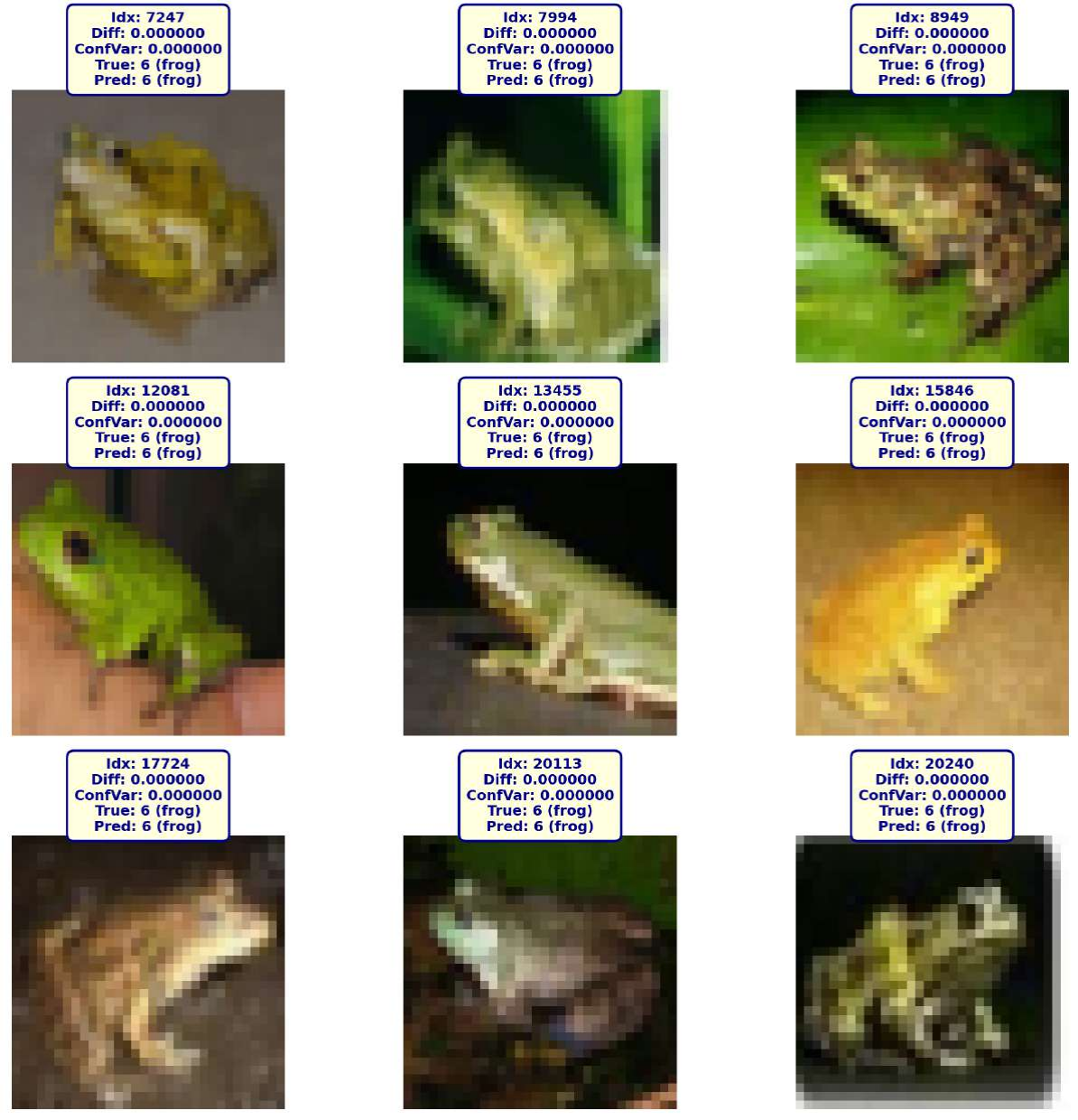}
    \caption{Easiest \textit{frog} samples.}
  \end{subfigure}
  \hfill
  \begin{subfigure}[]{0.48\textwidth}
    \includegraphics[width=\linewidth]{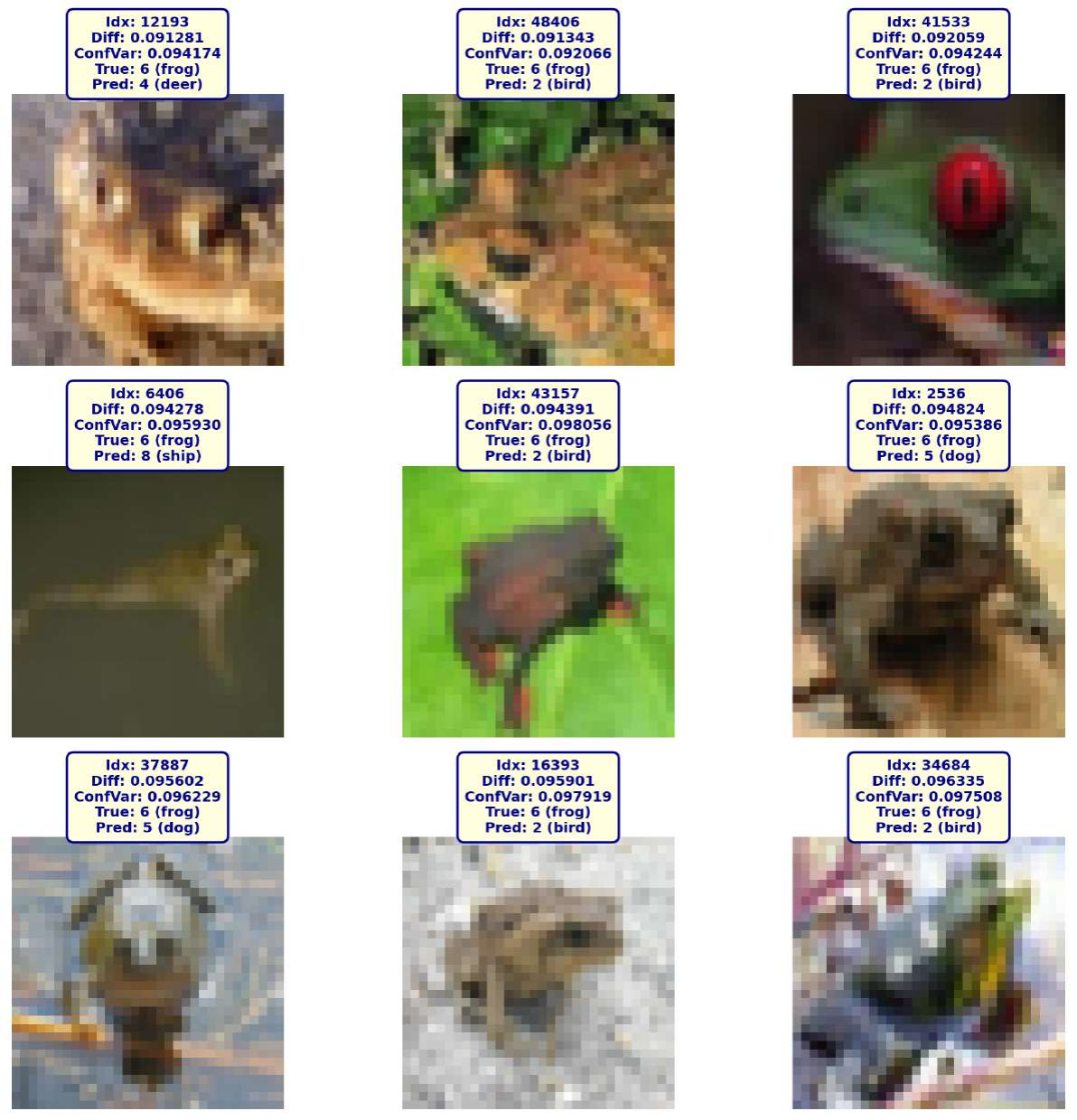}
    \caption{Most difficult \textit{frog} samples.}
  \end{subfigure}
  \caption{Frog class.}
  \label{fig:viz_frog}
\end{figure*}

\begin{figure*}[!t]
  \centering
  \begin{subfigure}[]{0.48\textwidth}
    \includegraphics[width=\linewidth]{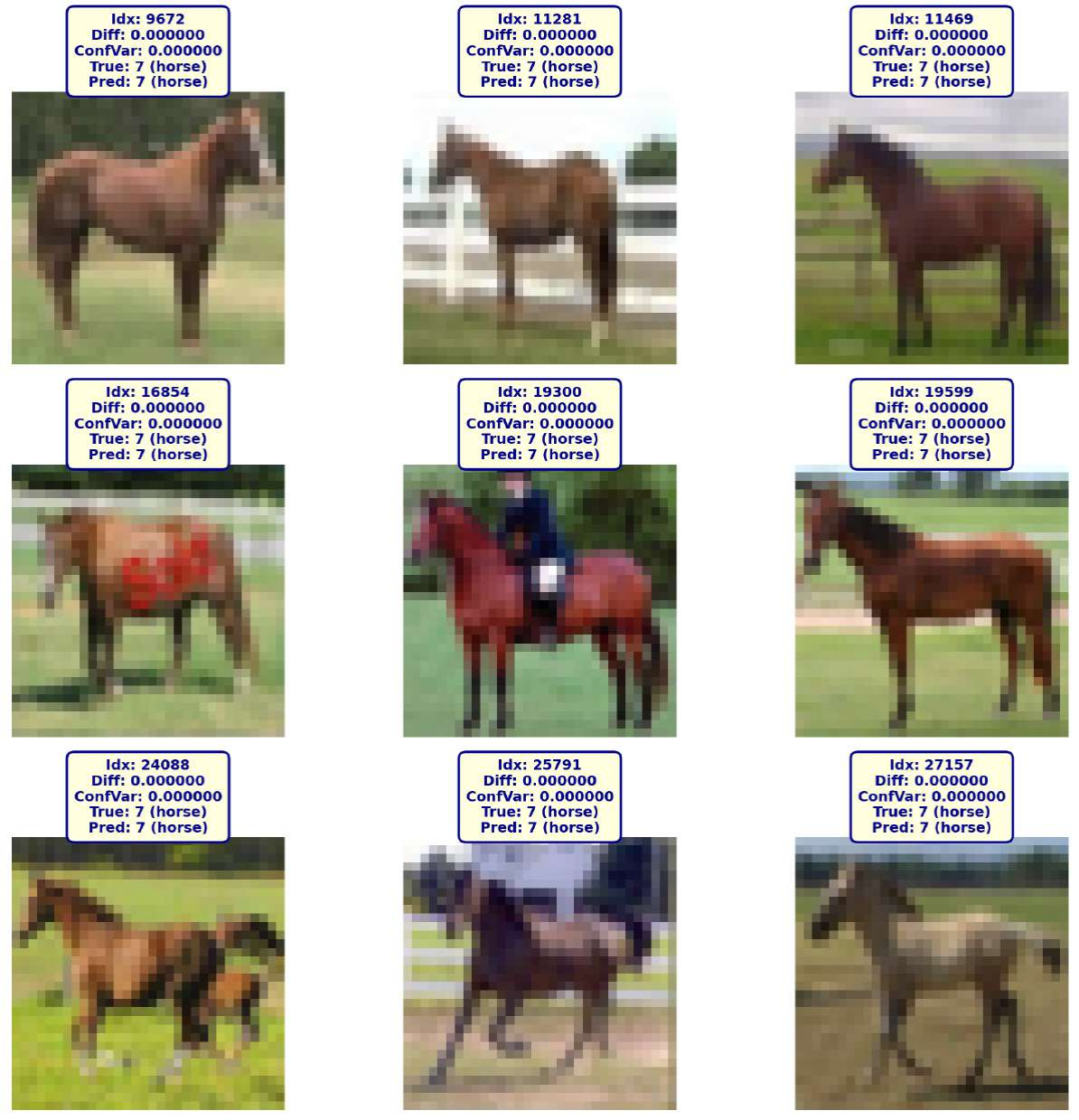}
    \caption{Easiest \textit{horse} samples.}
  \end{subfigure}
  \hfill
  \begin{subfigure}[]{0.48\textwidth}
    \includegraphics[width=\linewidth]{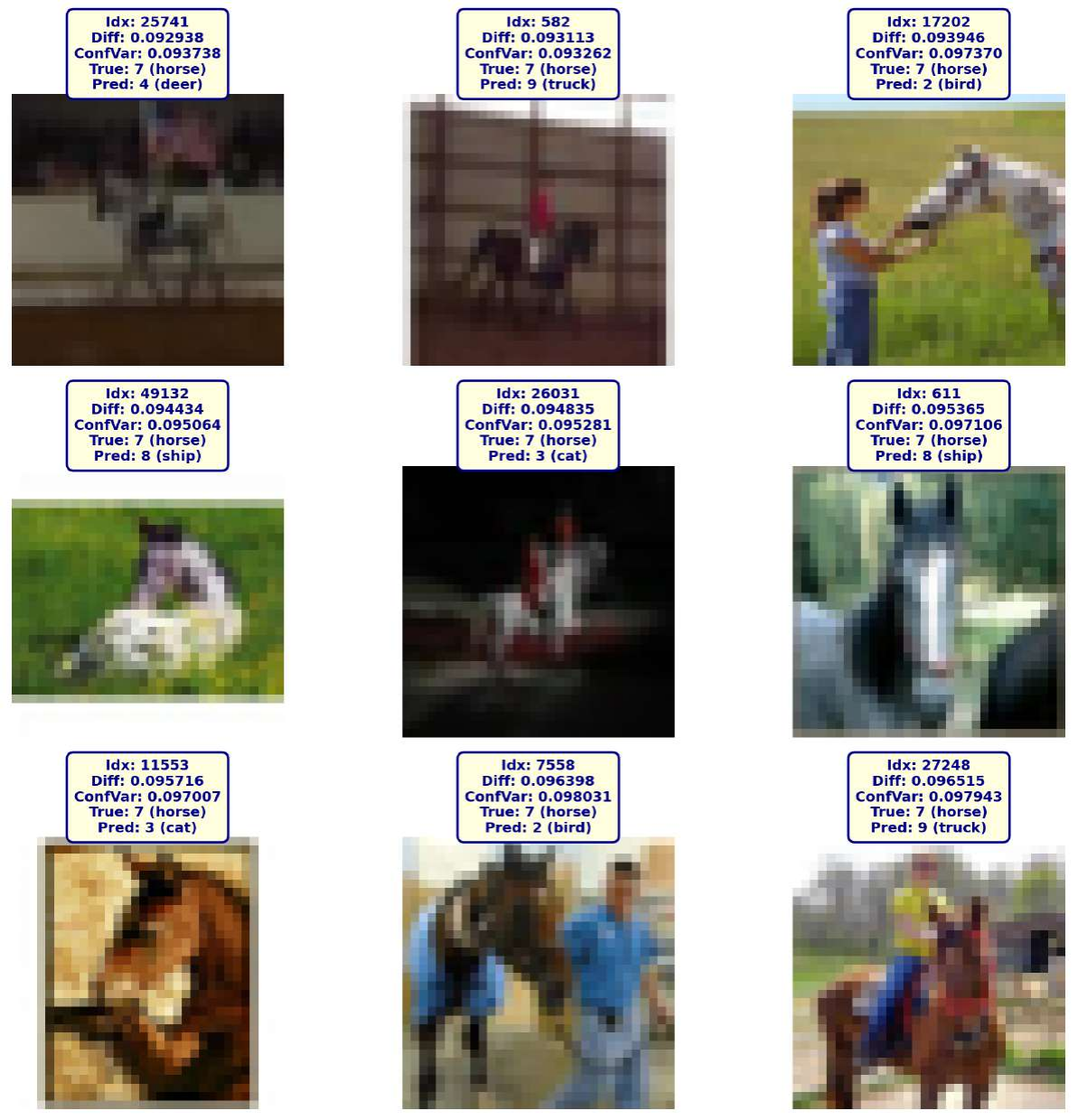}
    \caption{Most difficult \textit{horse} samples.}
  \end{subfigure}
  \caption{Horse class.}
  \label{fig:viz_horse}
\end{figure*}

\begin{figure*}[!t]
  \centering
  \begin{subfigure}[]{0.48\textwidth}
    \includegraphics[width=\linewidth]{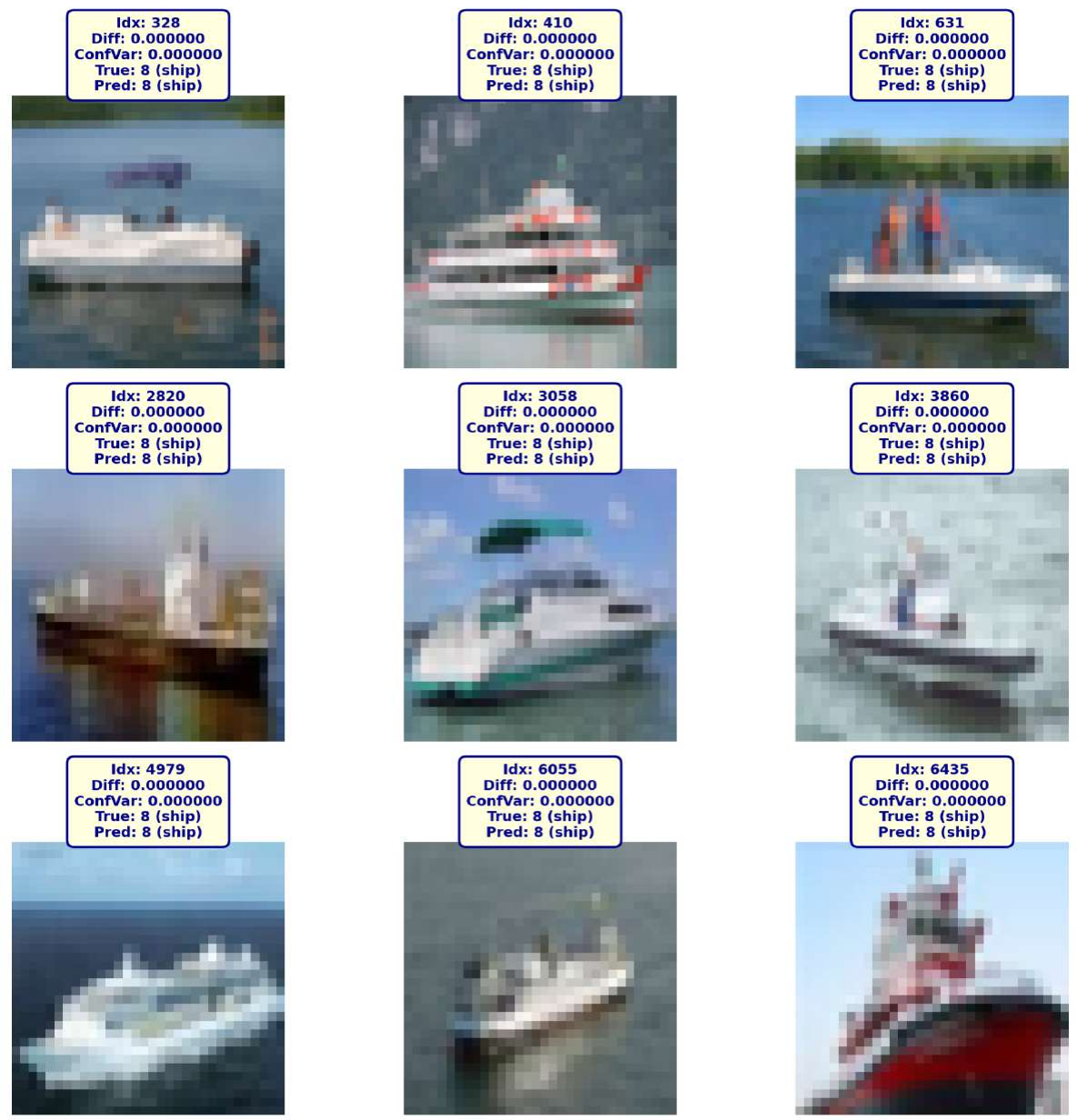}
    \caption{Easiest \textit{ship} samples.}
  \end{subfigure}
  \hfill
  \begin{subfigure}[]{0.48\textwidth}
    \includegraphics[width=\linewidth]{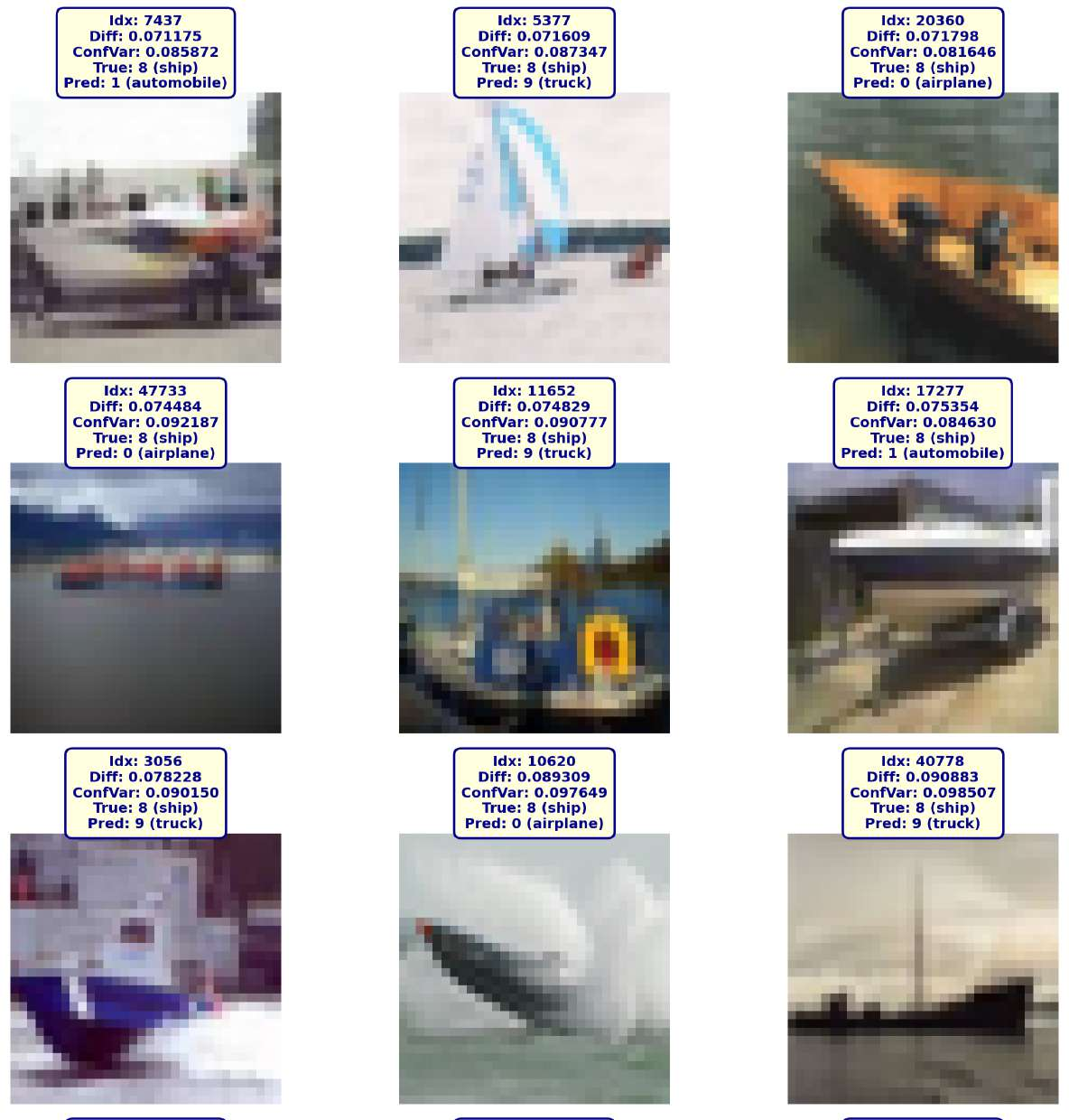}
    \caption{Most difficult \textit{ship} samples.}
  \end{subfigure}
  \caption{Ship class.}
  \label{fig:viz_ship}
\end{figure*}

\begin{figure*}[!t]
  \centering
  \begin{subfigure}[]{0.48\textwidth}
    \includegraphics[width=\linewidth]{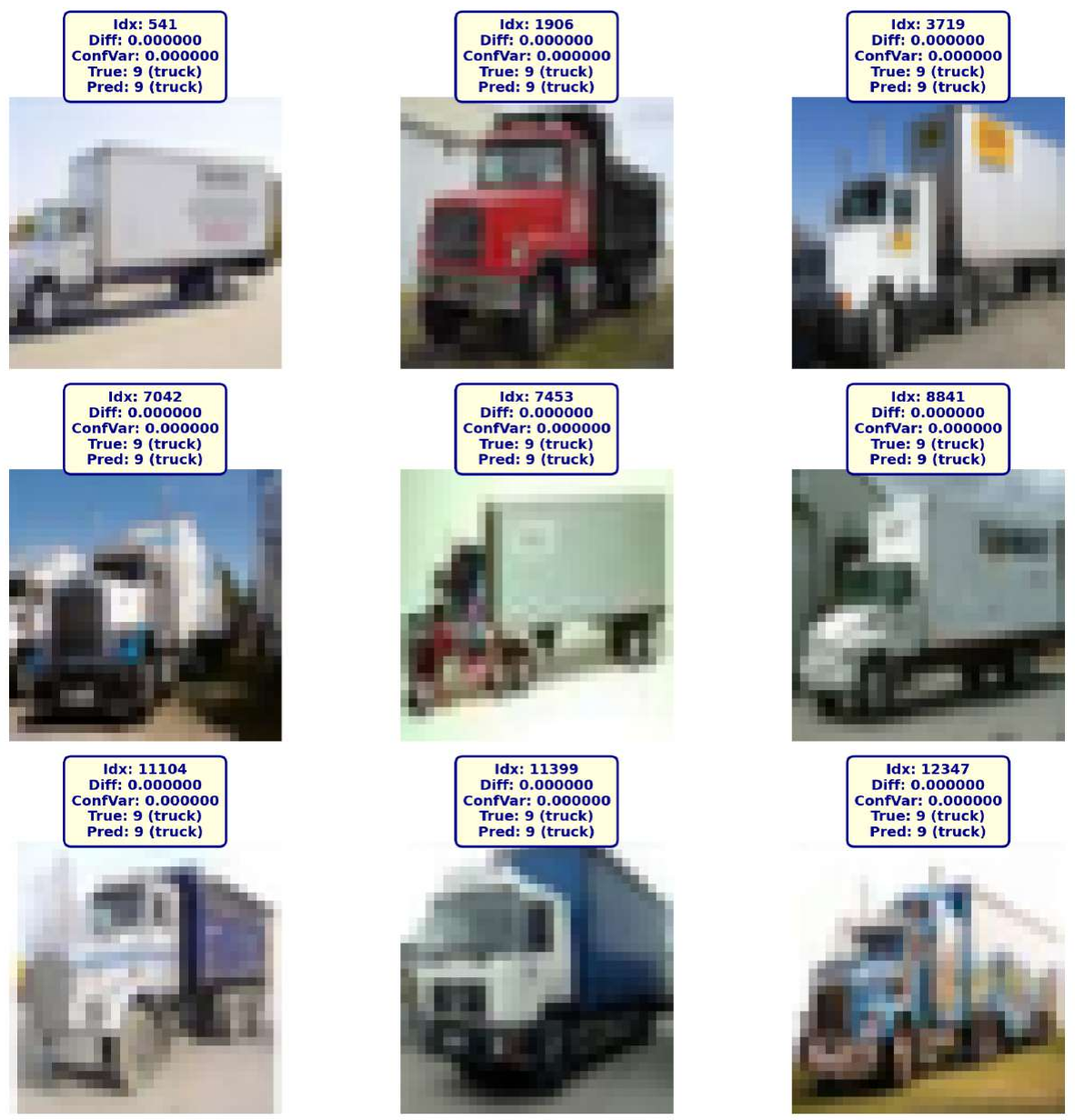}
    \caption{Easiest \textit{truck} samples.}
  \end{subfigure}
  \hfill
  \begin{subfigure}[]{0.48\textwidth}
    \includegraphics[width=\linewidth]{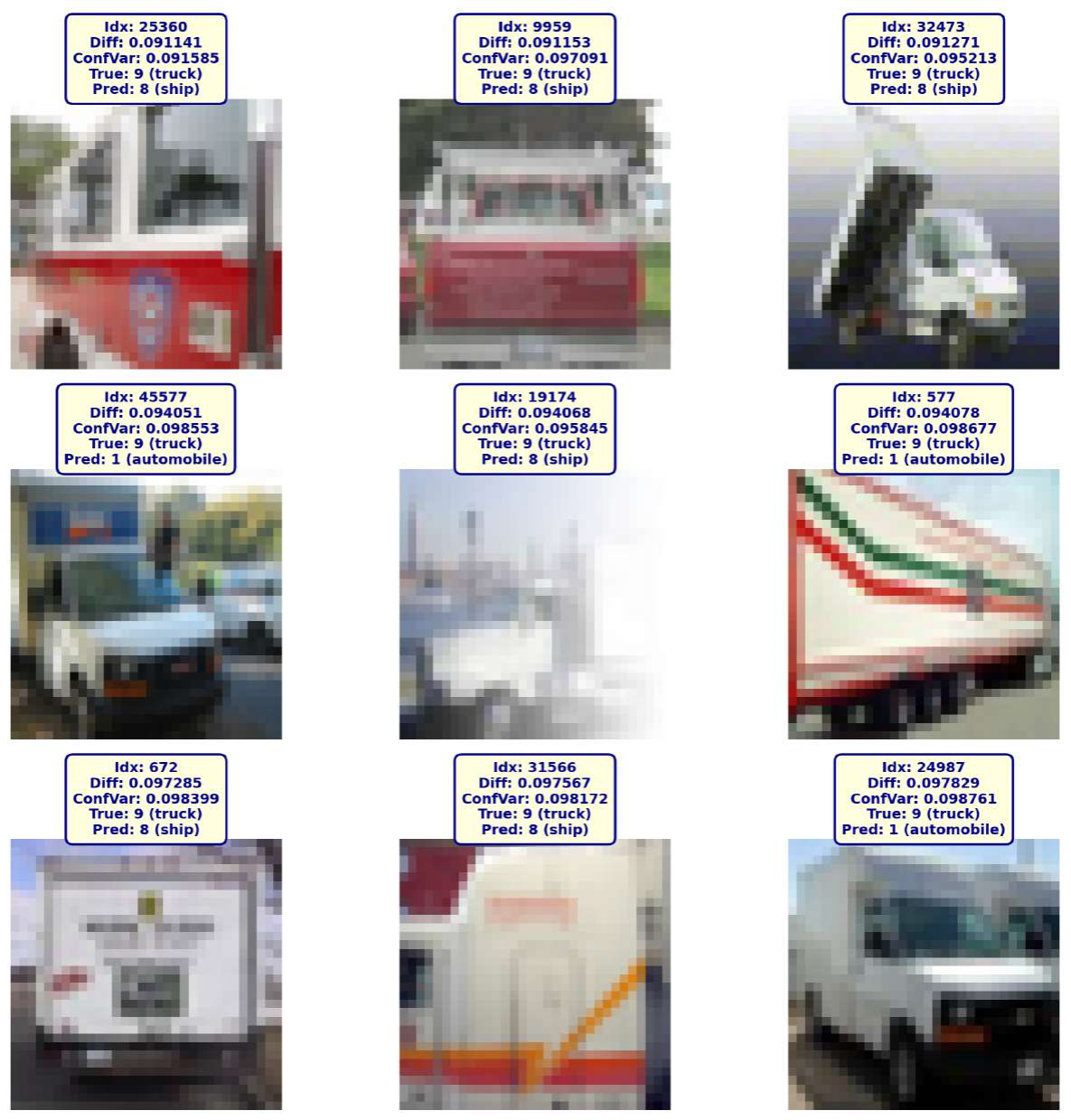}
    \caption{Most difficult \textit{truck} samples.}
  \end{subfigure}
  \caption{Truck class.}
  \label{fig:viz_truck}
\end{figure*}

Figures~\ref{fig:viz_airplane}--\ref{fig:viz_truck} show nine randomly 
sampled images from the 20 easiest and the 9 most difficult samples for 
each of the ten CIFAR-10 classes, ranked by $D(x)$. Difficulty scores on 
easy samples display as \texttt{0.000000} due to plot decimal precision; 
values are non-zero. The \texttt{Idx} field denotes the default sample index in 
the CIFAR-10 training set.

\subsection{Difficulty Score Validation}
\label{app:diff_validation}
\begin{figure}[!t]
    \centering
    \begin{subfigure}[t]{0.48\linewidth}
        \centering
        \includegraphics[width=\linewidth, height=6cm, keepaspectratio]{1c.pdf}
        \caption{100\% teacher --- valid ordering.}
        \label{fig:stage_100_appendix}
    \end{subfigure}
    \hfill
    \begin{subfigure}[t]{0.48\linewidth}
        \centering
        \includegraphics[width=\linewidth, height=6cm, keepaspectratio]{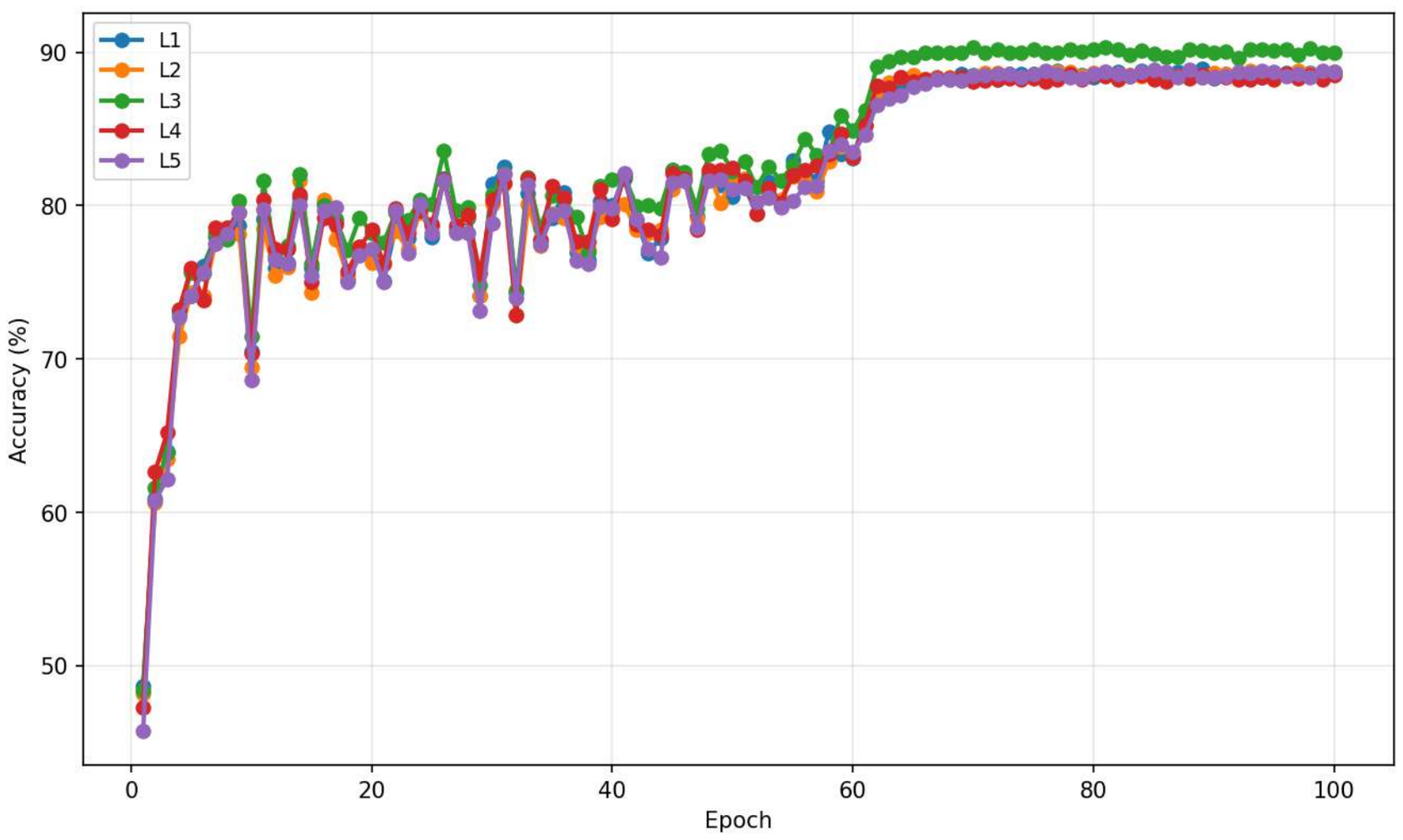}
        \caption{10\% teacher --- invalid ordering.}
        \label{fig:stage_10_appendix}
    \end{subfigure}
    \caption{Stage-accuracy curves across 100 training epochs on fixed difficulty bins L1--L5. \textbf{(a)} The 100\% teacher produces clear monotone separation. \textbf{(b)} The 10\% teacher produces overlapping curves, confirming teacher expressiveness is necessary for a valid difficulty ordering.}
    \label{fig:stage_gradient_appendix}
\end{figure}

Figure~\ref{fig:stage_100_appendix} and \ref{fig:stage_10_appendix} show the stage-wise test accuracy on non-CL baseline where the test subsets are obtained from a 100\% teacher model and a 10\% teacher model respectively. The 100\% teacher produces a clear monotone
gradient that holds throughout training ( L1 64.00\% → L2
52.25\% → L3 44.45\% → L4 35.50\% → L5 29.60\%) and 10\% teacher produces a near-flat
gradient ( L1 44.75\% → L2 42.25\% → L3 42.95\% → L4
42.65\% → L5 43.40\%, range 42.25\%–44.75\%).

% \onecolumn
% \clearpage
\FloatBarrier
\clearpage

\section{Performance Evaluation Plots}

\renewcommand{\thefigure}{B\arabic{figure}}
\setcounter{figure}{0}
\renewcommand{\thetable}{B\arabic{table}}
\setcounter{table}{0}
\label{app:AppendixB}

\begin{figure}[h]
    \centering
    \begin{subfigure}[]{0.48\textwidth}
        \centering
        \includegraphics[width=\textwidth]{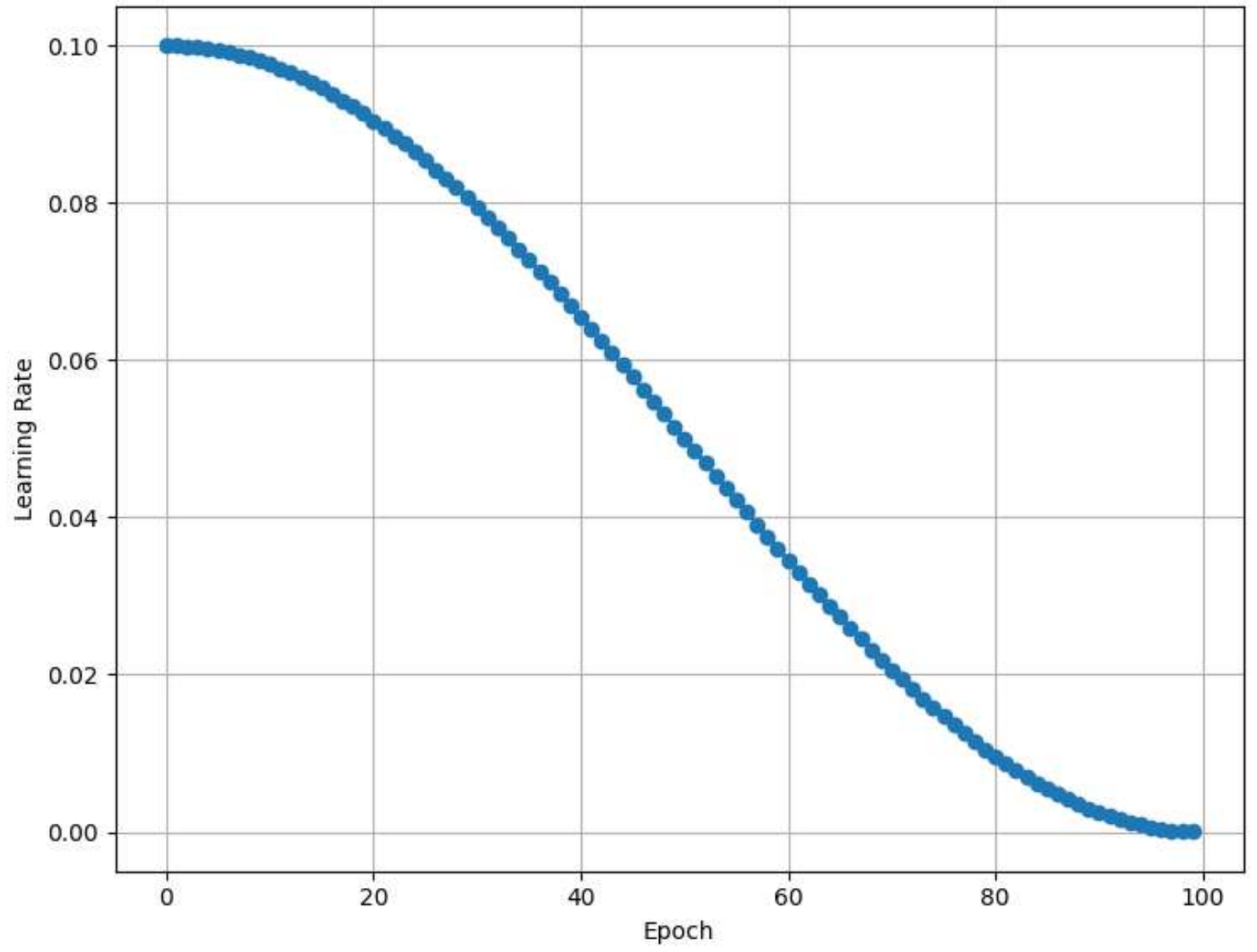}
        \caption{Cosine annealing.}
        \label{fig:lr_cosine}
    \end{subfigure}
    \hfill
    \begin{subfigure}[]{0.48\textwidth}
        \centering
        \includegraphics[width=\textwidth]{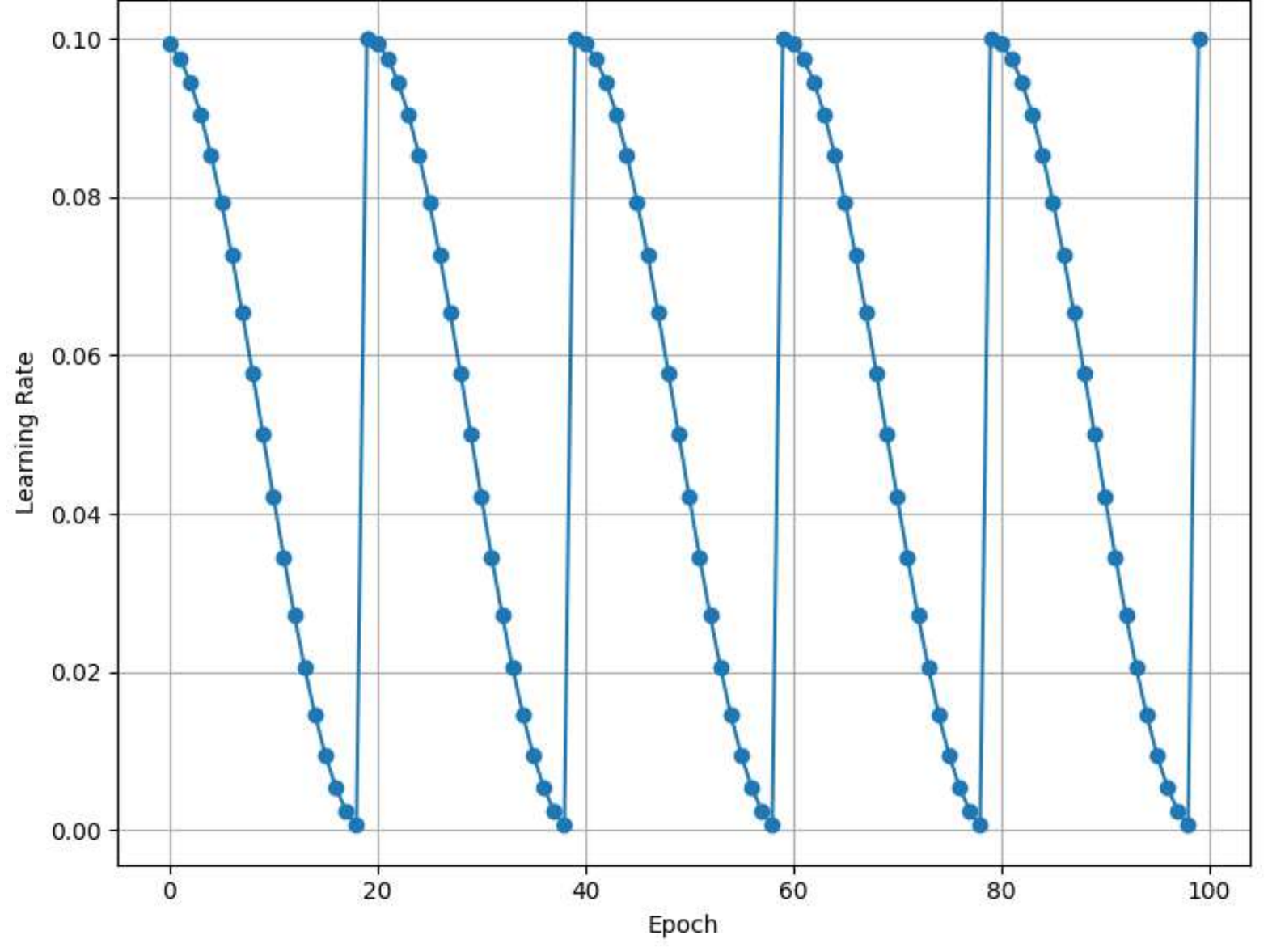}
        \caption{Stage-cosine annealing.}
        \label{fig:lr_stage_cosine}
    \end{subfigure}
    \caption{The two learning rate schedules used across all experiments, 
    both initialised at $\eta_0 = 0.1$. 
    (a) Standard cosine annealing decays monotonically to 0 over 100 epochs. 
    (b) Stage-cosine annealing restarts the cosine decay every 20 epochs, 
    producing five identical warm-restart cycles. 
    The stage boundaries coincide with the curriculum pacing stages, 
    so each time the training set expands the learning rate resets to 
    $\eta_0$, allowing the model to re-adapt to the newly introduced samples.}
    \label{fig:lr_schedules}
\end{figure}

\begin{figure*}[!t]
    \centering
    \begin{subfigure}[]{0.32\textwidth}
        \centering
        \includegraphics[width=\textwidth]{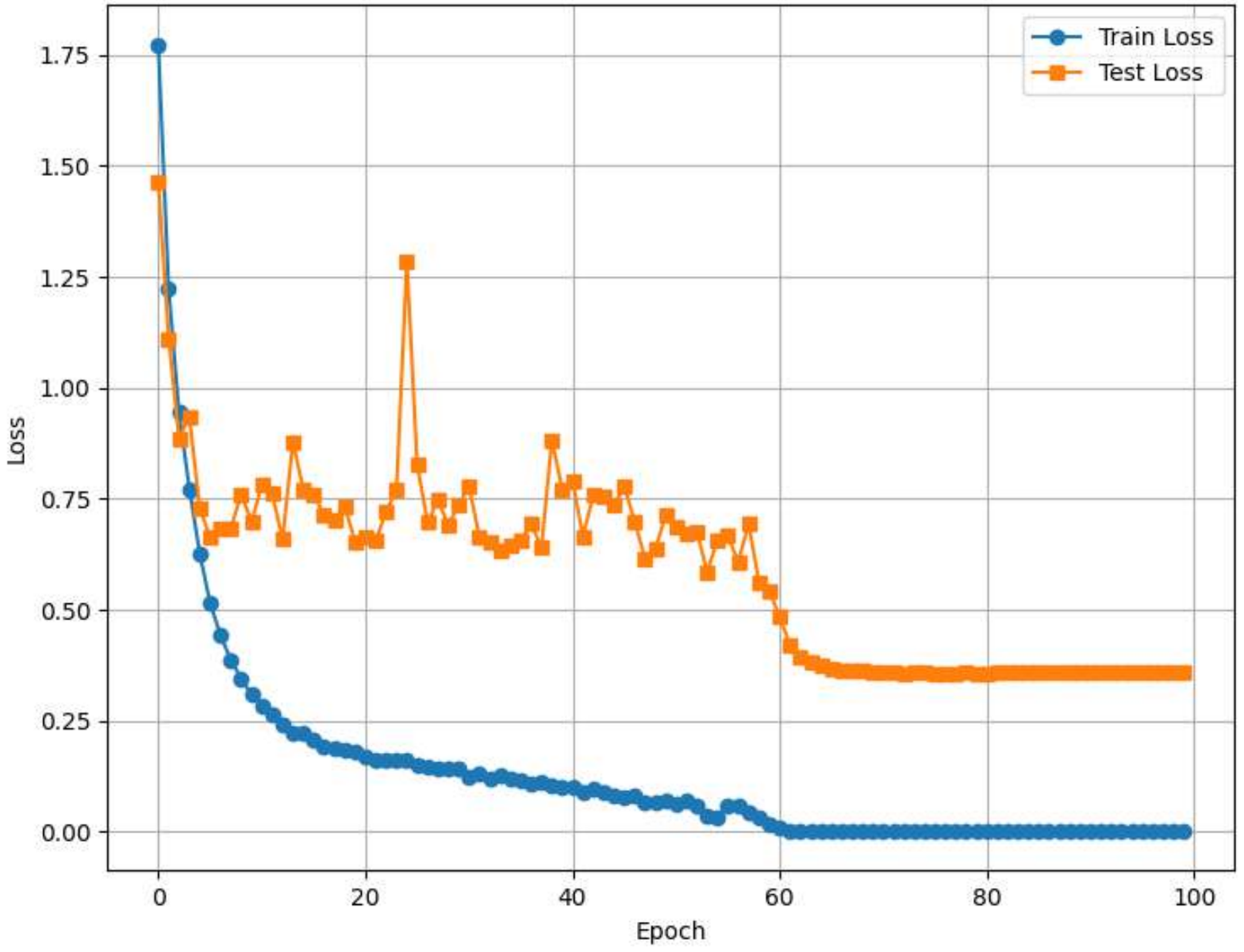}
        \caption{Training and test loss.}
        \label{fig:resnet18_baseline_loss}
    \end{subfigure}
    \hfill
    \begin{subfigure}[]{0.32\textwidth}
        \centering
        \includegraphics[width=\textwidth]{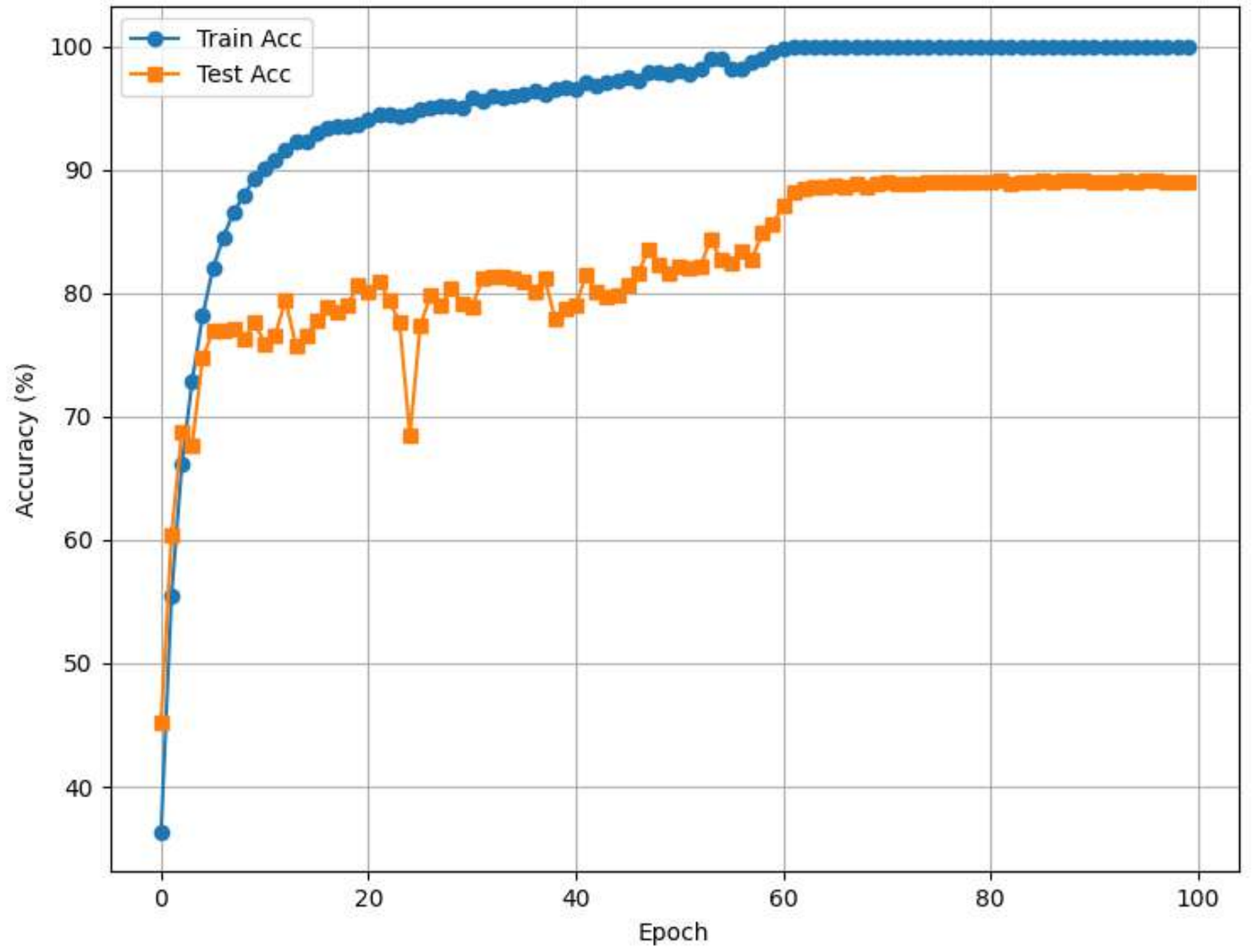}
        \caption{Training and test accuracy.}
        \label{fig:resnet18_baseline_acc}
    \end{subfigure}
    \hfill
    \begin{subfigure}[]{0.32\textwidth}
        \centering
        \includegraphics[width=\textwidth]{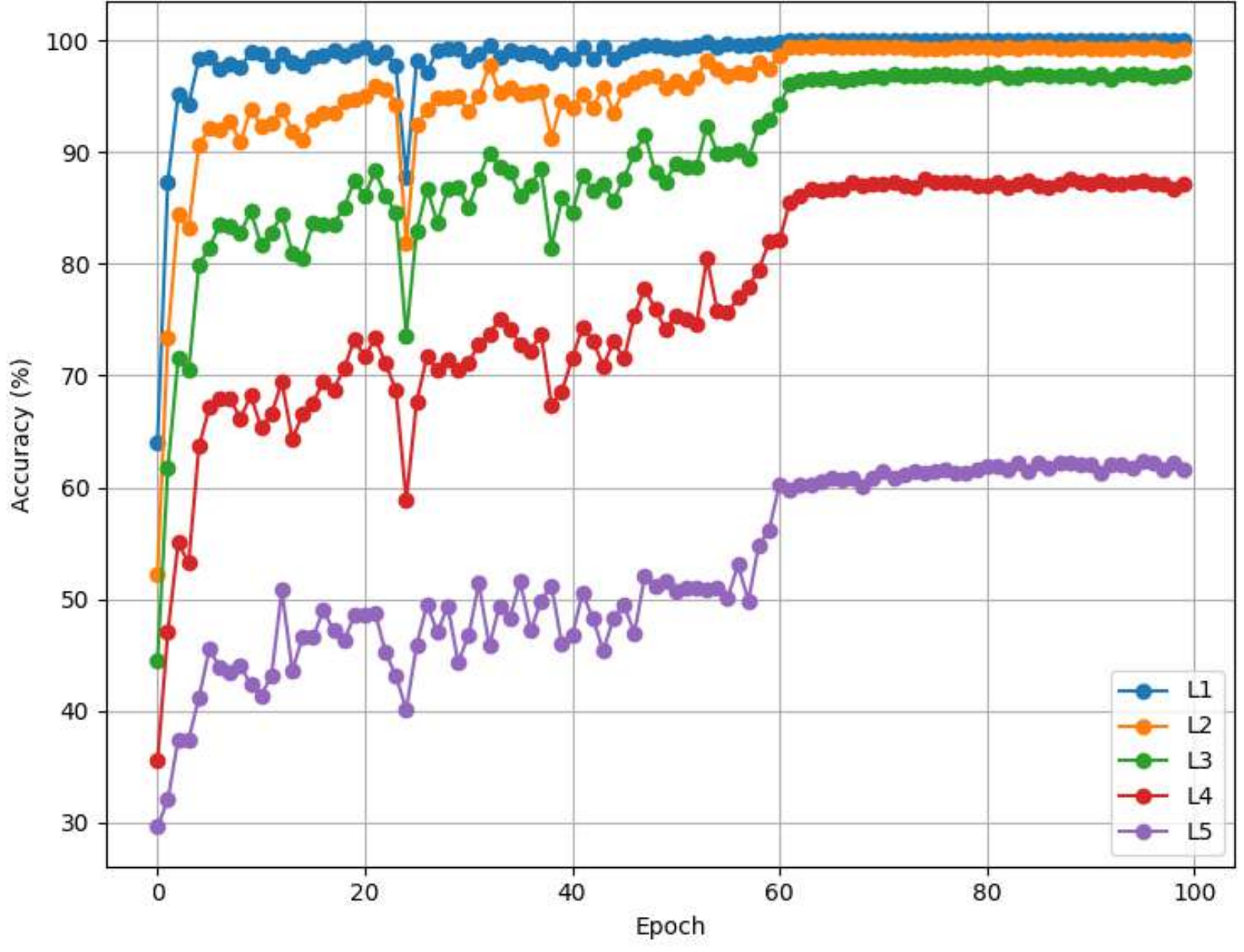}
        \caption{Stage-wise test accuracy.}
        \label{fig:resnet18_baseline_stages}
    \end{subfigure}
    
   \caption{Training curves for \textbf{ResNet-18 Baseline, Cosine LR}.}
    \label{fig:appendix_resnet18_baseline}
\end{figure*}

\begin{figure*}[h]
    \centering
    \begin{subfigure}[]{0.32\textwidth}
        \centering
        \includegraphics[width=\textwidth]{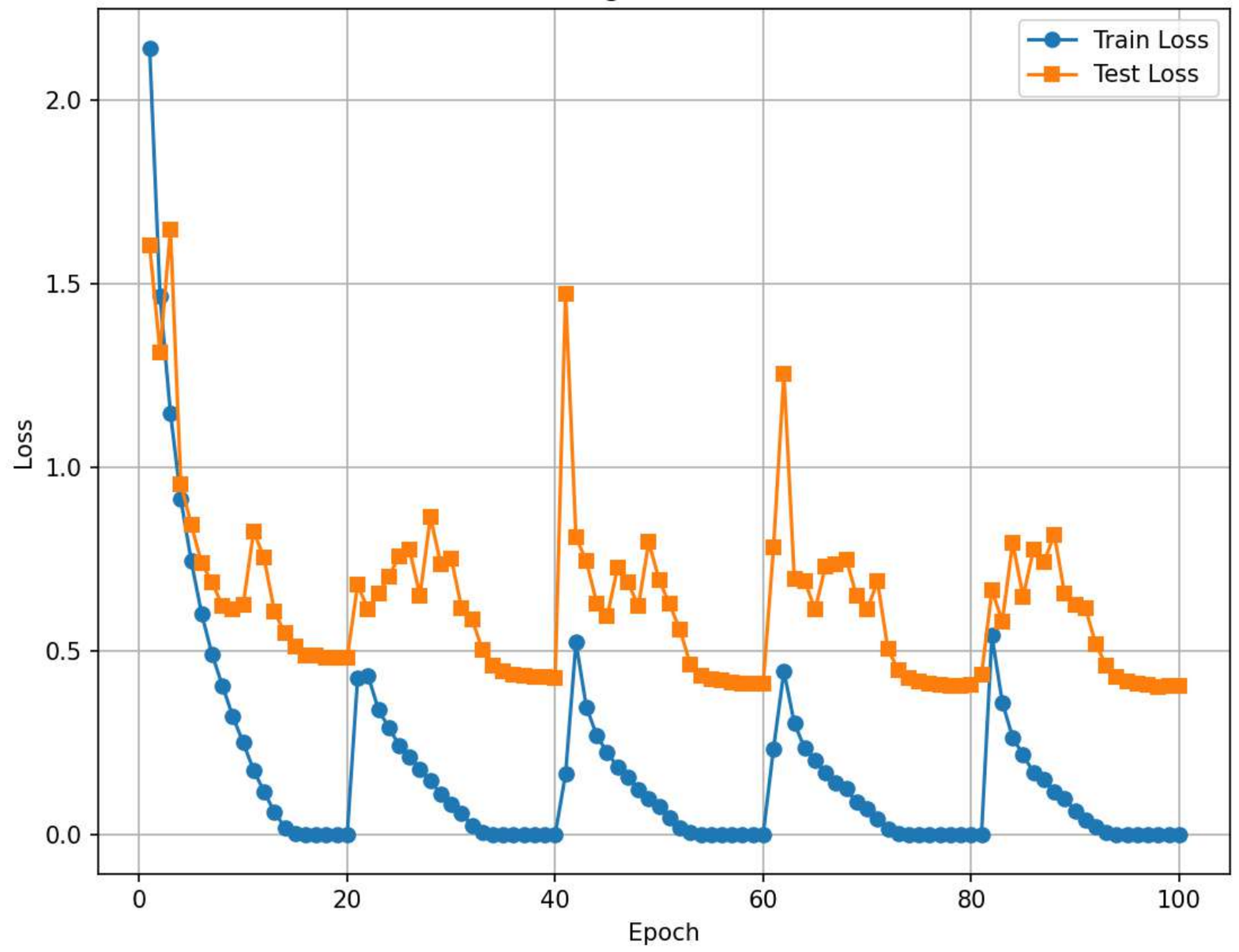}
        \caption{Training and test loss.}
        \label{fig:resnet18_baseline_stage_loss}
    \end{subfigure}
    \hfill
    \begin{subfigure}[]{0.32\textwidth}
        \centering
        \includegraphics[width=\textwidth]{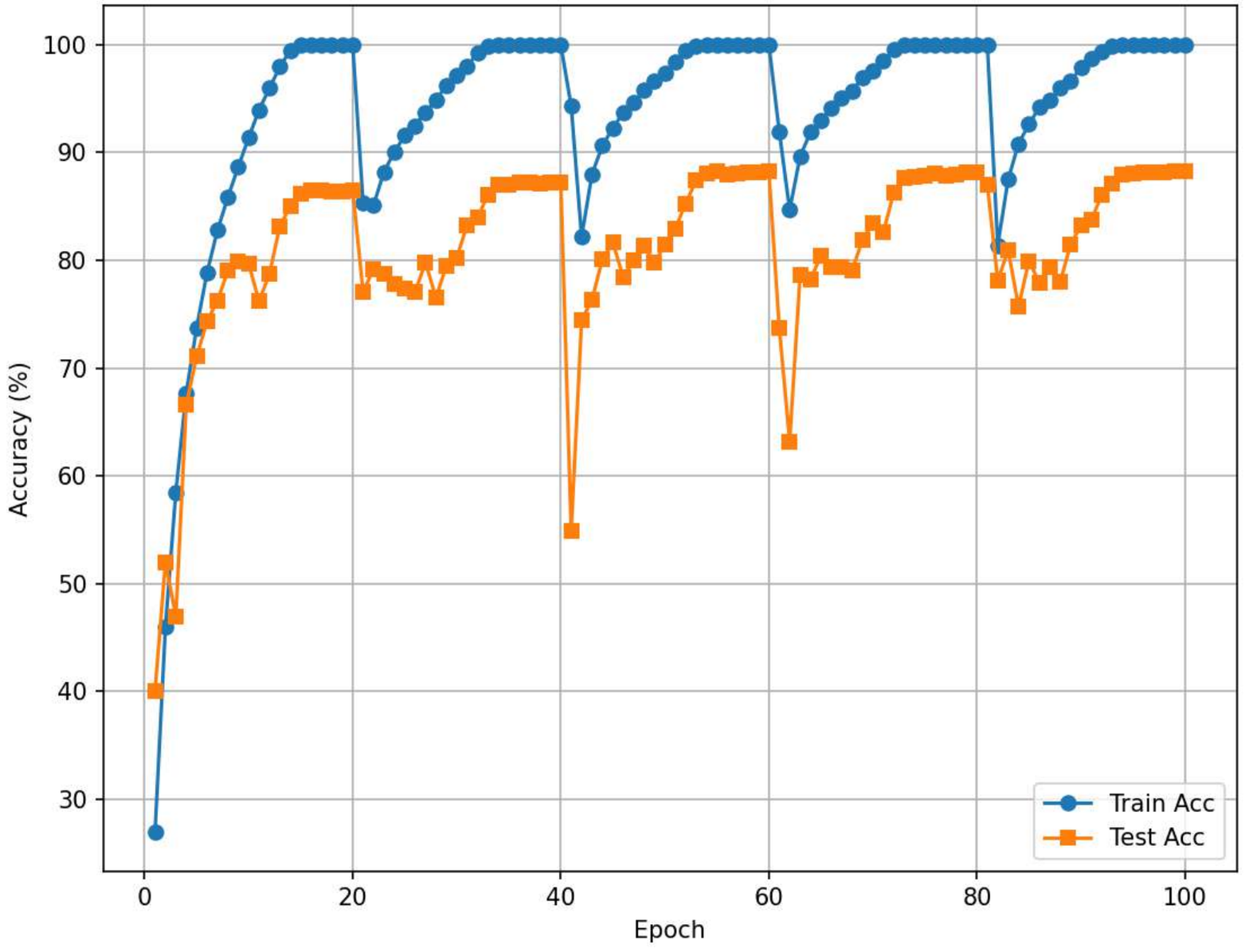}
        \caption{Training and test accuracy.}
        \label{fig:resnet18_baseline_stage_acc}
    \end{subfigure}
    \hfill
    \begin{subfigure}[]{0.32\textwidth}
        \centering
        \includegraphics[width=\textwidth]{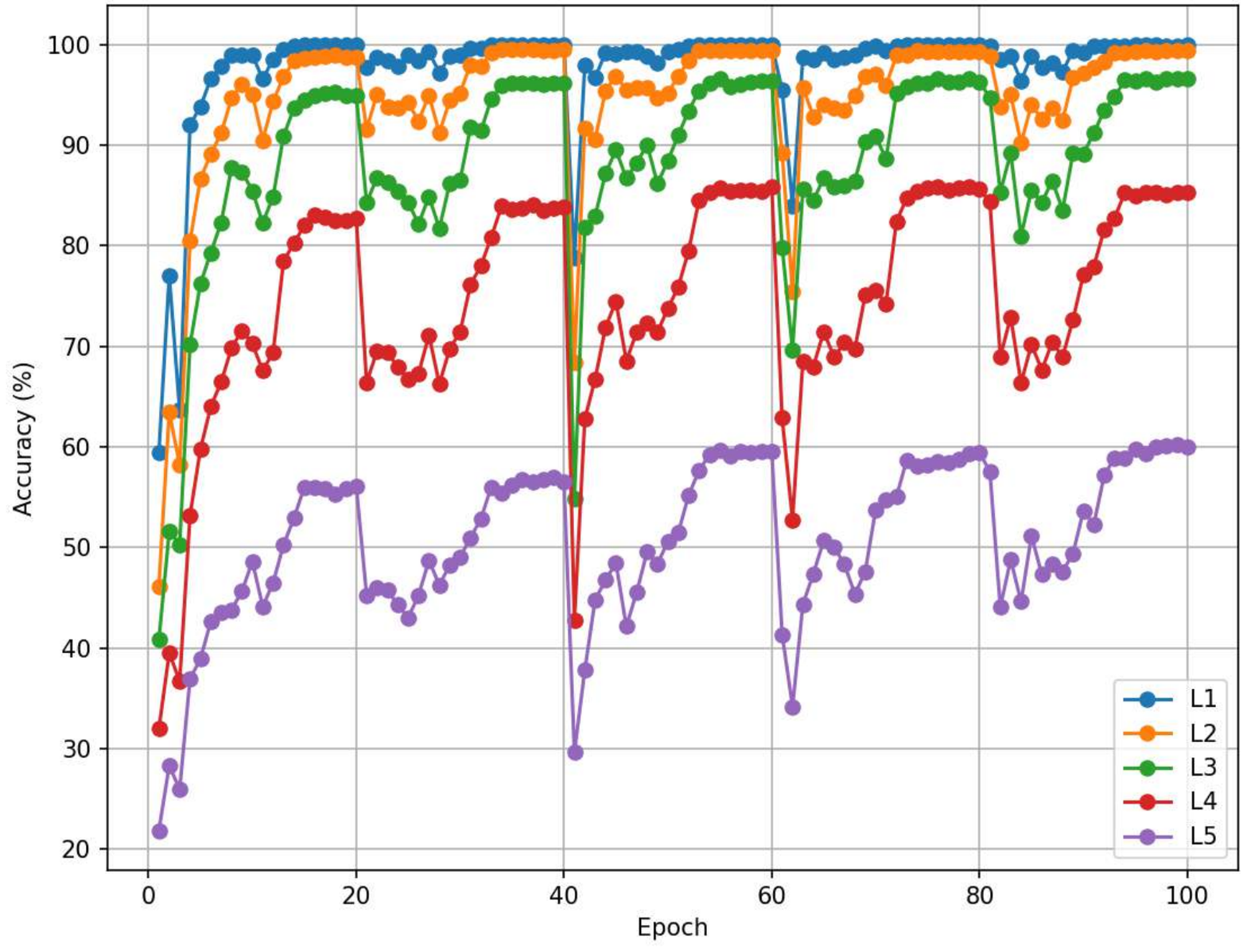}
        \caption{Stage-wise test accuracy.}
        \label{fig:resnet18_baseline_stage_stages}
    \end{subfigure}
    
    \caption{Training curves for \textbf{ResNet-18 Baseline, Stage-Cosine LR}.}
\label{fig:appendix_resnet18_baseline_stage}

\end{figure*}

\begin{figure*}[ht]
    \centering
    \begin{subfigure}[]{0.32\textwidth}
        \centering
        \includegraphics[width=\textwidth]{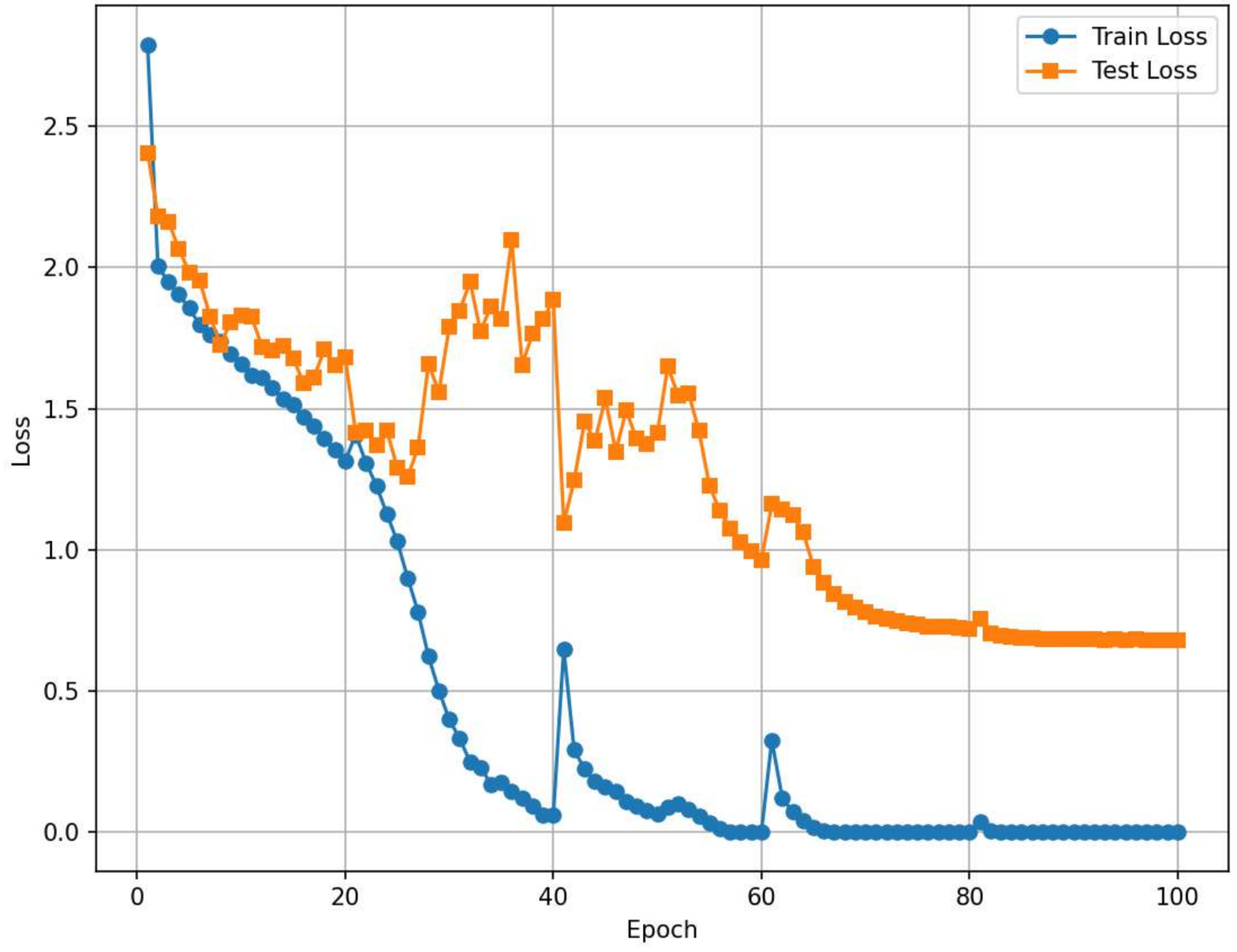}
        \caption{Training and test loss.}
        \label{fig:resnet18_anticoisne_loss}
    \end{subfigure}
    \hfill
    \begin{subfigure}[]{0.32\textwidth}
        \centering
        \includegraphics[width=\textwidth]{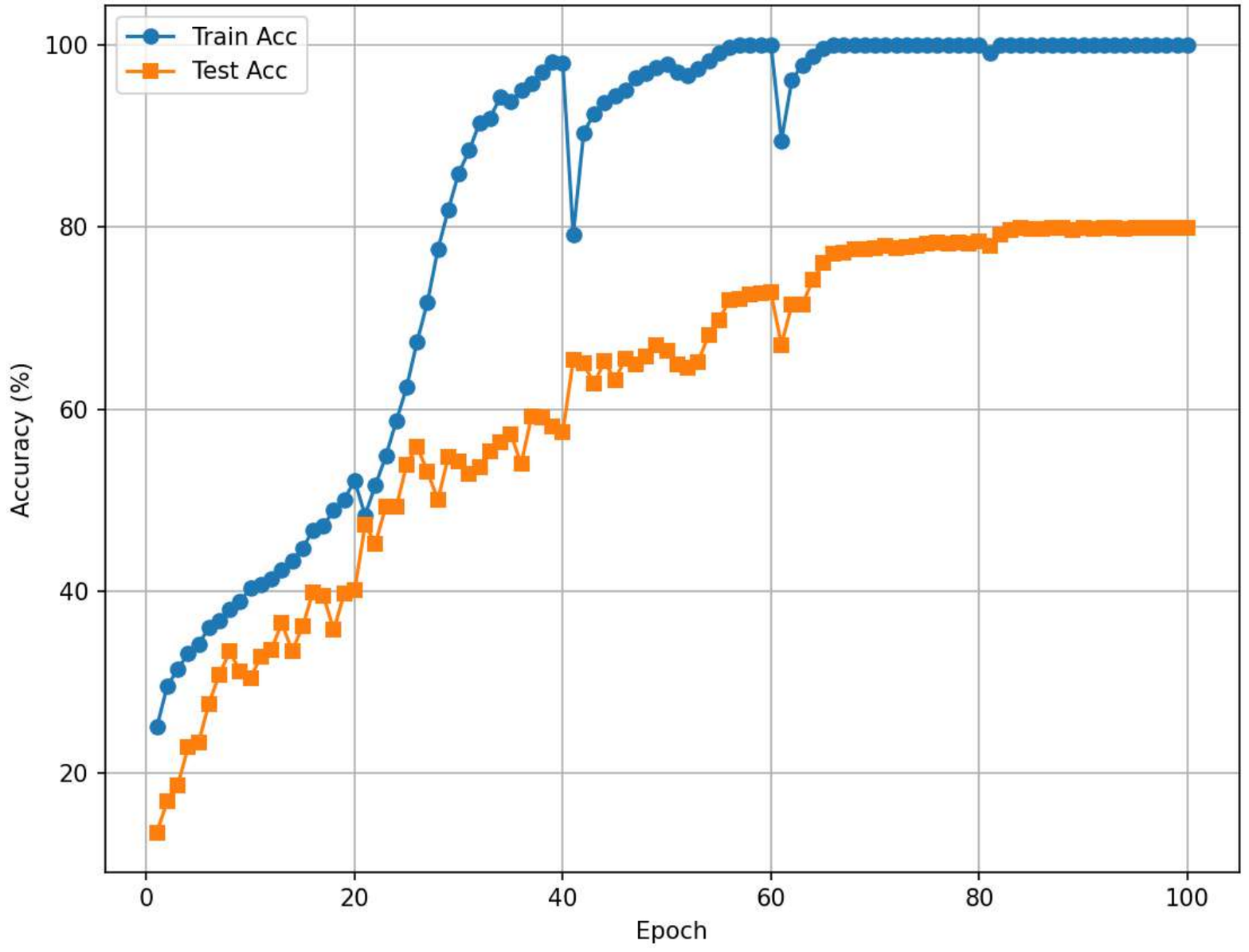}
        \caption{Training and test accuracy.}
        \label{fig:resnet18_anticosine_acc}
    \end{subfigure}
    \hfill
    \begin{subfigure}[]{0.32\textwidth}
        \centering
        \includegraphics[width=\textwidth]{
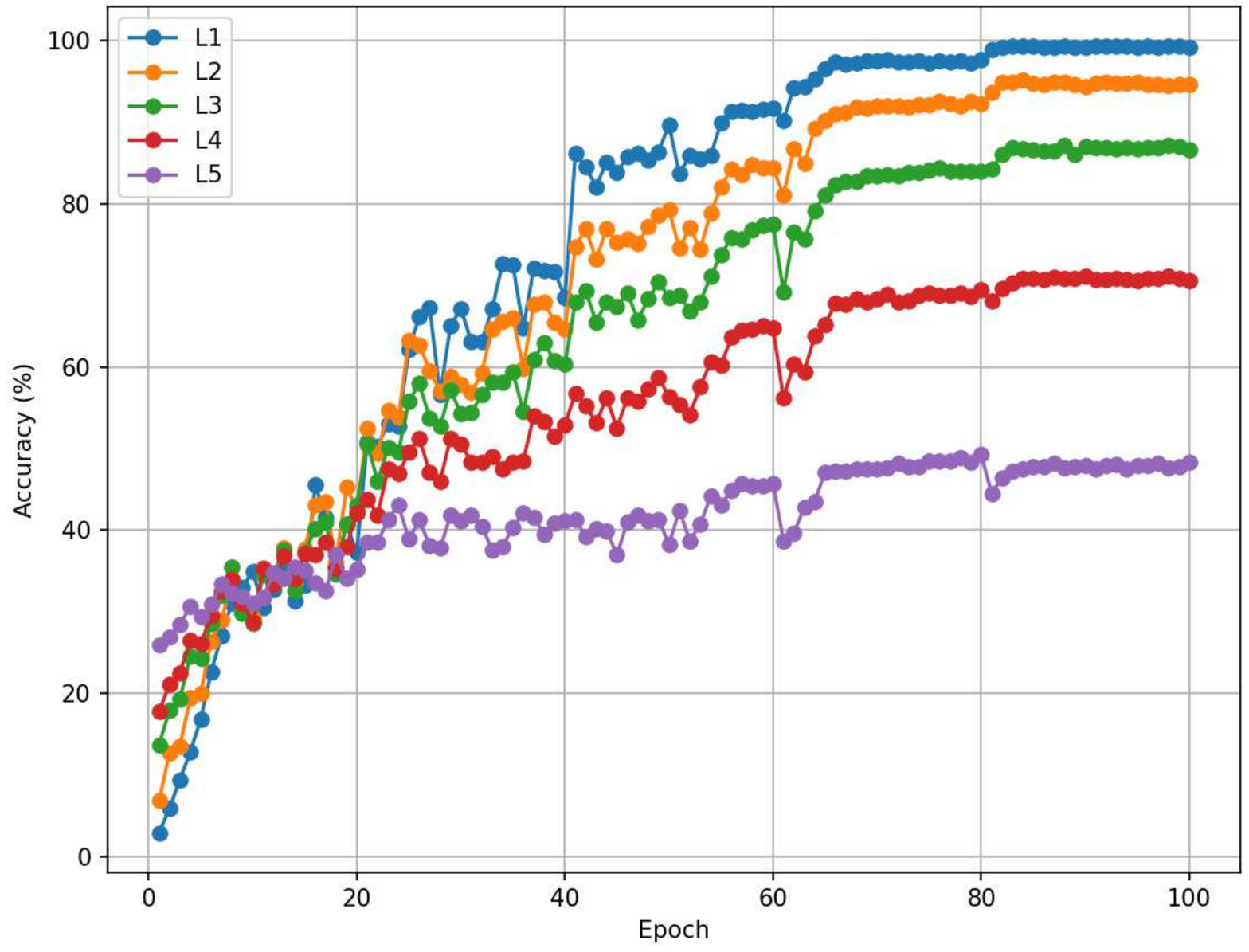}
        \caption{Stage-wise test accuracy.}
        \label{fig:resnet18_anticosoine_stages}
    \end{subfigure}
    
    \caption{Training curves for \textbf{ResNet-18 Anti-Curriculum, Cosine LR}.}
\label{fig:appendix_resnet18_anticosine}
\end{figure*}

\begin{figure*}[h]
    \centering
    \begin{subfigure}[]{0.32\textwidth}
        \centering
        \includegraphics[width=\textwidth]{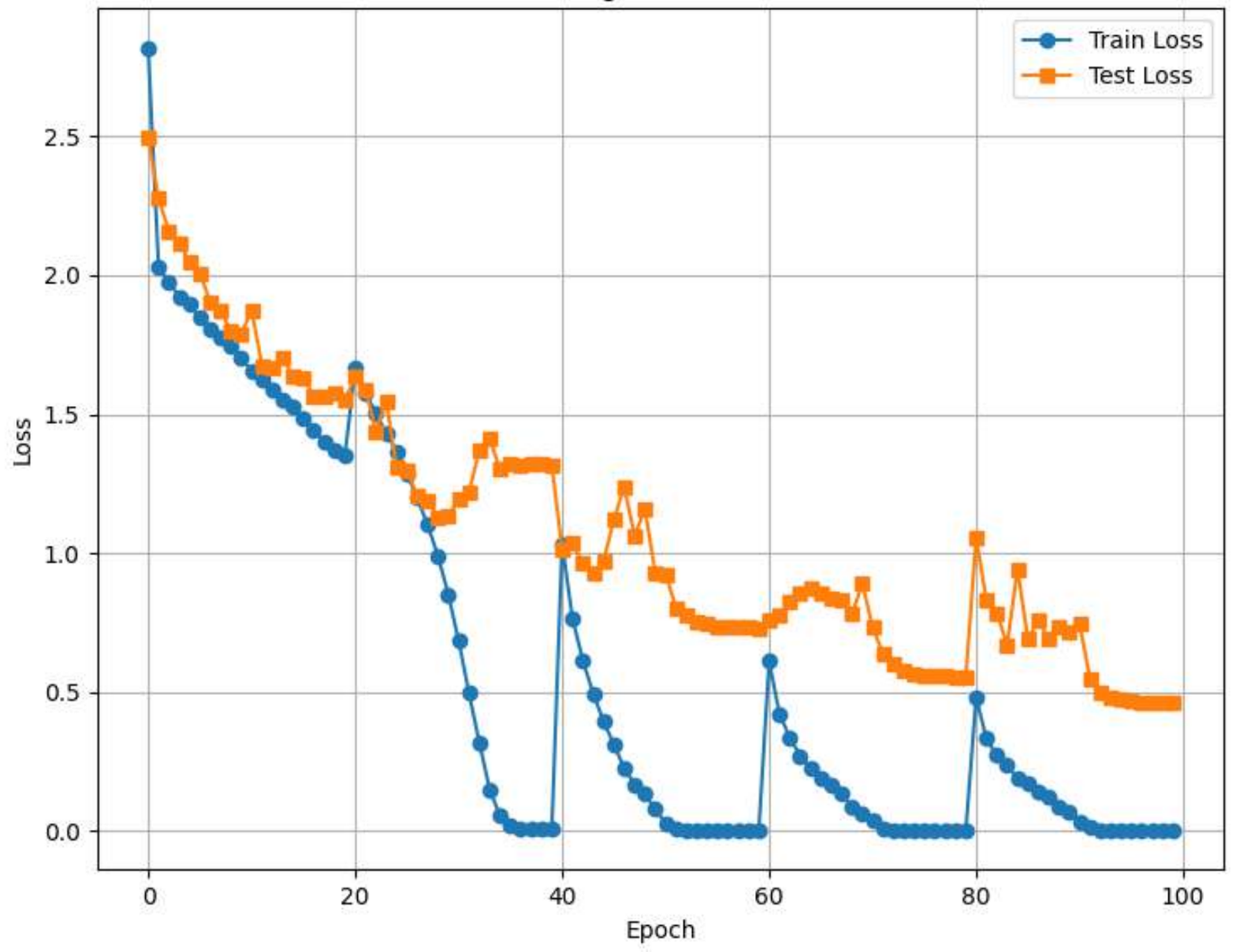}
        \caption{Training and test loss.}
        \label{fig:resnet18_anti_coisne_stage_loss}
    \end{subfigure}
    \hfill
    \begin{subfigure}[]{0.32\textwidth}
        \centering
        \includegraphics[width=\textwidth]{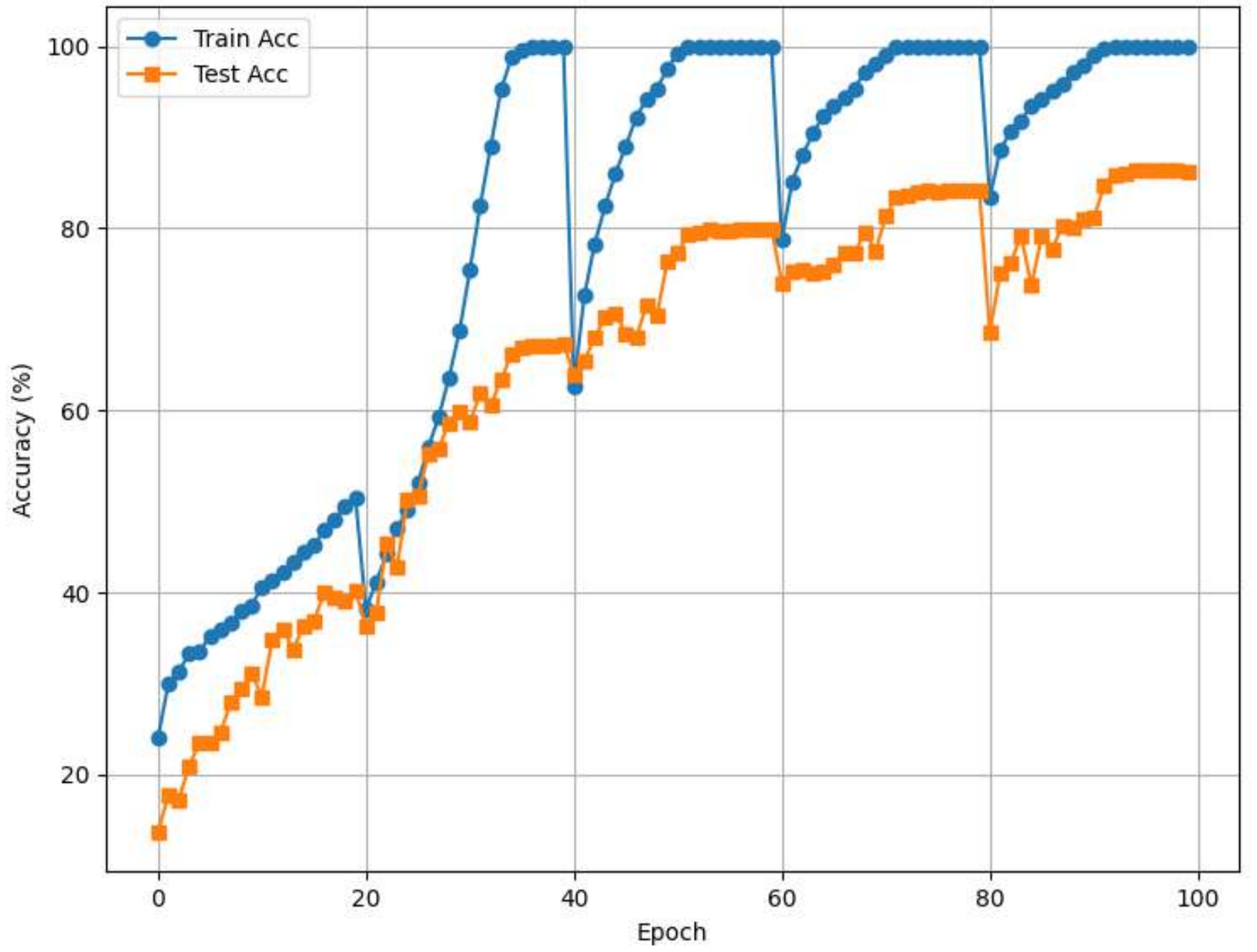}
        \caption{Training and test accuracy.}
        \label{fig:resnet18_anti_cosine_stage_accuracy}
    \end{subfigure}
    \hfill
    \begin{subfigure}[]{0.32\textwidth}
        \centering
        \includegraphics[width=\textwidth]{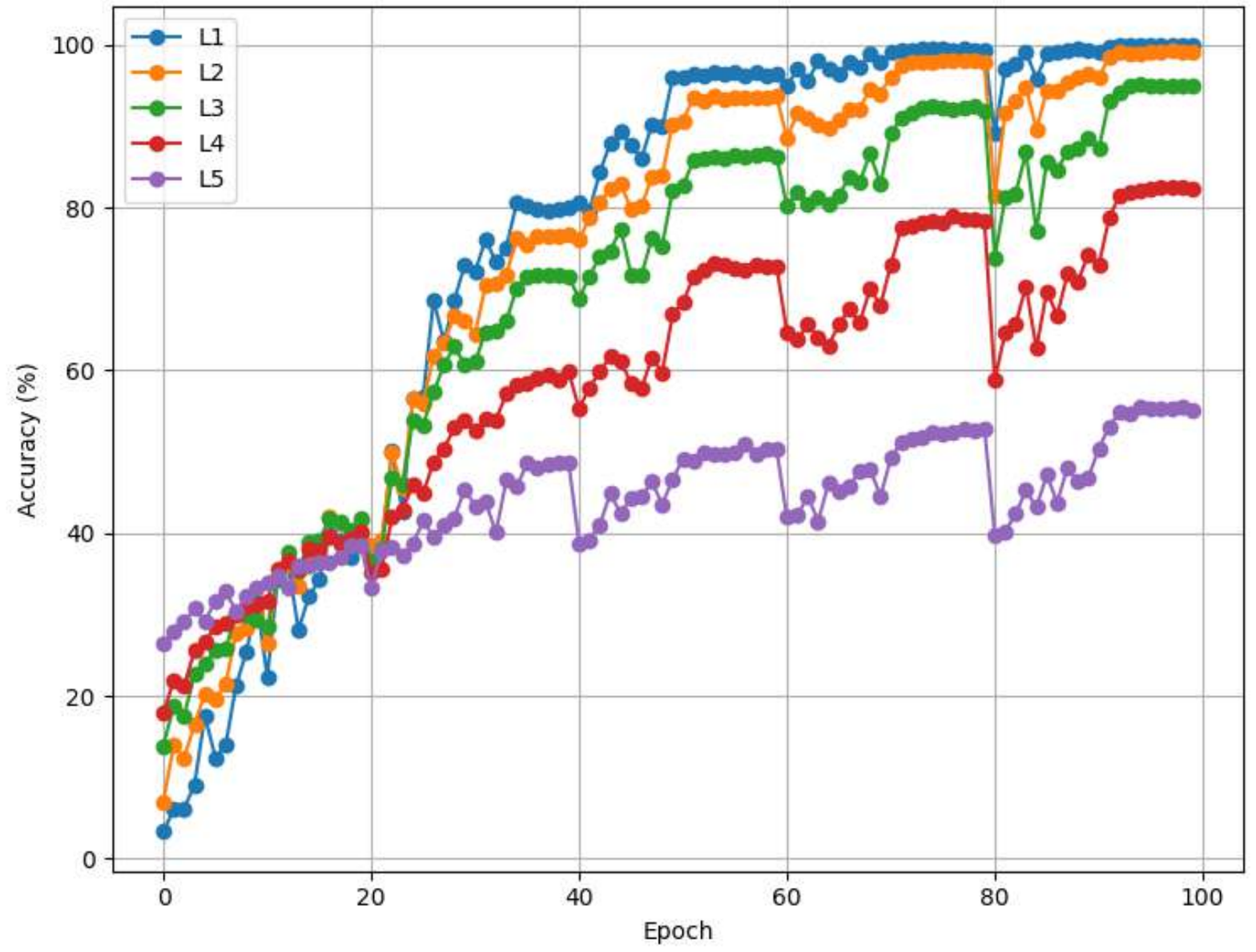}
        \caption{Stage-wise test accuracy.}
        \label{fig:resnet18_anti_cosoine_stages}
    \end{subfigure}
    
   \caption{Training curves for \textbf{VGG-16 Baseline, Cosine LR}.}
\label{fig:appendix_vgg16_baseline}
\end{figure*}

\begin{figure*}[h]
    \centering
    \begin{subfigure}[]{0.32\textwidth}
        \centering
        \includegraphics[width=\textwidth]{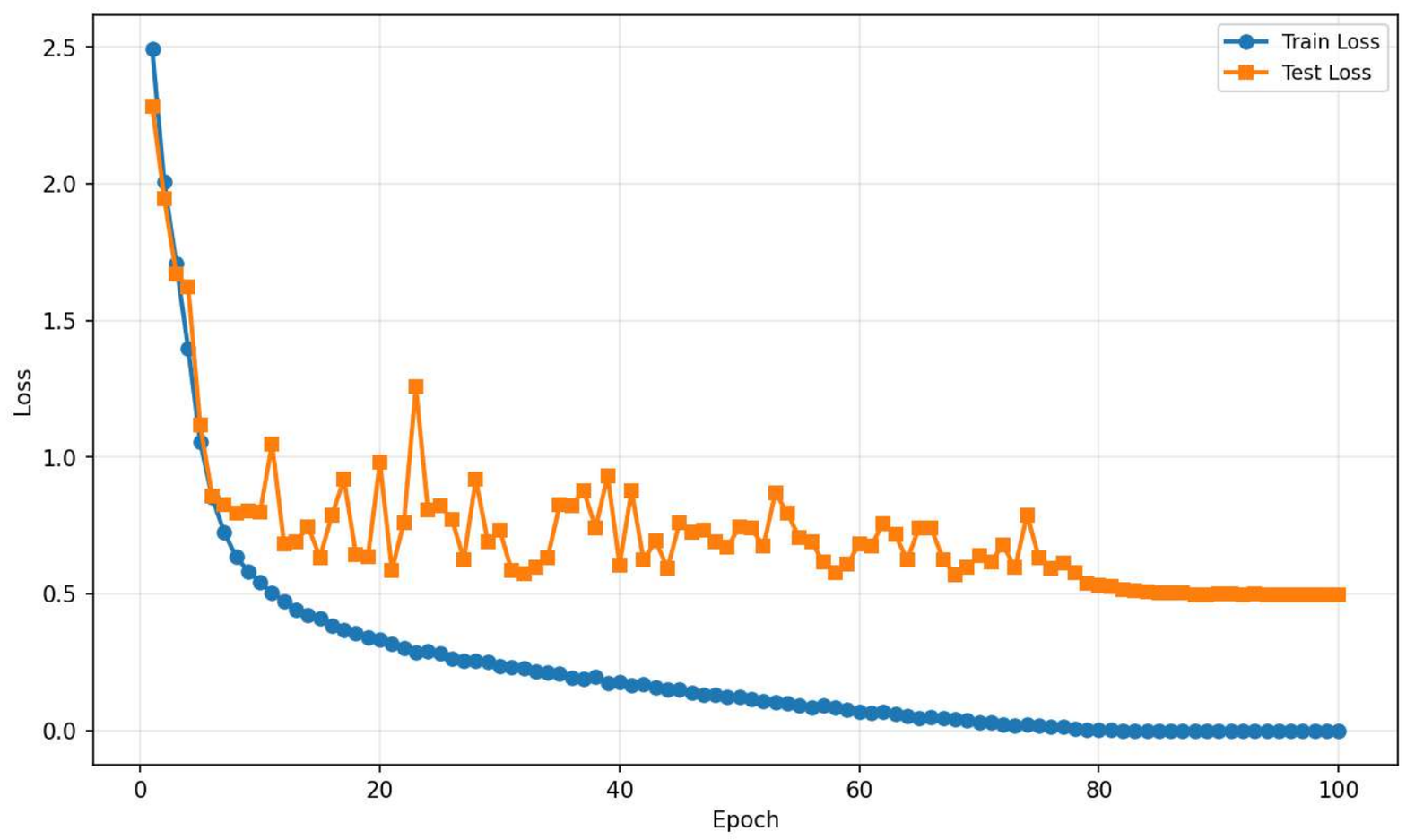}
        \caption{Training and test loss.}
        \label{fig:vgg_16_baseline_loss}
    \end{subfigure}
    \hfill
    \begin{subfigure}[]{0.32\textwidth}
        \centering
        \includegraphics[width=\textwidth]{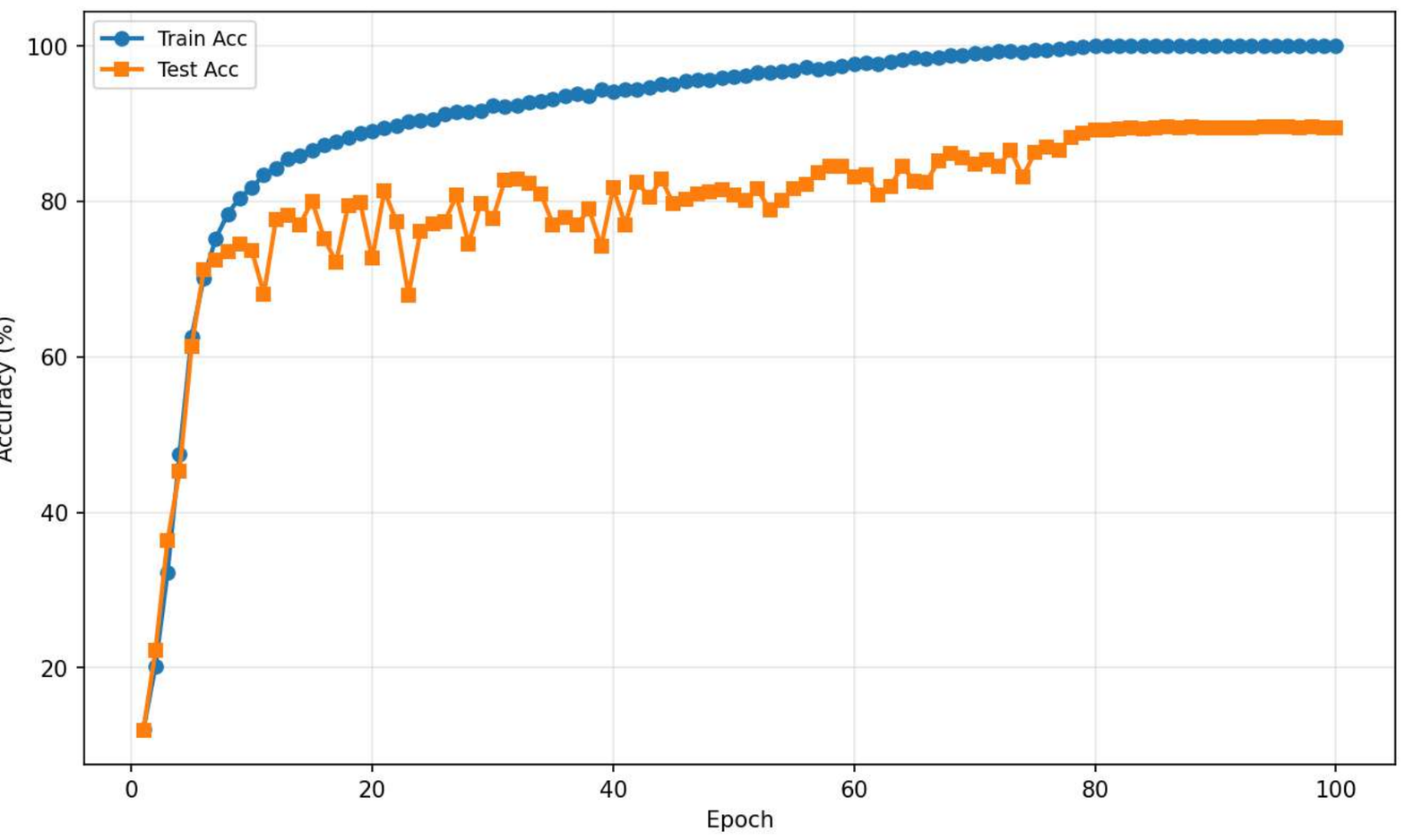}
        \caption{Training and test accuracy.}
        \label{fig:vgg_16_baseline_acc}
    \end{subfigure}
    \hfill
    \begin{subfigure}[]{0.32\textwidth}
        \centering
        \includegraphics[width=\textwidth]{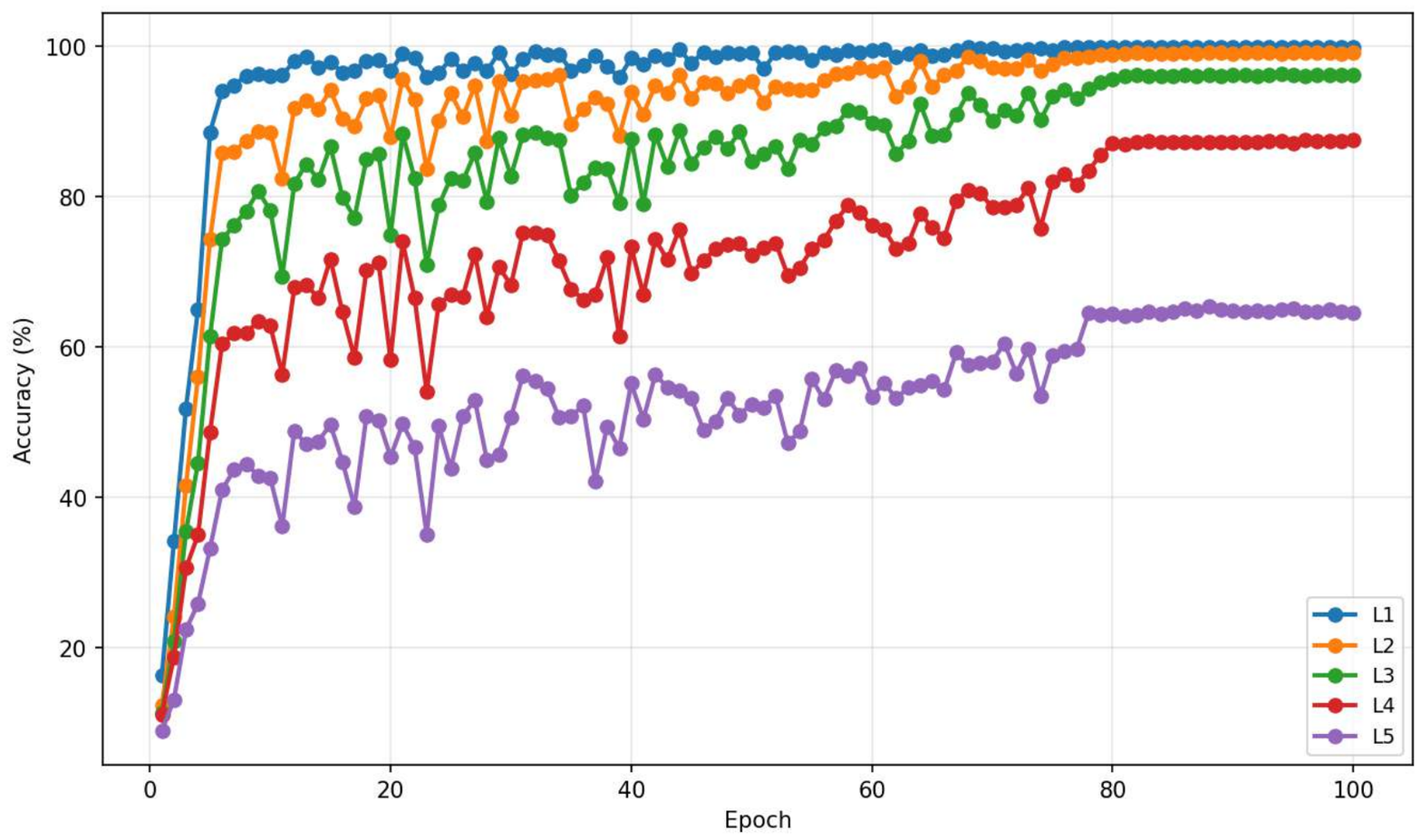}
        \caption{Stage-wise test accuracy.}
        \label{fig:vgg_16_baseline_stages}
    \end{subfigure}
    
   \caption{Training curves for \textbf{VGG-16 Baseline, Stage-Cosine LR}.}
\label{fig:appendix_vgg16_baseline_stages}

\end{figure*}

\begin{figure*}[h]
    \centering
    \begin{subfigure}[]{0.32\textwidth}
        \centering
        \includegraphics[width=\textwidth]{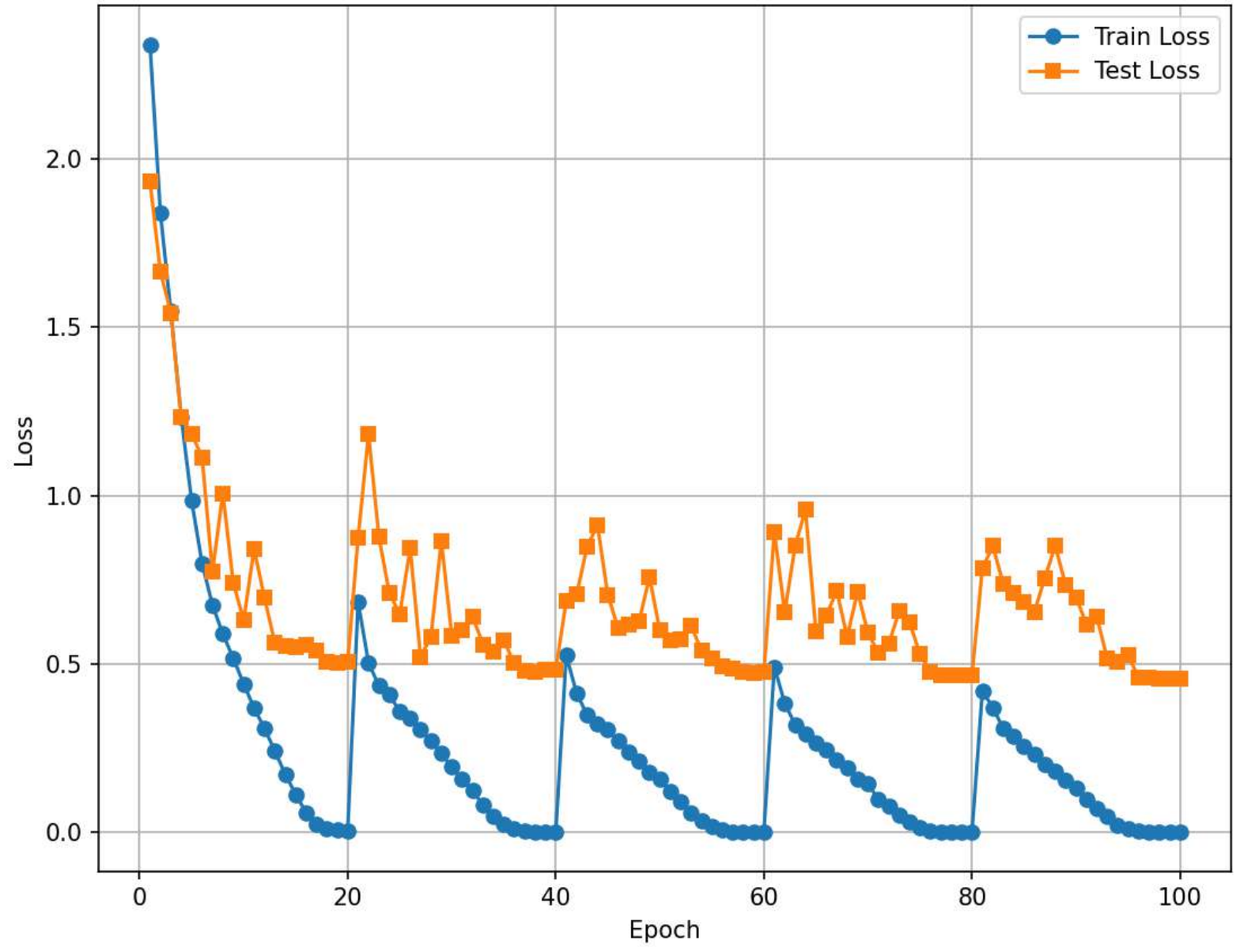}
        \caption{Training and test loss.}
        \label{fig:vgg_16_baseline_stage_loss}
    \end{subfigure}
    \hfill
    \begin{subfigure}[]{0.32\textwidth}
        \centering
        \includegraphics[width=\textwidth]{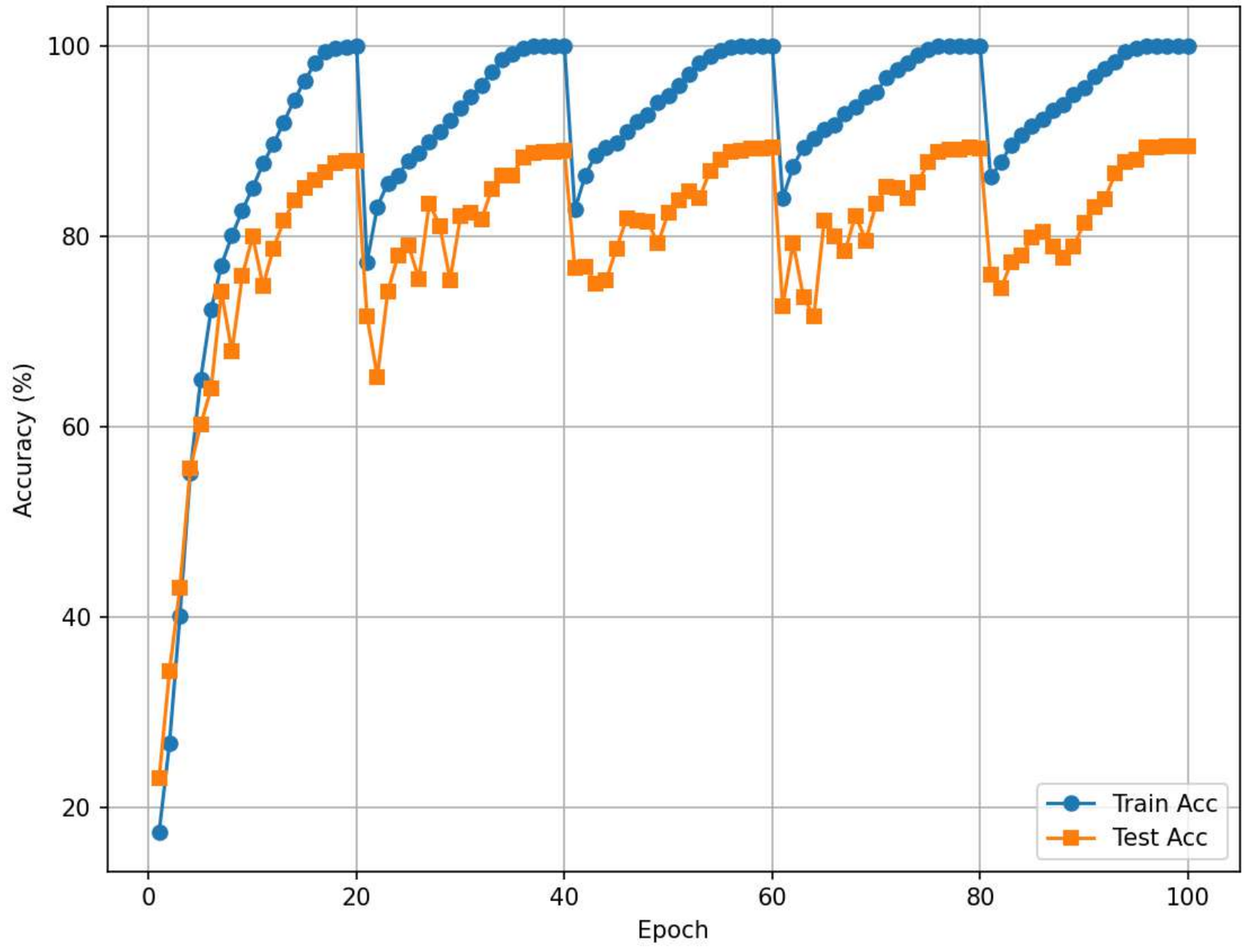}
        \caption{Training and test accuracy.}
        \label{fig:vgg_16_baseline_stage_acc}
    \end{subfigure}
    \hfill
    \begin{subfigure}[]{0.32\textwidth}
        \centering
        \includegraphics[width=\textwidth]{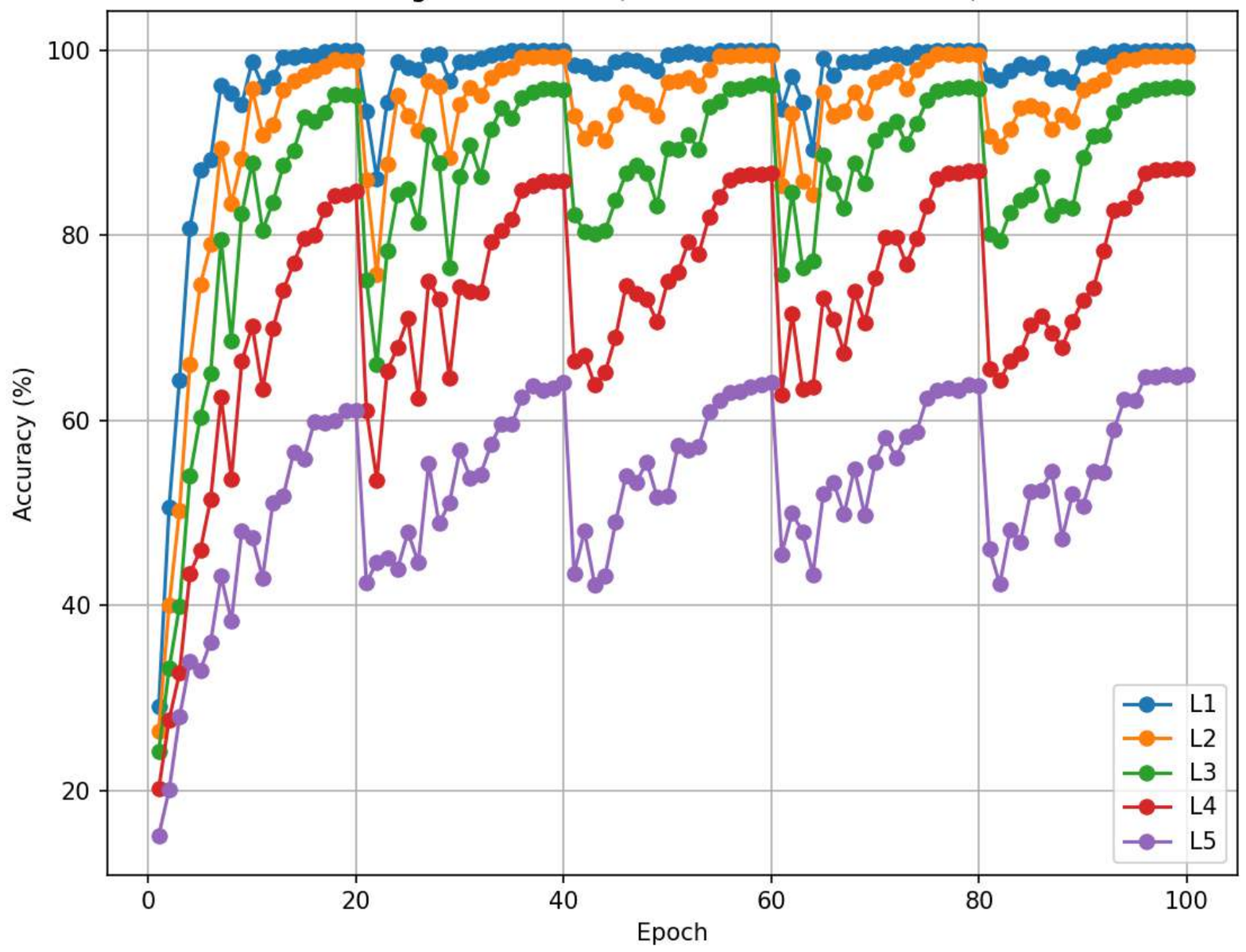}
        \caption{Stage-wise test accuracy.}
        \label{fig:vgg_16_baseline_stage_stages}
    \end{subfigure}
    
    \caption{Training curves for \textbf{ResNet-18 Curriculum, Cosine LR}.}
\label{fig:appendix_restnet18_curriculum}
\end{figure*}

\begin{figure*}[h]
    \centering
    \begin{subfigure}[]{0.32\textwidth}
        \centering
        \includegraphics[width=\textwidth]{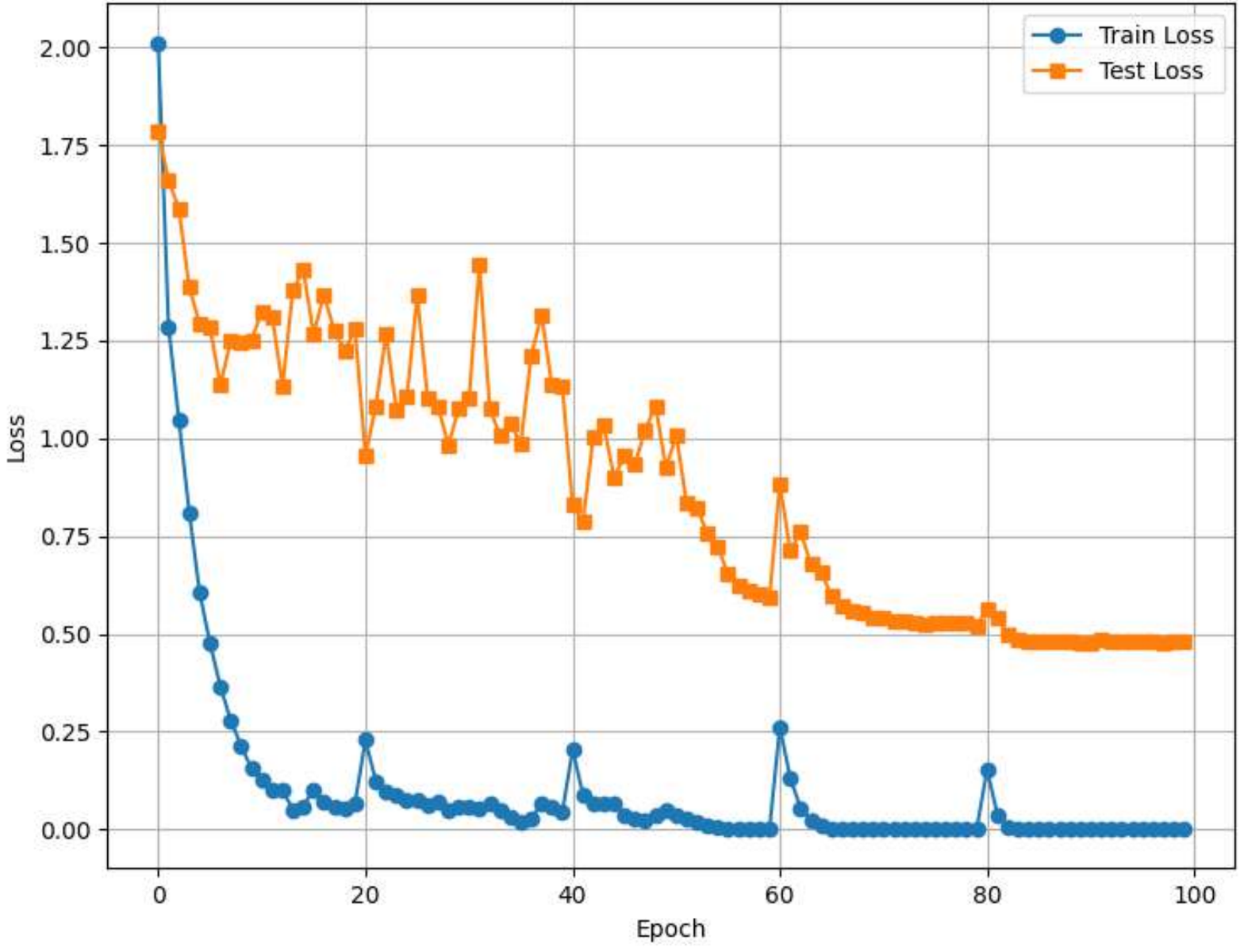}
        \caption{Training and test loss.}
        \label{fig:restnet18_curriculum_loss}
    \end{subfigure}
    \hfill
    \begin{subfigure}[]{0.32\textwidth}
        \centering
        \includegraphics[width=\textwidth]{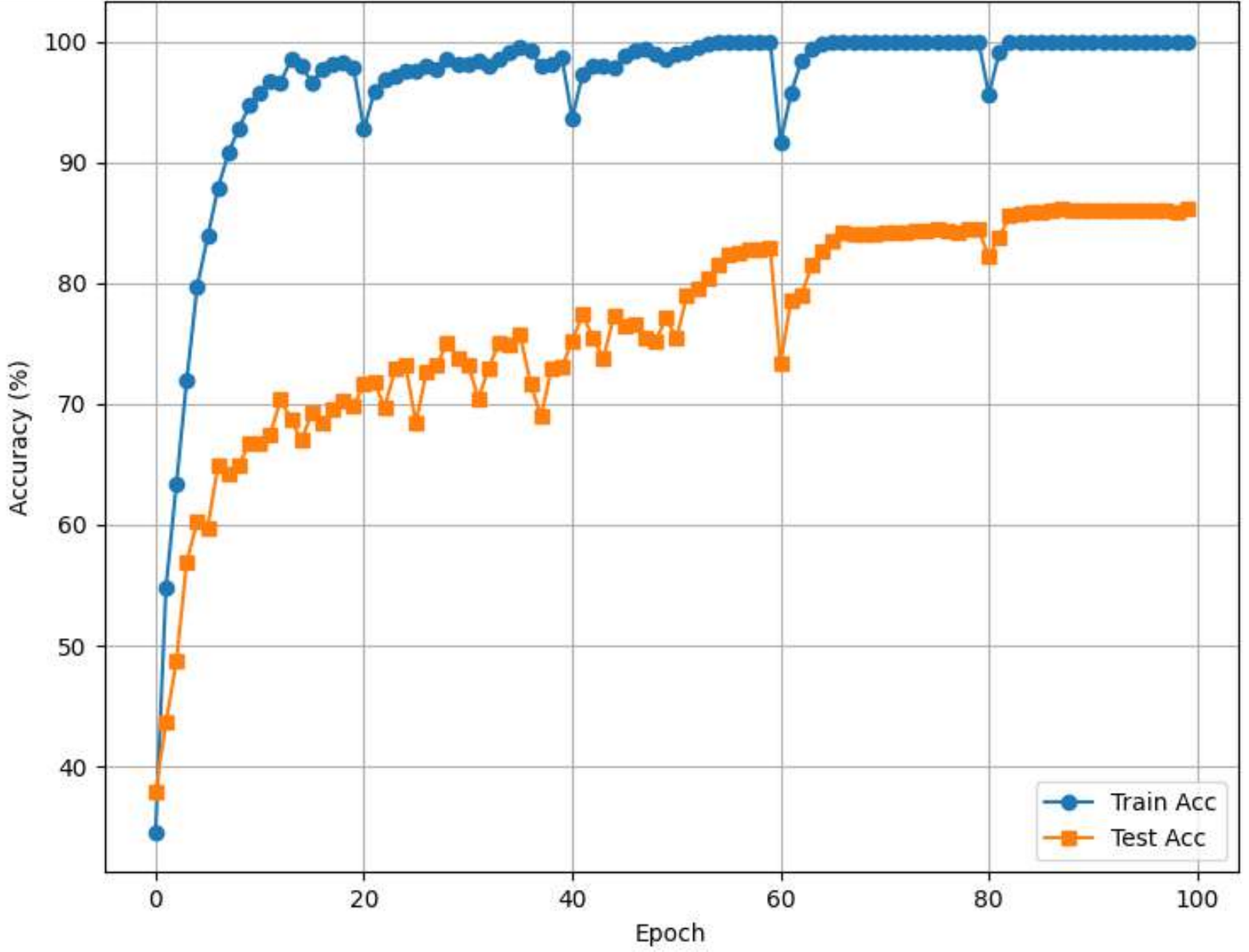}
        \caption{Training and test accuracy.}
        \label{fig:restnet18_curriculum_acc}
    \end{subfigure}
    \hfill
    \begin{subfigure}[]{0.32\textwidth}
        \centering
        \includegraphics[width=\textwidth]{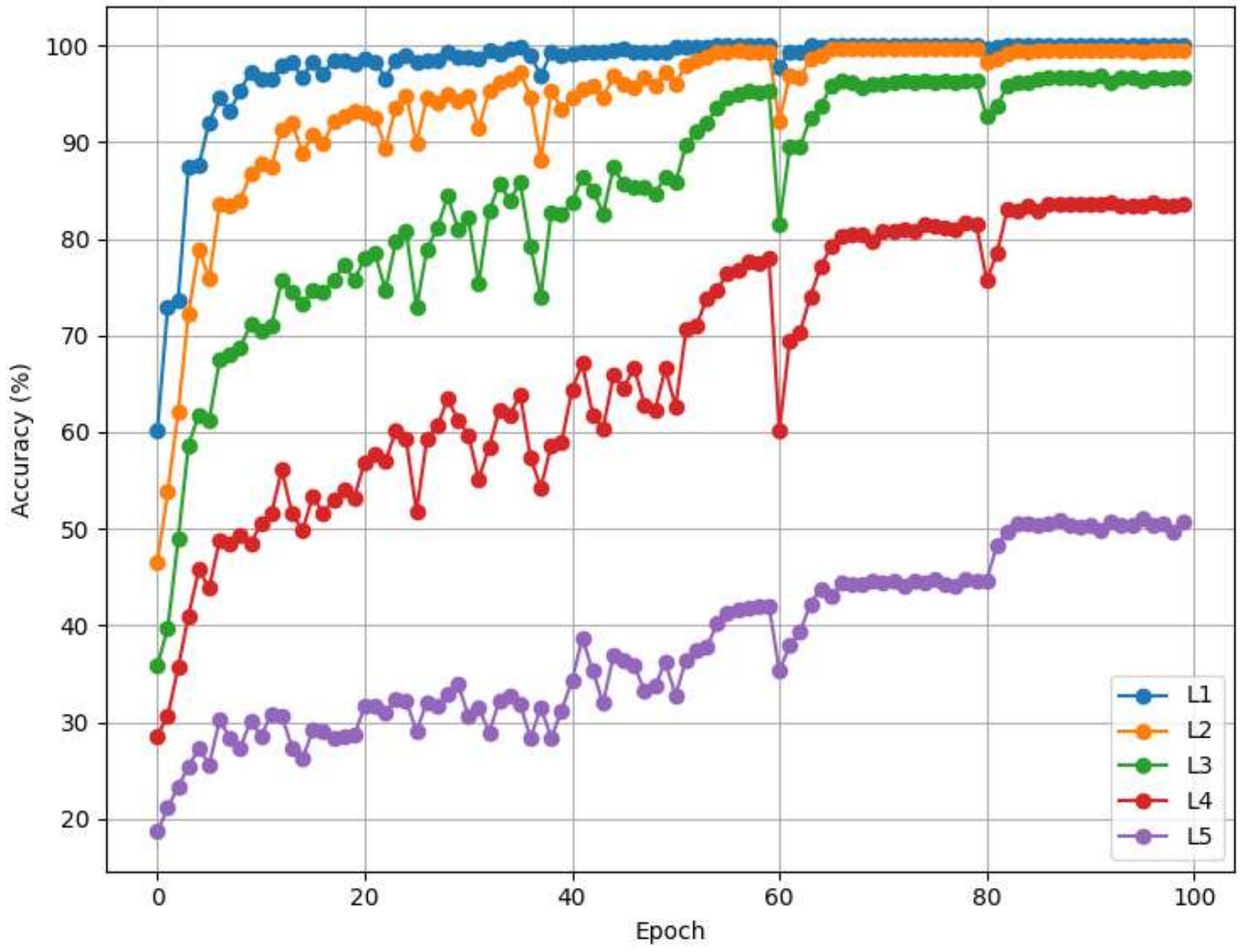}
        \caption{Stage-wise test accuracy.}
        \label{fig:restnet18_curriculum_stages}
    \end{subfigure}
    
   \caption{Training curves for \textbf{ResNet-18 Curriculum, Stage-Cosine LR}.}
\label{fig:appendix_restnet18_curriculum_stages}

\end{figure*}

\begin{figure*}[h]
    \centering
    \begin{subfigure}[]{0.32\textwidth}
        \centering
        \includegraphics[width=\textwidth]{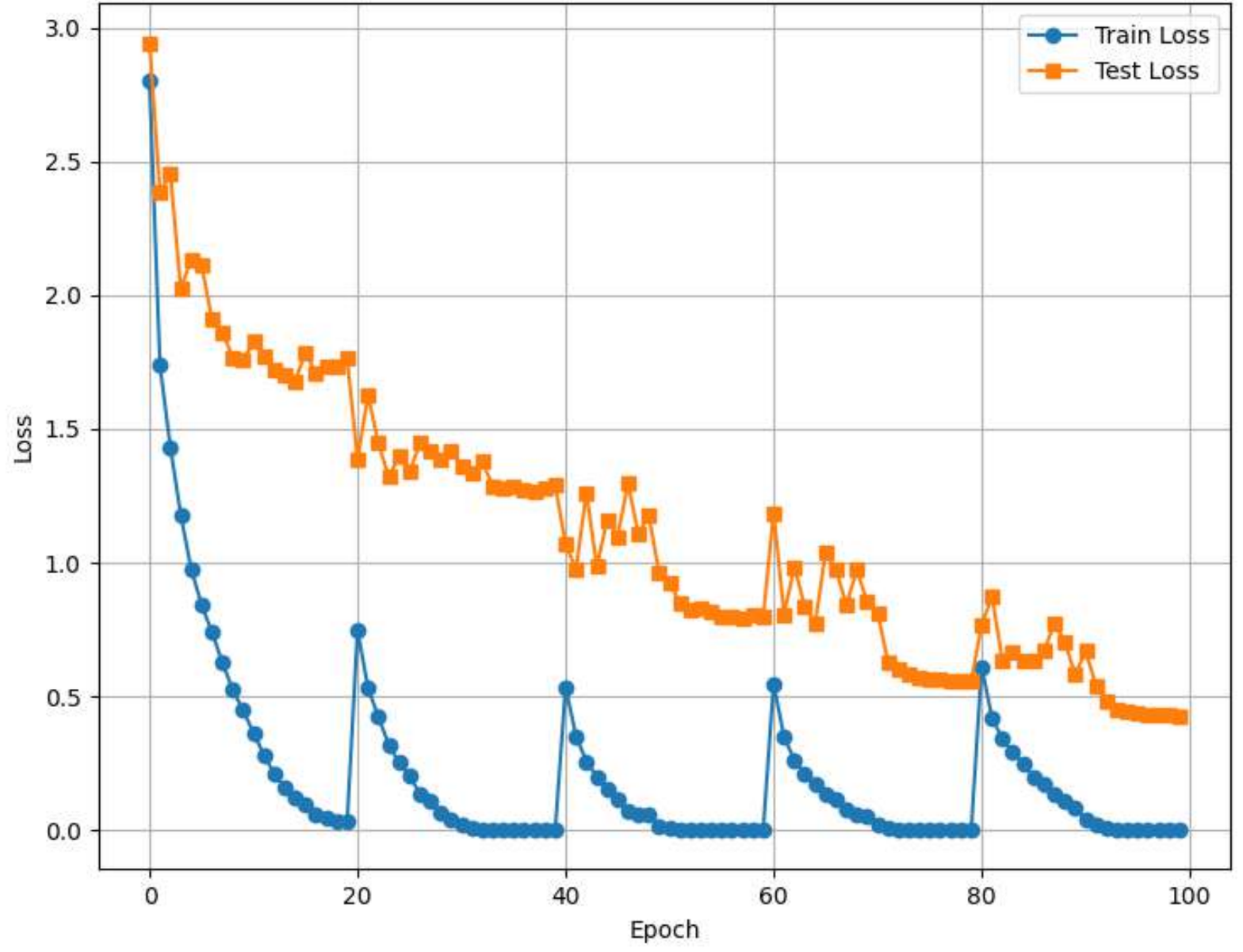}
        \caption{Training and test loss.}
        \label{fig:restnet_curriculum_stage_loss}
    \end{subfigure}
    \hfill
    \begin{subfigure}[]{0.32\textwidth}
        \centering
        \includegraphics[width=\textwidth]{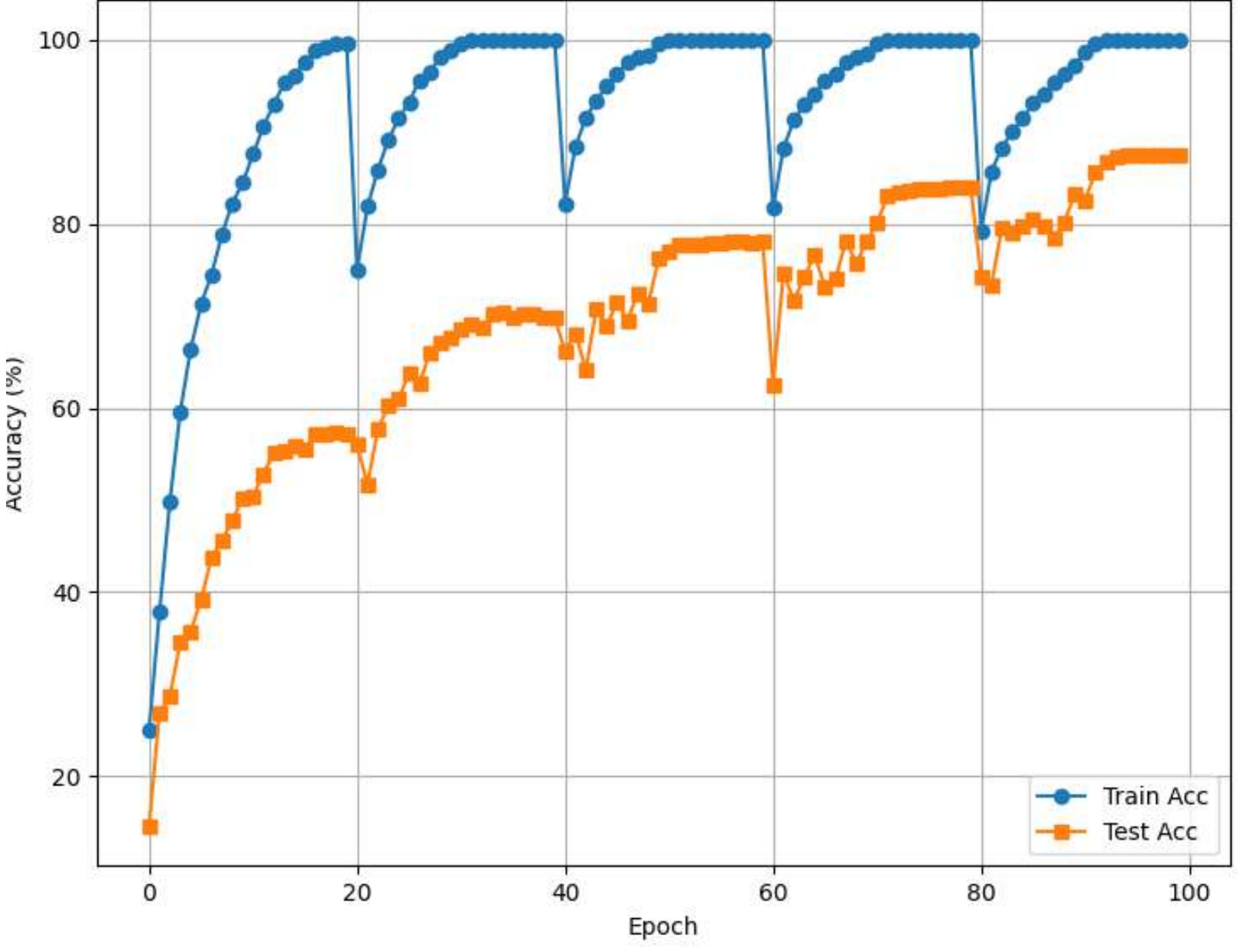}
        \caption{Training and test accuracy.}
        \label{fig:restnet_curriculum_stage_acc}
    \end{subfigure}
    \hfill
    \begin{subfigure}[]{0.32\textwidth}
        \centering
        \includegraphics[width=\textwidth]{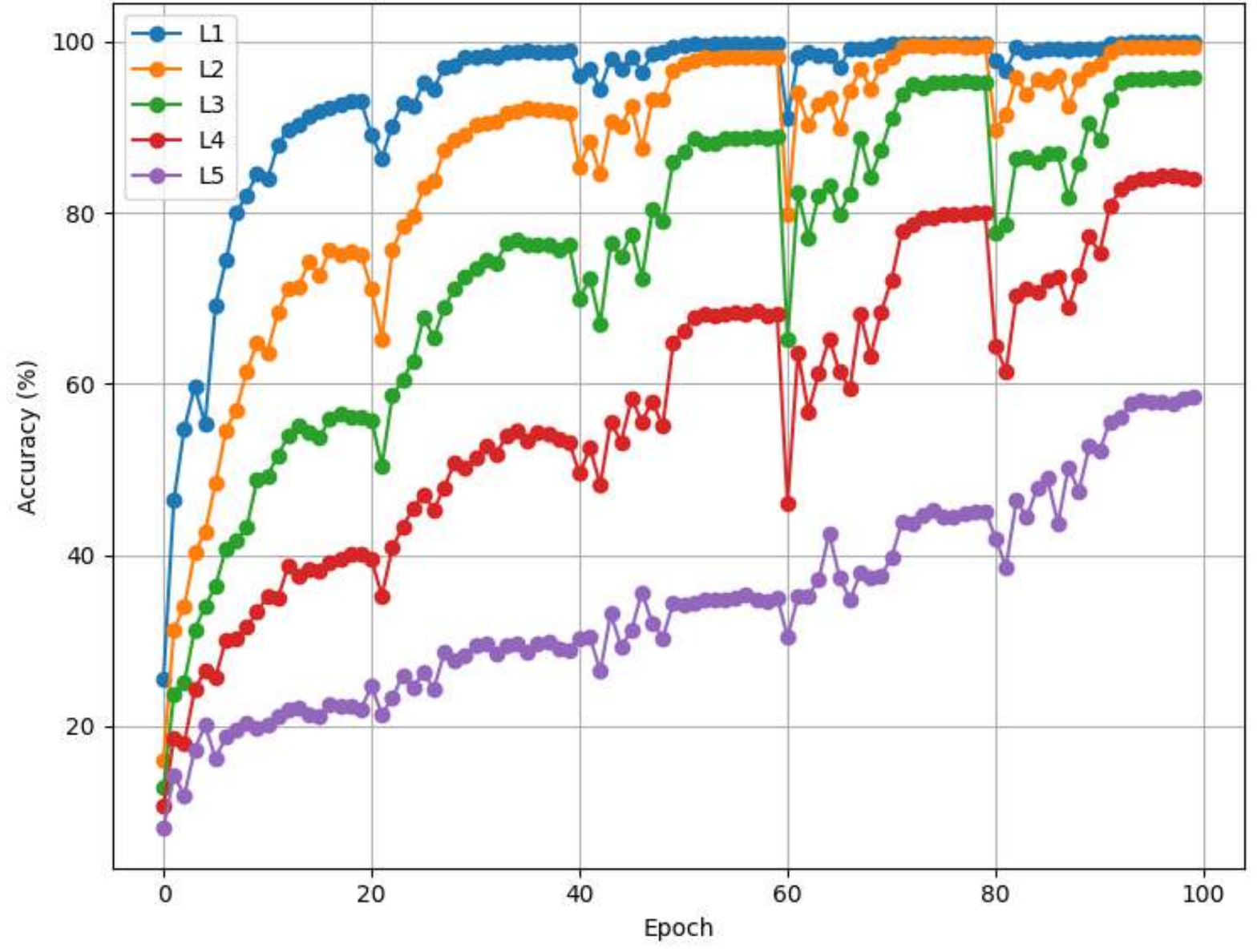}
        \caption{Stage-wise test accuracy.}
        \label{fig:restnet18_curriculum_stage_stages}
    \end{subfigure}
    
    \caption{Training curves for \textbf{VGG-16 Curriculum, Stage-Cosine LR}.}
\label{fig:appendix_vgg16_curriculum_stages}
\end{figure*}

\begin{figure*}[h]
    \centering
    \begin{subfigure}[]{0.32\textwidth}
        \centering
        \includegraphics[width=\textwidth]{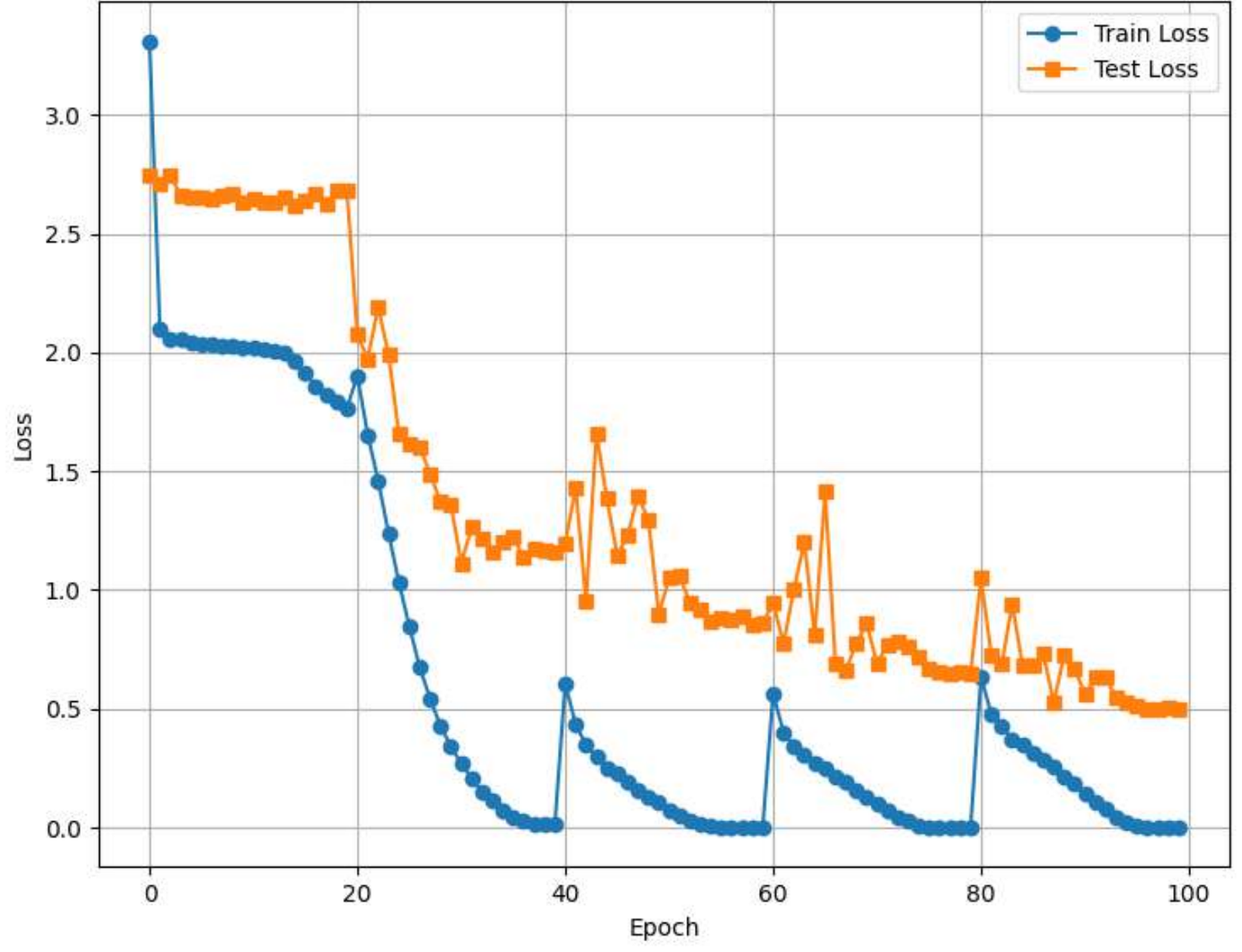}
        \caption{Training and test loss.}
        \label{fig:vgg16_curriculum_stage_loss}
    \end{subfigure}
    \hfill
    \begin{subfigure}[]{0.32\textwidth}
        \centering
        \includegraphics[width=\textwidth]{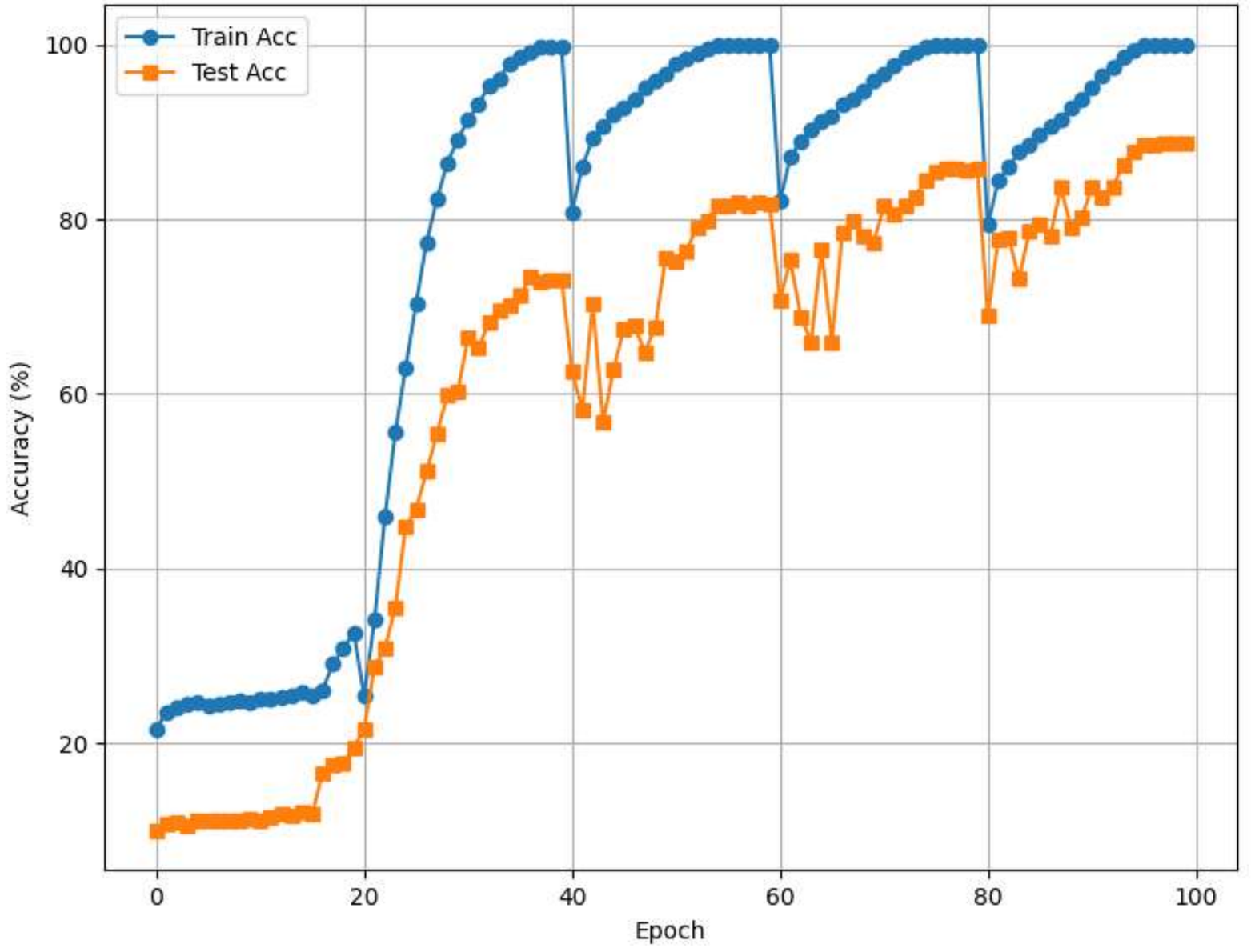}
        \caption{Training and test accuracy.}
        \label{fig:vgg16_curriculum_stage_acc}
    \end{subfigure}
    \hfill
    \begin{subfigure}[]{0.32\textwidth}
        \centering
        \includegraphics[width=\textwidth]{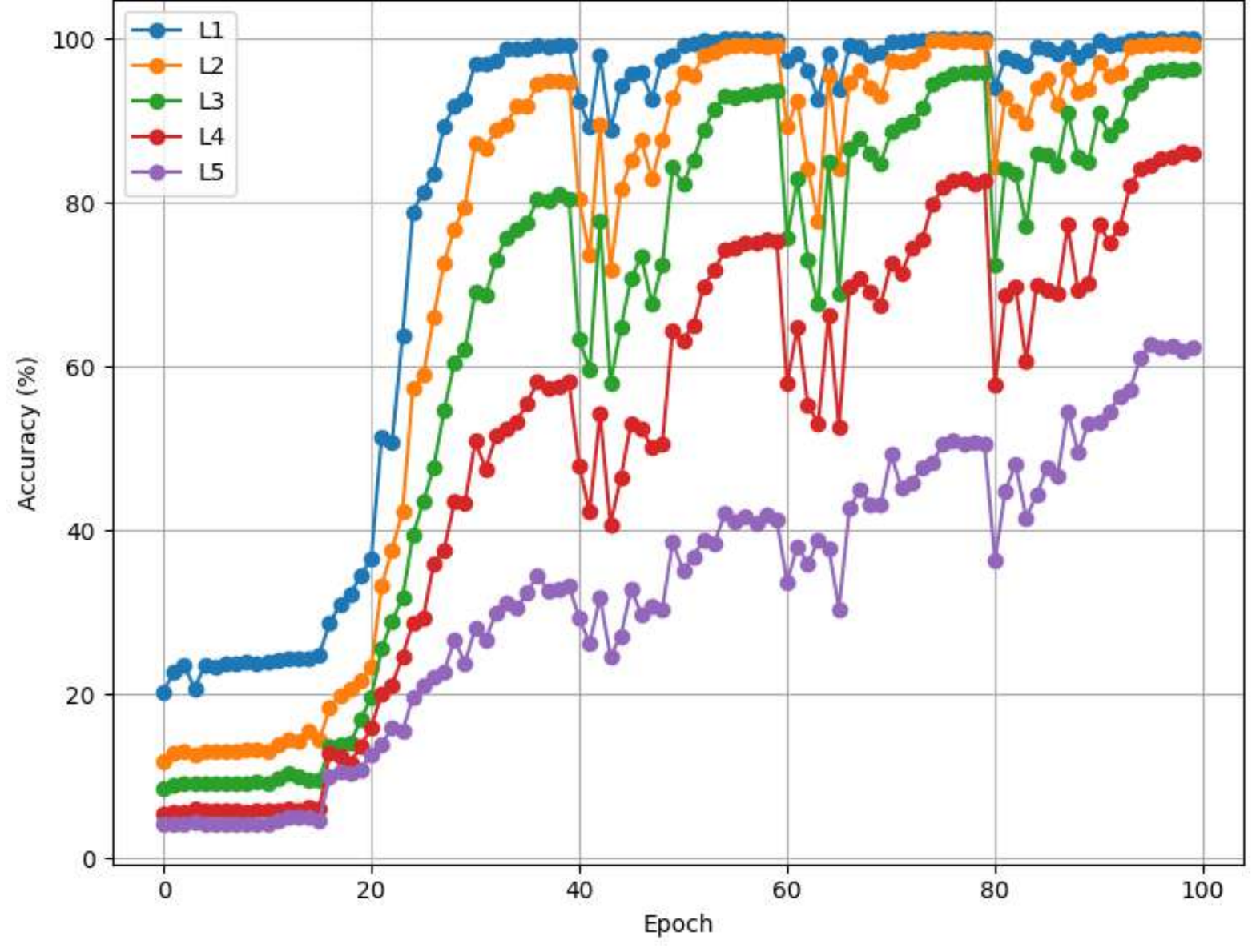}
        \caption{Stage-wise test accuracy.}
        \label{fig:vgg16_curriculum_stage_stages}
    \end{subfigure}
    
   \caption{Training curves for \textbf{VGG-16 Curriculum, Cosine LR}.}
\label{fig:appendix_vgg16_curriculum}
\end{figure*}

\begin{figure*}[h]
    \centering
    \begin{subfigure}[]{0.32\textwidth}
        \centering
        \includegraphics[width=\textwidth]{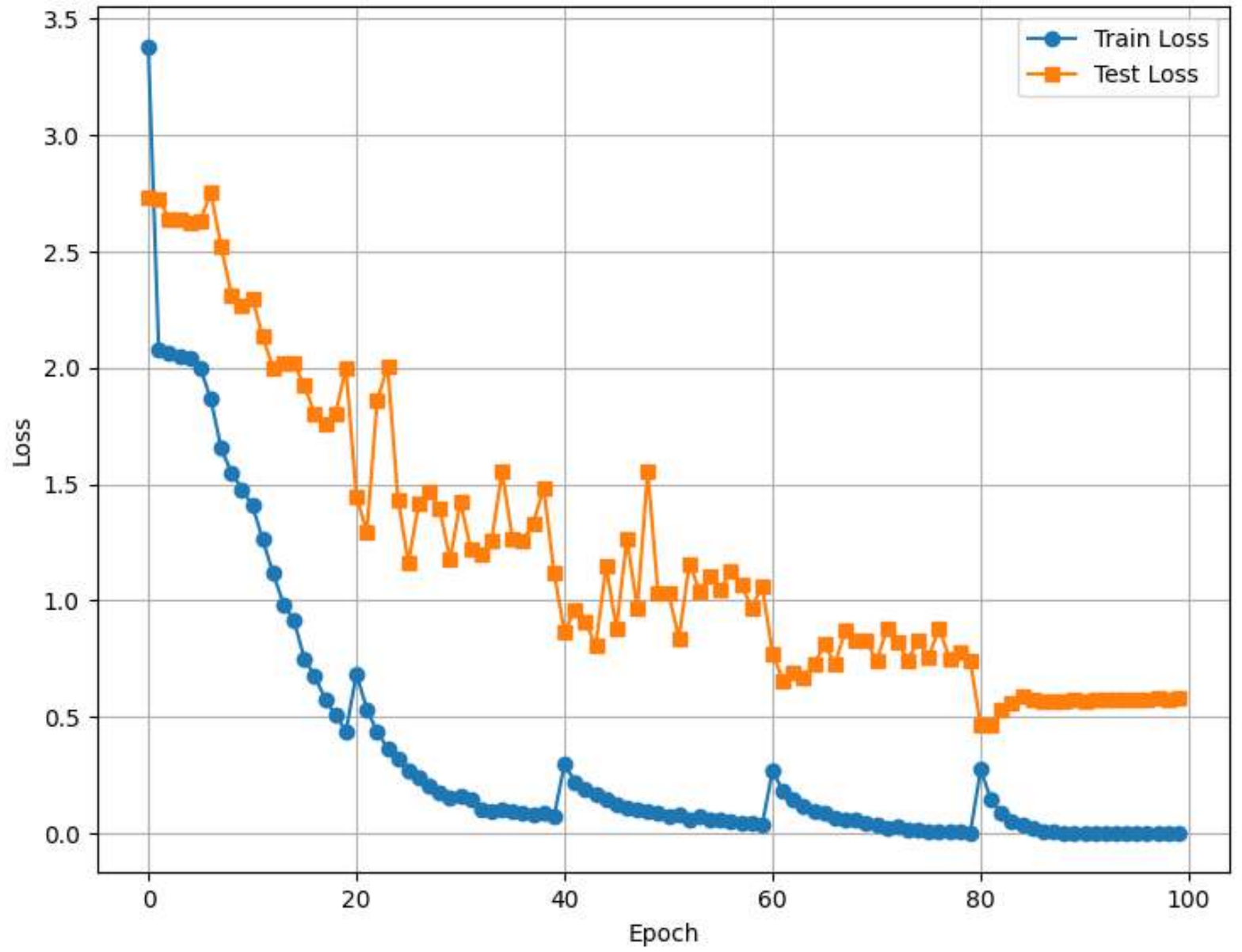}
        \caption{Training and test loss.}
        \label{fig:vgg16_curriculum_loss}
    \end{subfigure}
    \hfill
    \begin{subfigure}[]{0.32\textwidth}
        \centering
        \includegraphics[width=\textwidth]{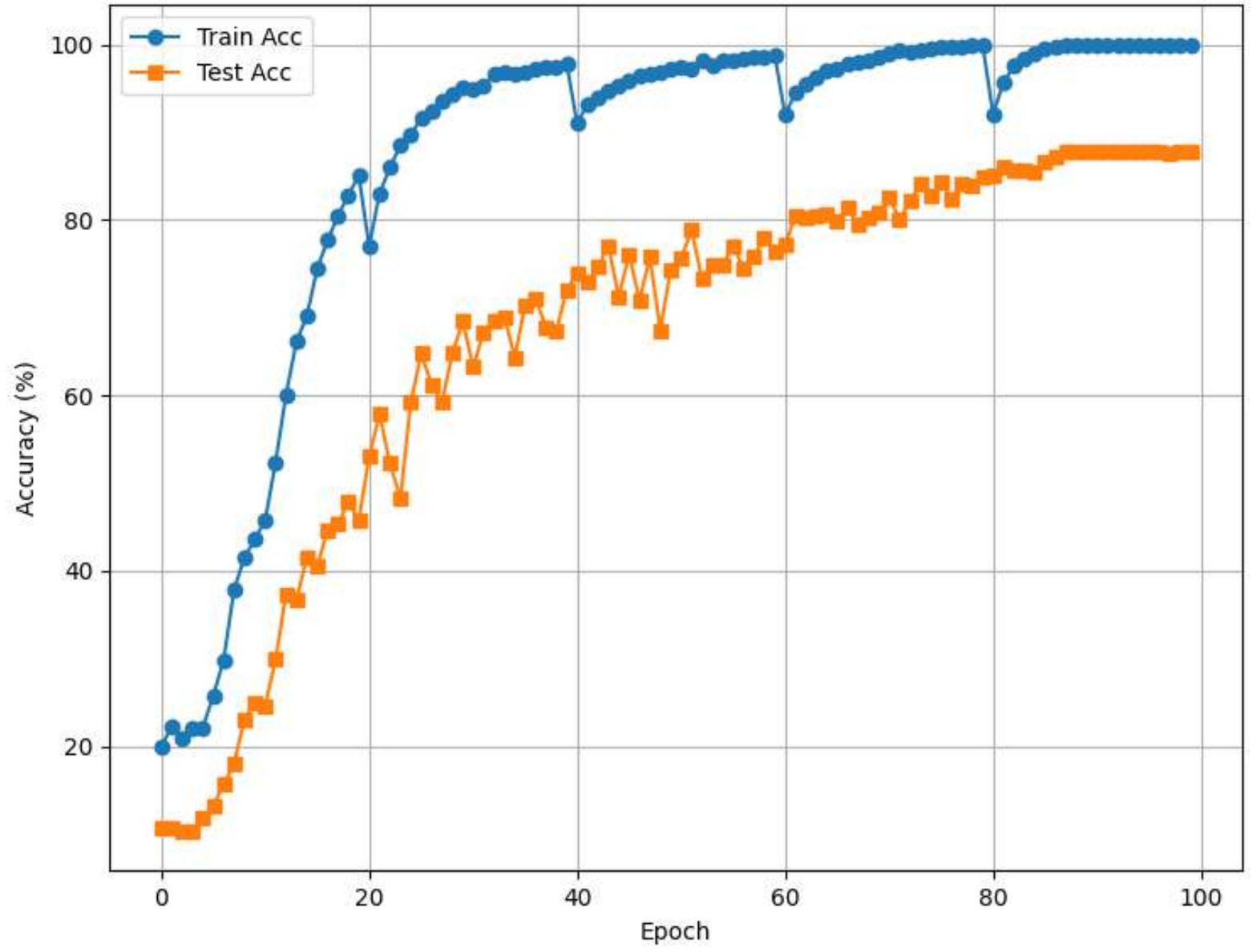}
        \caption{Training and test accuracy.}
        \label{fig:vgg16_curriculum_acc}
    \end{subfigure}
    \hfill
    \begin{subfigure}[]{0.32\textwidth}
        \centering
        \includegraphics[width=\textwidth]{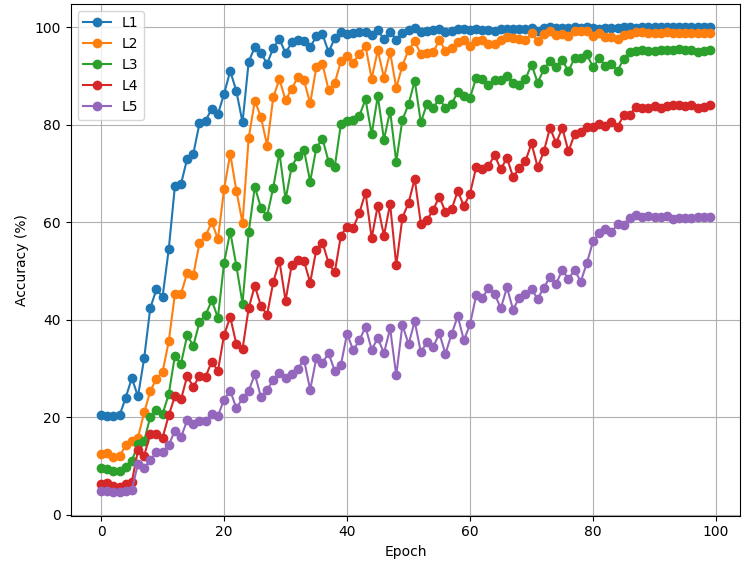}
        \caption{Stage-wise test accuracy.}
        \label{fig:vgg16_curriculum_stages}
    \end{subfigure}
    
    \caption{Training curves for \textbf{VGG-16 Anti-Curriculum, Cosine LR}.}
\label{fig:appendix_vgg16_anticosine}
\end{figure*}

\begin{figure*}[h]
    \centering
    \begin{subfigure}[]{0.32\textwidth}
        \centering
        \includegraphics[width=\textwidth]{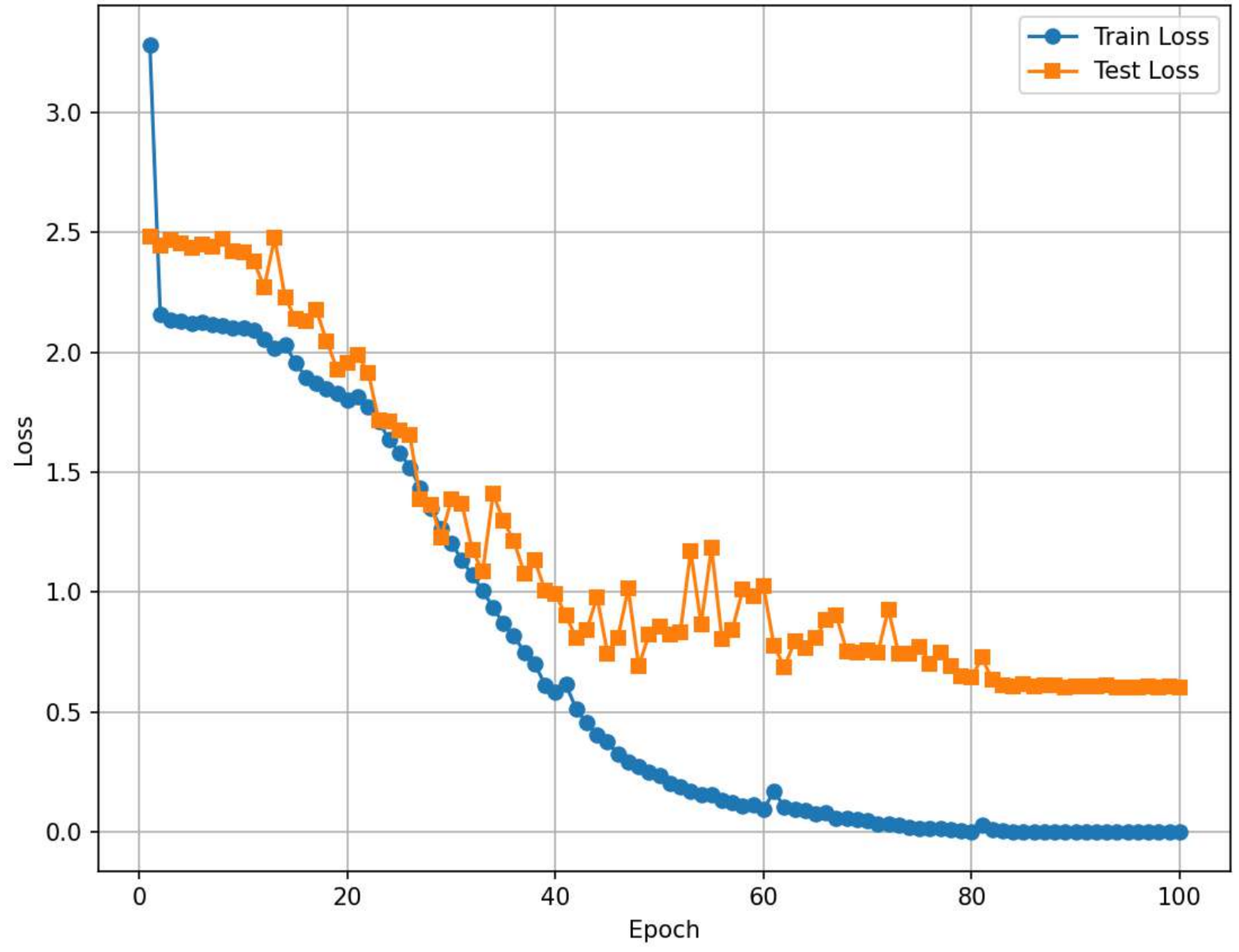}
        \caption{Training and test loss.}
        \label{fig:vgg_16_anticosine_loss}
    \end{subfigure}
    \hfill
    \begin{subfigure}[]{0.32\textwidth}
        \centering
        \includegraphics[width=\textwidth]{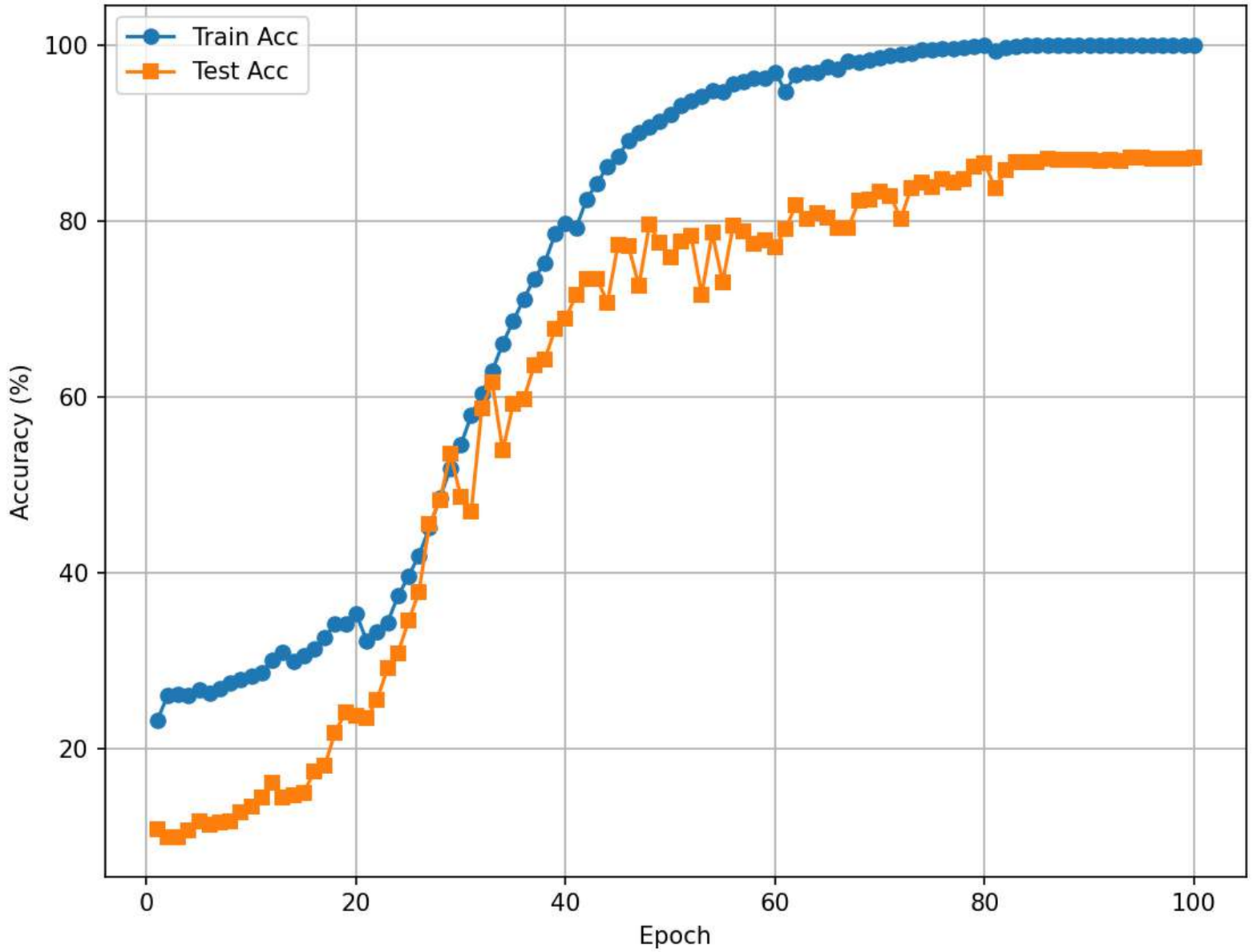}
        \caption{Training and test accuracy.}
        \label{fig:vgg_16_anticosine_acc}
    \end{subfigure}
    \hfill
    \begin{subfigure}[]{0.32\textwidth}
        \centering
        \includegraphics[width=\textwidth]{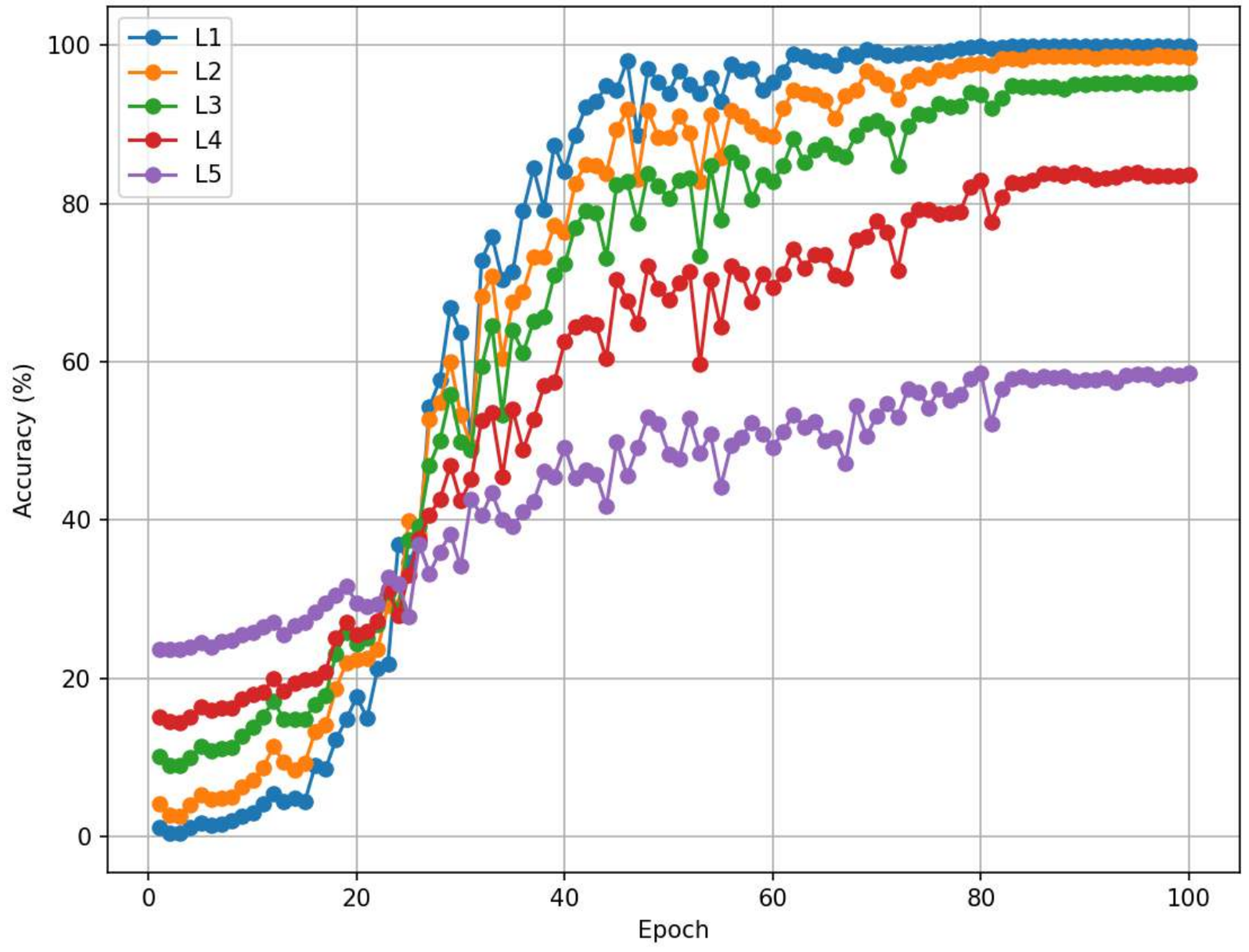}
        \caption{Stage-wise test accuracy.}
        \label{fig:vgg_16_anticosine_stages}
    \end{subfigure}
    
    \caption{Training curves for \textbf{VGG-16 Baseline + Pacing, Cosine LR}.}
\label{fig:appendix_vgg16_Pacing}
\end{figure*}

\begin{figure*}[h]
    \centering
    \begin{subfigure}[]{0.32\textwidth}
        \centering
        \includegraphics[width=\textwidth]{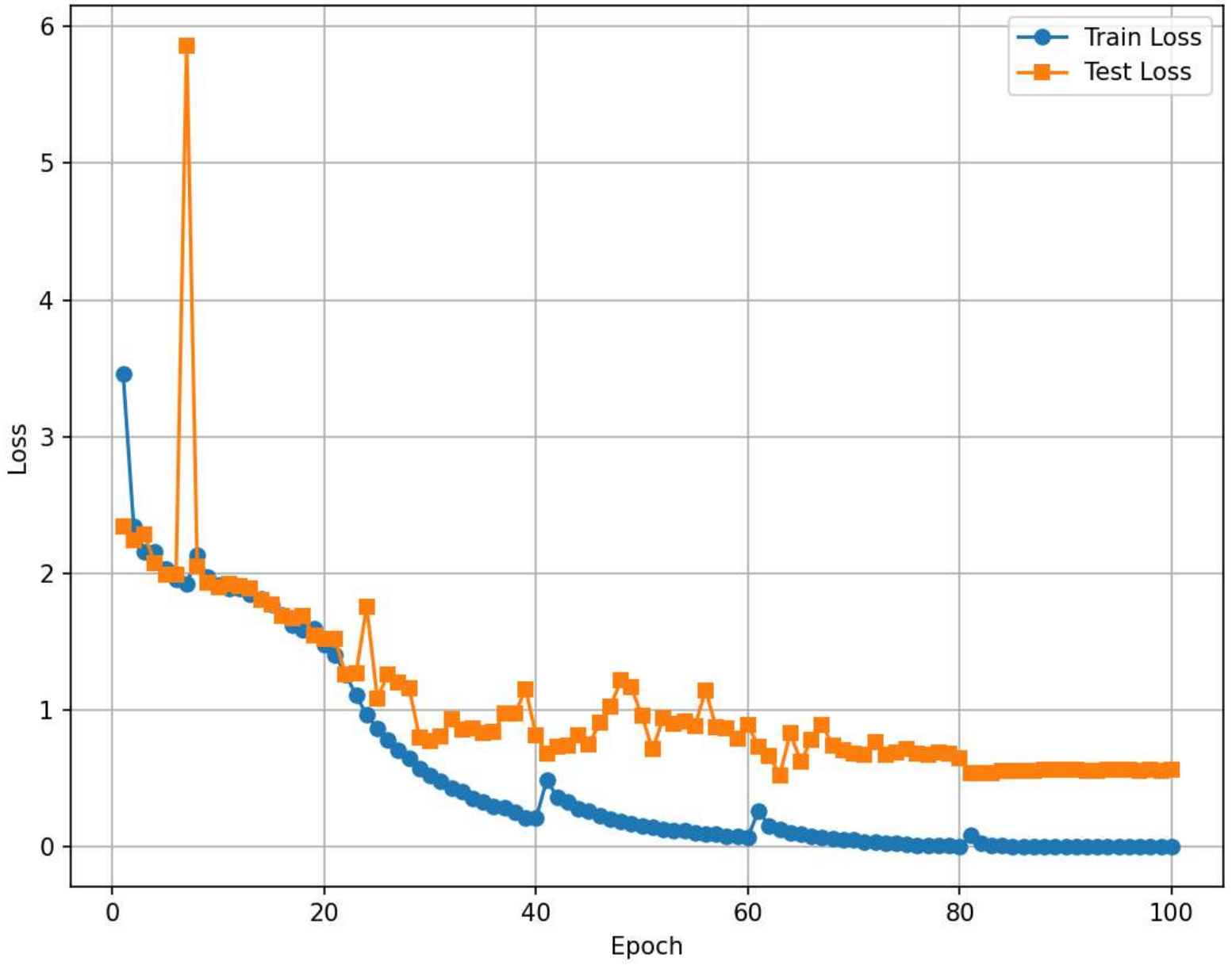}
        \caption{Training and test loss.}
        \label{fig:vgg16_pacing_loss}
    \end{subfigure}
    \hfill
    \begin{subfigure}[]{0.32\textwidth}
        \centering
        \includegraphics[width=\textwidth]{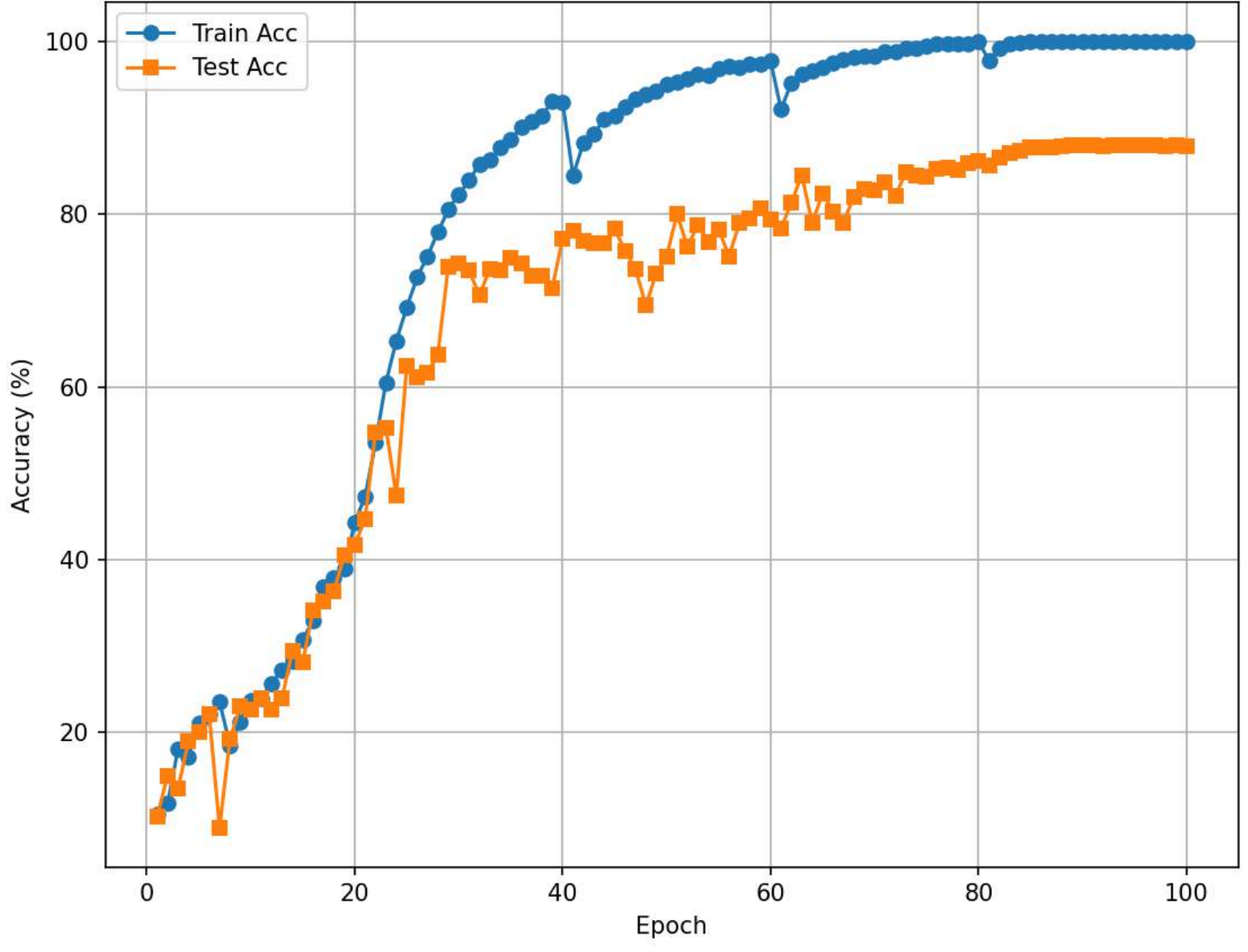}
        \caption{Training and test accuracy.}
        \label{fig:vgg16_pacing_acc}
    \end{subfigure}
    \hfill
    \begin{subfigure}[]{0.32\textwidth}
        \centering
        \includegraphics[width=\textwidth]{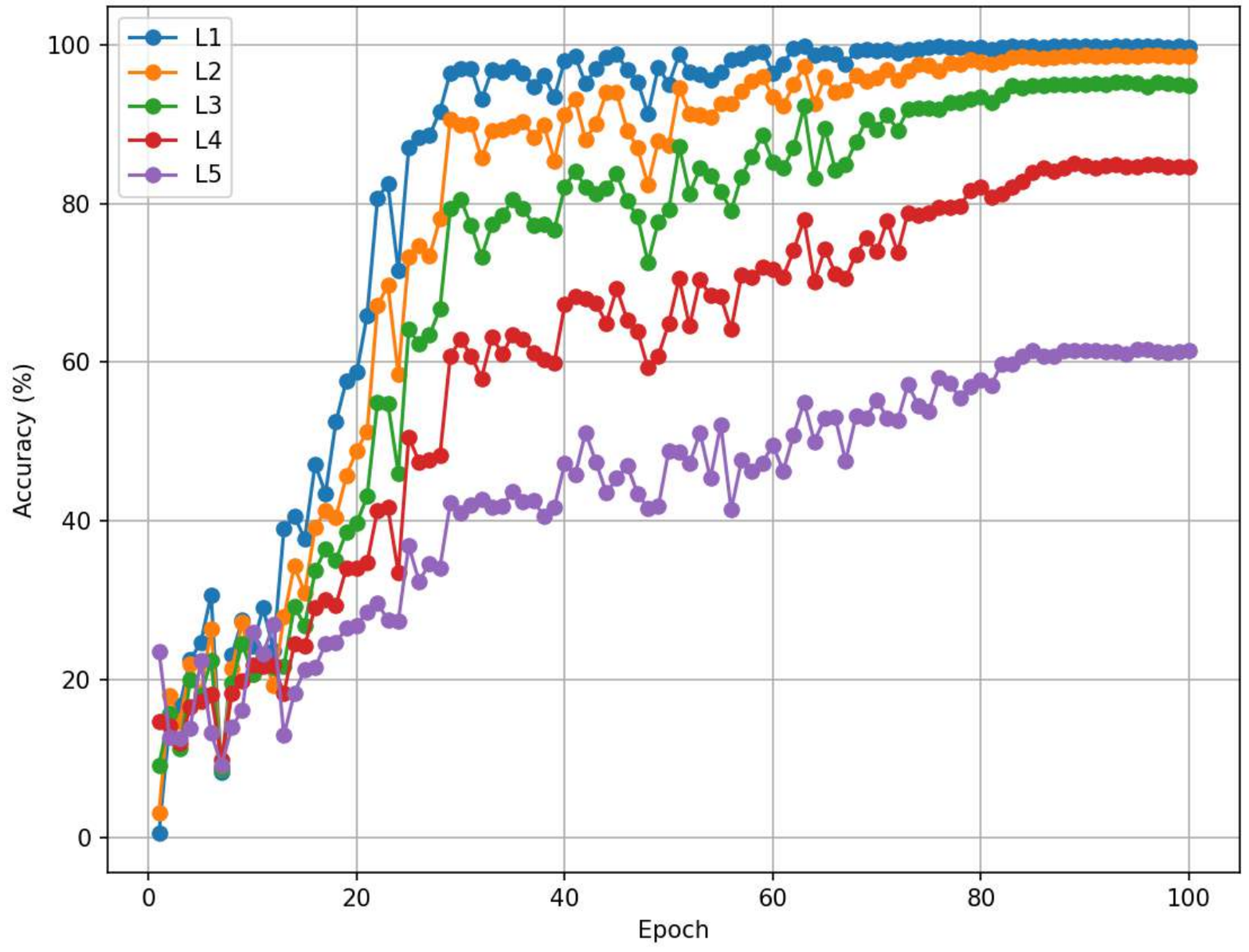}
        \caption{Stage-wise test accuracy.}
        \label{fig:vgg16_pacing_stages}
    \end{subfigure}
    
    \caption{Training curves for \textbf{ResNet-18 Baseline + Pacing, Cosine LR}.}
\label{fig:appendix_restnet_Paceing}
\end{figure*}

\begin{figure*}[h]
    \centering
    \begin{subfigure}[]{0.32\textwidth}
        \centering
        \includegraphics[width=\textwidth]{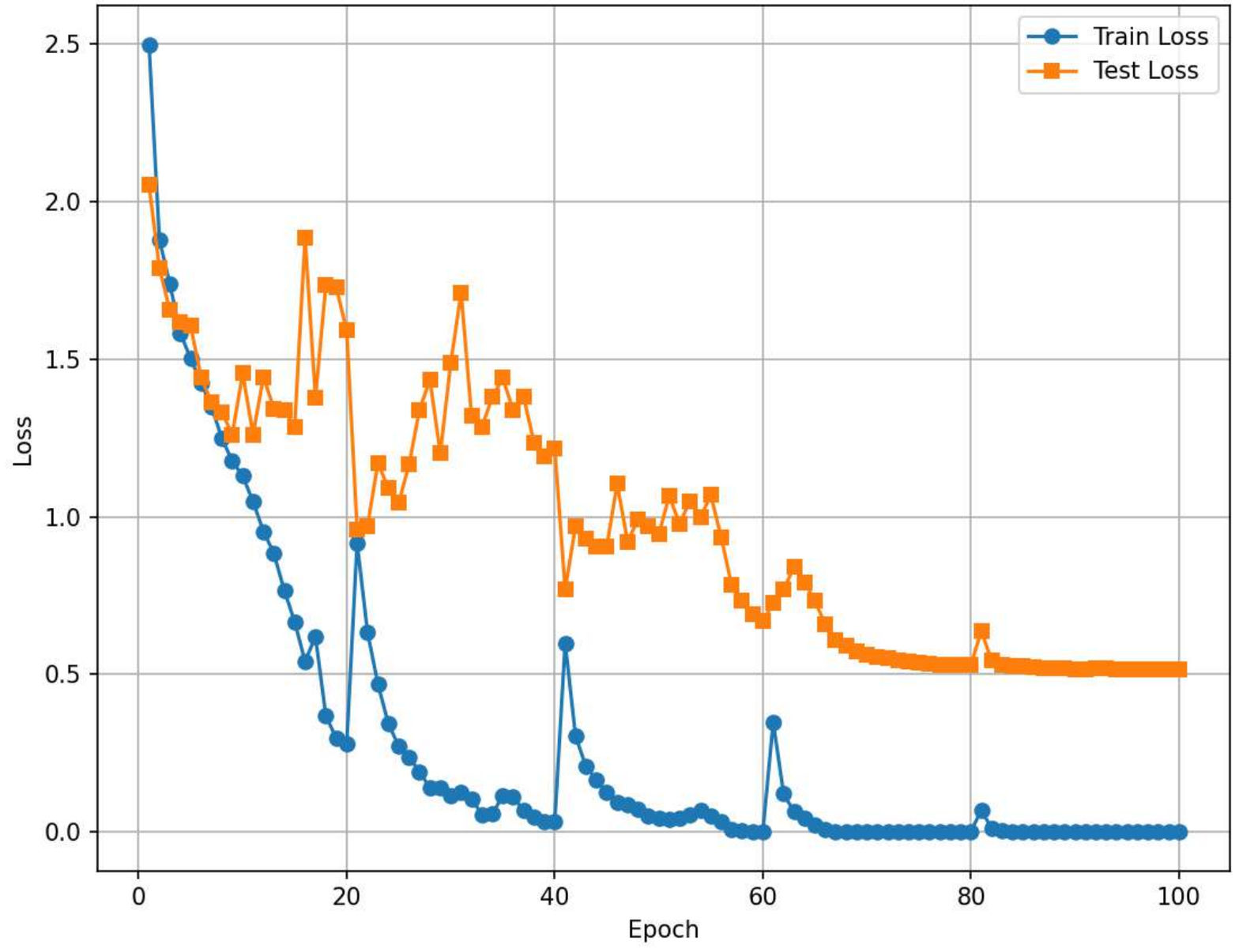}
        \caption{Training and test loss.}
        \label{fig:restnet_pacing_loss}
    \end{subfigure}
    \hfill
    \begin{subfigure}[]{0.32\textwidth}
        \centering
        \includegraphics[width=\textwidth]{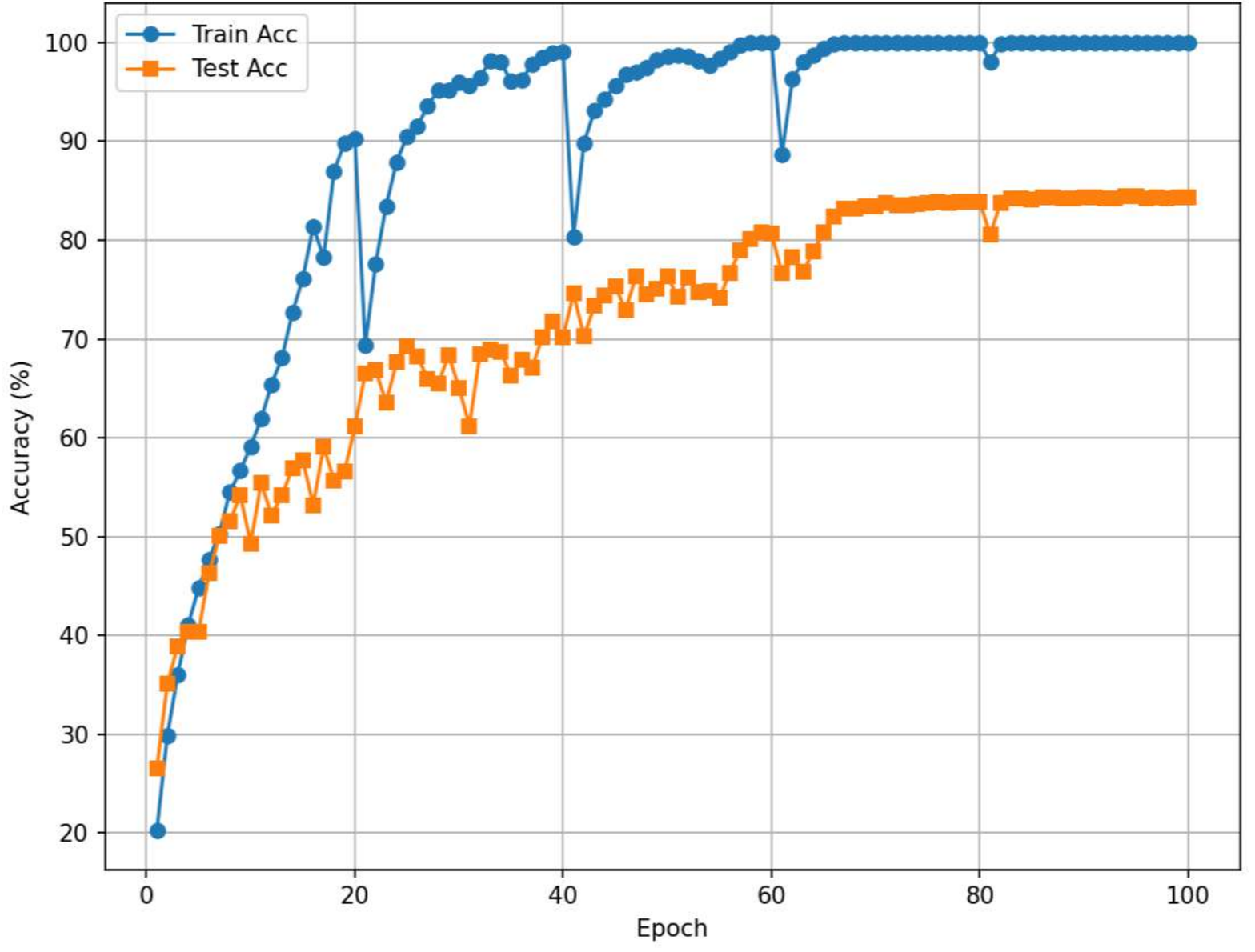}
        \caption{Training and test accuracy.}
        \label{fig:restnet_pacing_acc}
    \end{subfigure}
    \hfill
    \begin{subfigure}[]{0.32\textwidth}
        \centering
        \includegraphics[width=\textwidth]{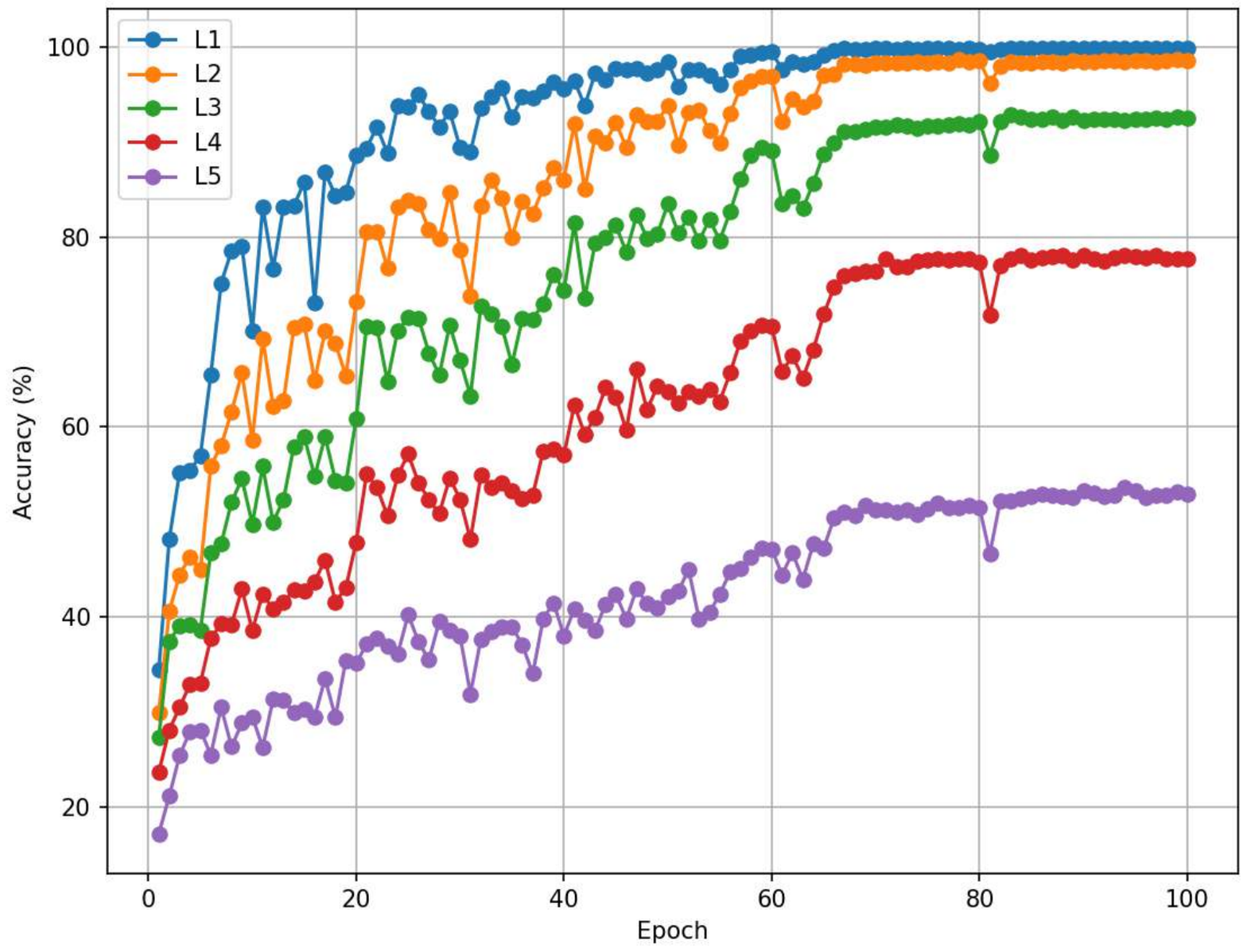}
        \caption{Stage-wise test accuracy.}
        \label{fig:restnet_pacing_stages}
    \end{subfigure}
    
  \caption{Training curves for \textbf{ResNet-18 Baseline + Pacing, Stage-Cosine LR}.}
\label{fig:appendix_restnet_Pacing_stagecosine}

\end{figure*}

\begin{figure*}[h]
    \centering
    \begin{subfigure}[]{0.32\textwidth}
        \centering
        \includegraphics[width=\textwidth]{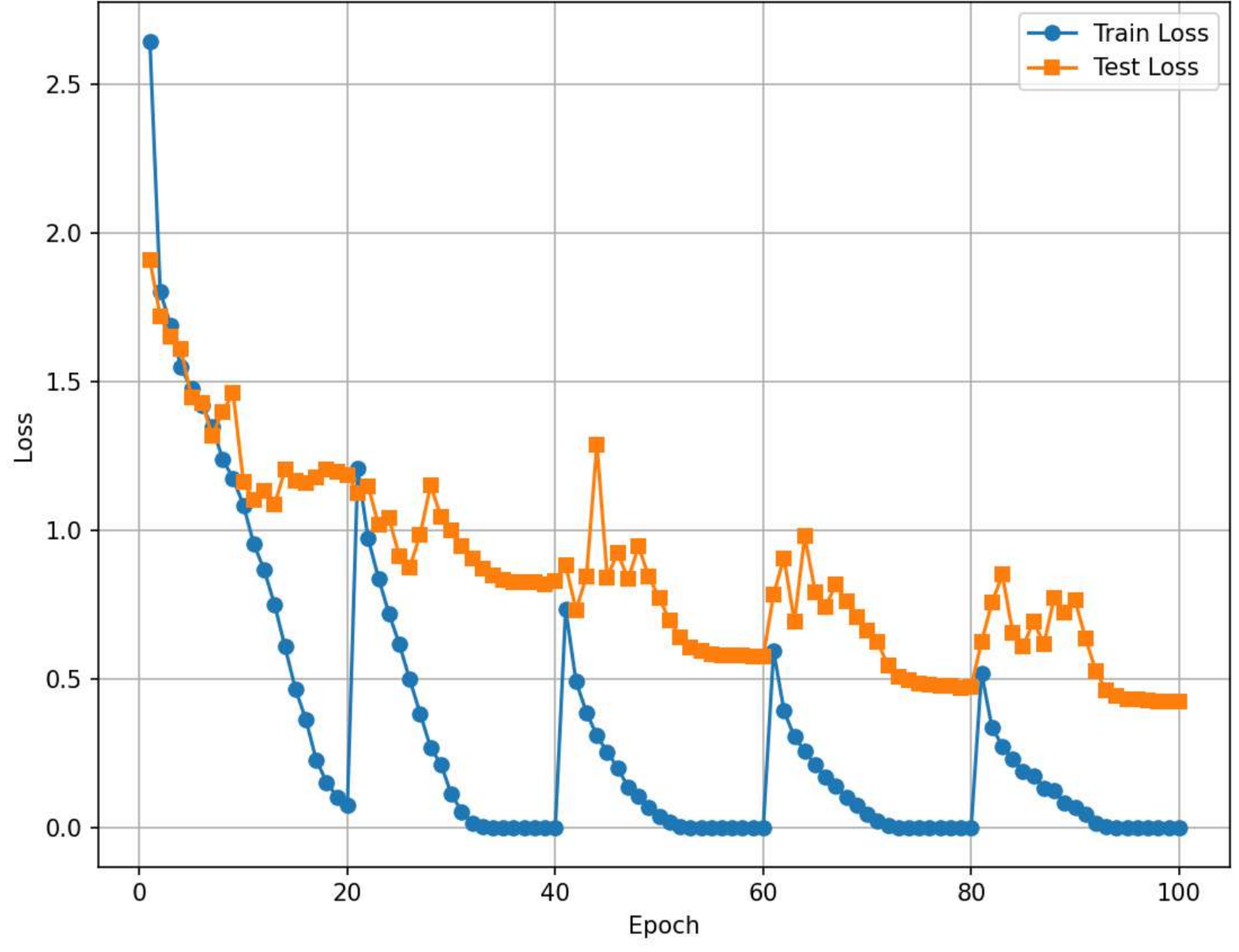}
        \caption{Training and test loss.}
        \label{fig:restnet_pacing_stage_loss}
    \end{subfigure}
    \hfill
    \begin{subfigure}[]{0.32\textwidth}
        \centering
        \includegraphics[width=\textwidth]{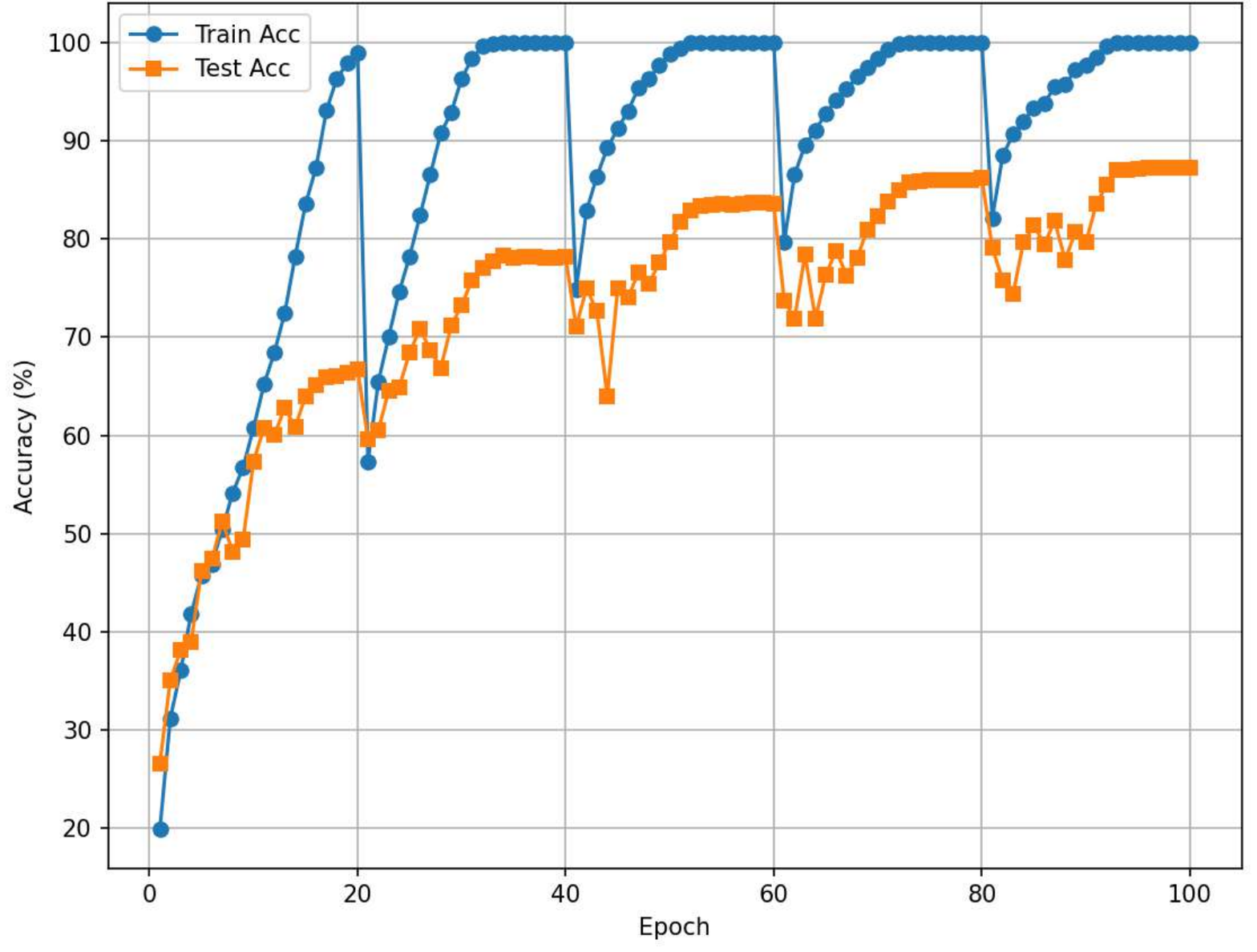}
        \caption{Training and test accuracy.}
        \label{fig:restnet_pacing_stage_acc}
    \end{subfigure}
    \hfill
    \begin{subfigure}[]{0.32\textwidth}
        \centering
        \includegraphics[width=\textwidth]{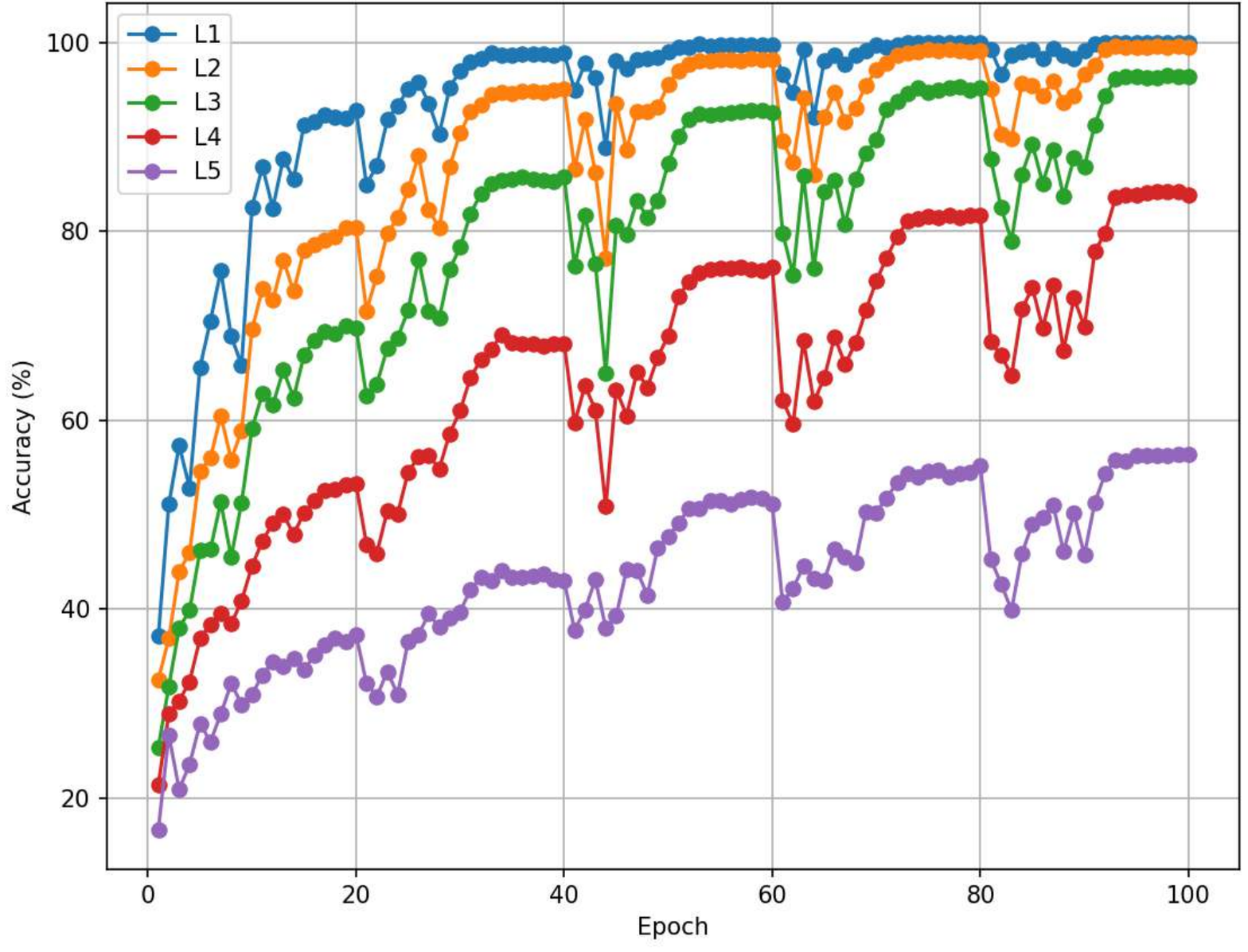}
        \caption{Stage-wise test accuracy.}
        \label{fig:restnet_pacing_stage_stages}
    \end{subfigure}
    
   \caption{Training curves for \textbf{ResNet-18 Baseline + Pacing, Stage-Cosine LR}.}
\label{fig:appendix_restnet18_Pacing_stagecosine}
\end{figure*}

\begin{figure*}[h]
    \centering
    \begin{subfigure}[]{0.32\textwidth}
        \centering
        \includegraphics[width=\textwidth]{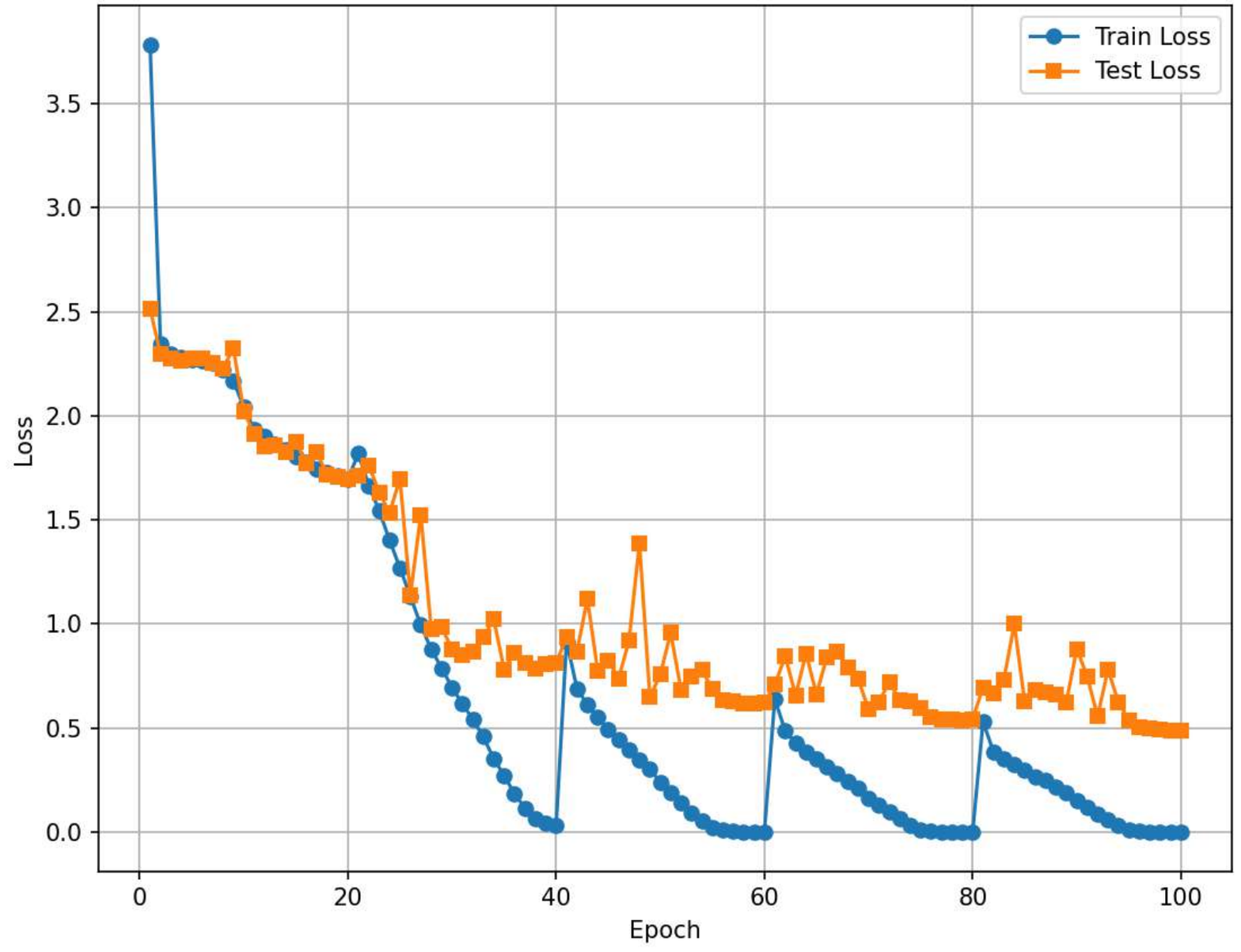}
        \caption{Training and test loss.}
        \label{fig:vgg16_pacing_stage_loss}
    \end{subfigure}
    \hfill
    \begin{subfigure}[]{0.32\textwidth}
        \centering
        \includegraphics[width=\textwidth]{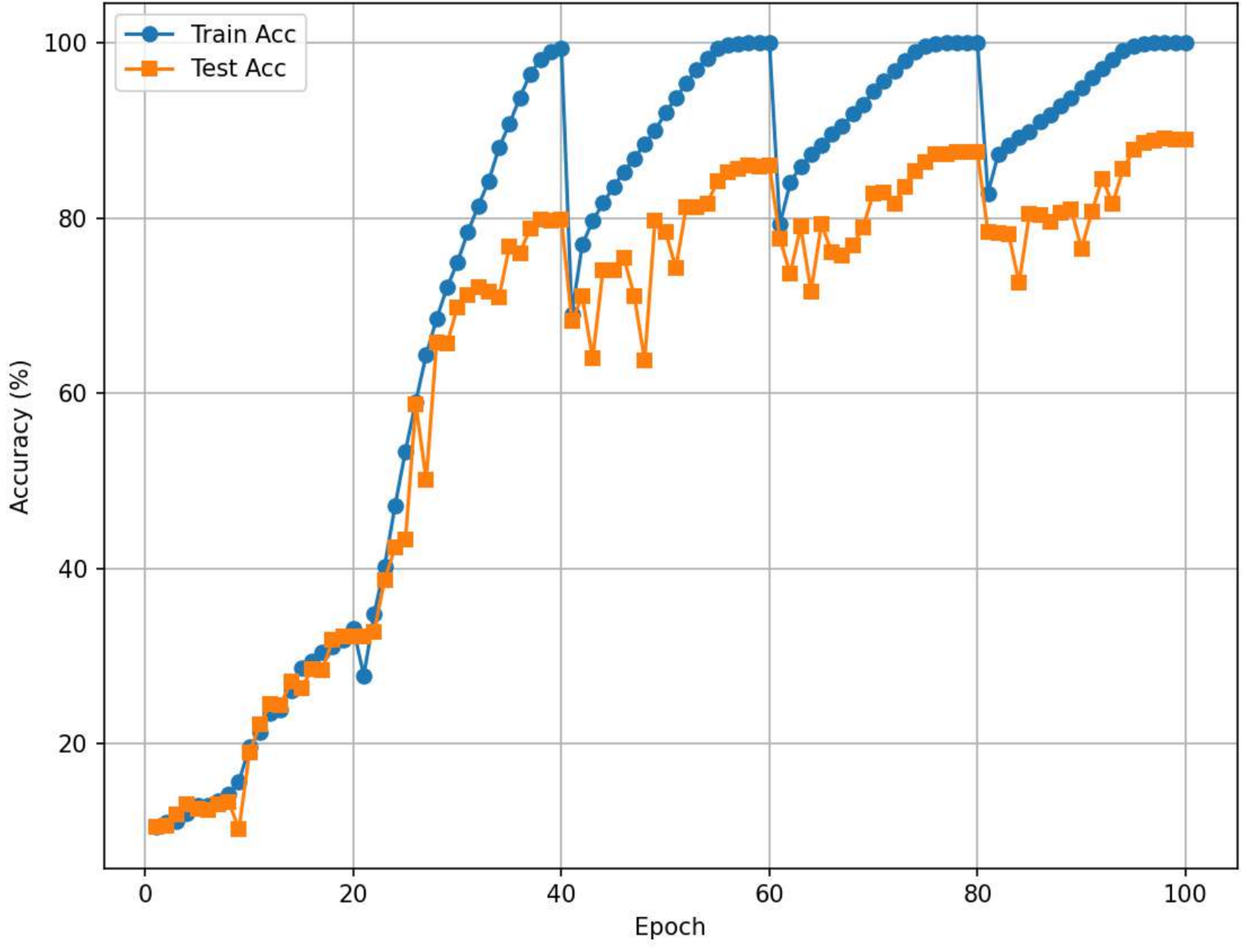}
        \caption{Training and test accuracy.}
        \label{fig:vgg16_pacing_stage_acc}
    \end{subfigure}
    \hfill
    \begin{subfigure}[]{0.32\textwidth}
        \centering
        \includegraphics[width=\textwidth]{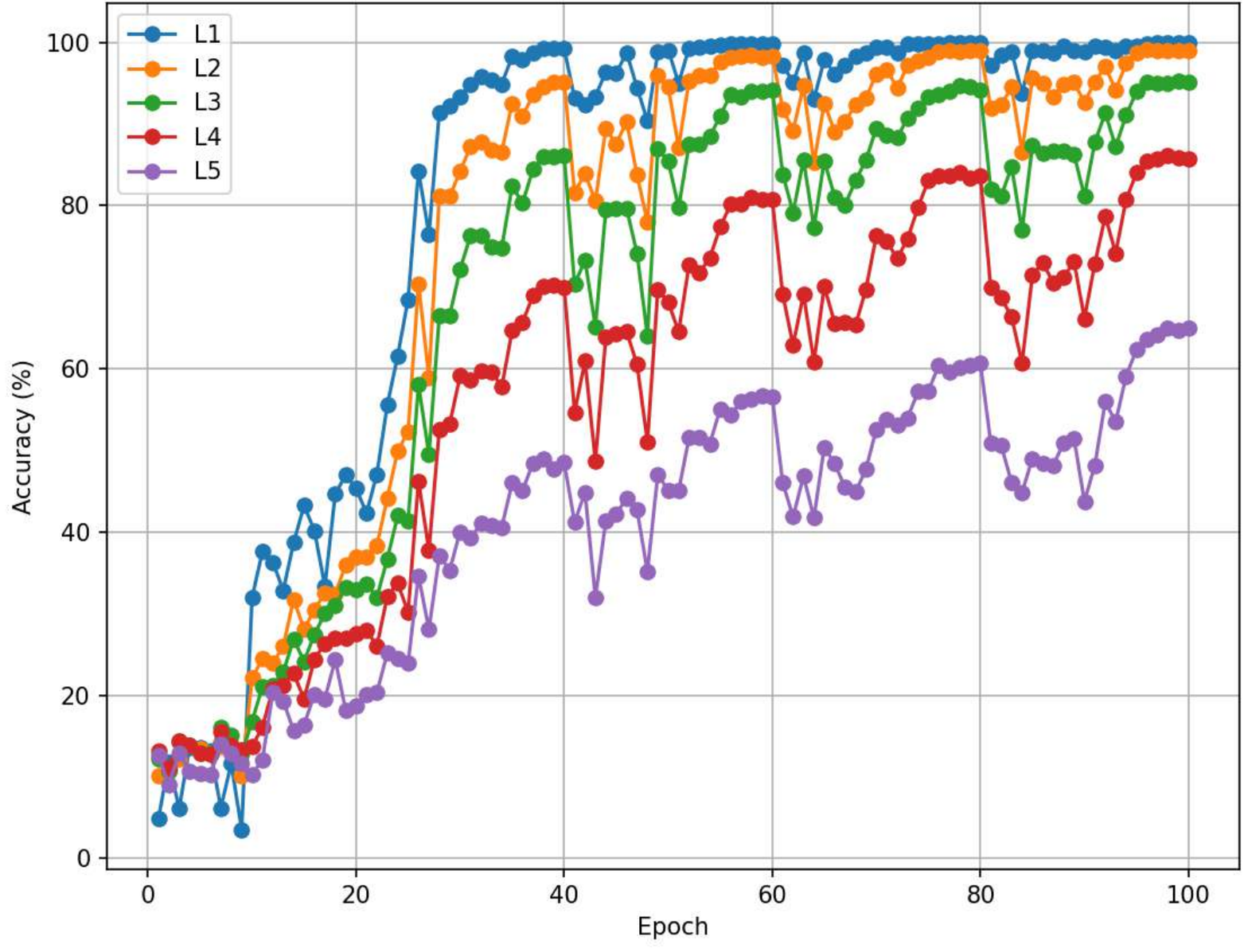}
        \caption{Stage-wise test accuracy.}
        \label{fig:vgg16_pacing_stage_stages}
    \end{subfigure}
    
    \caption{Training curves for \textbf{VGG16 pacing Stage Cosine} (Cosine LR, 100 epochs). 
    (a) Loss converges smoothly; test loss shows high variance in mid-training 
    before stabilising after epoch 60. 
    (b) Train accuracy reaches 100\% while test accuracy plateaus at 89.20\%. 
    (c) Stage accuracies confirm a clear difficulty gradient: L1 (easiest) 
    approaches 100\% early while L5 (hardest) plateaus near 62\%, 
    reflecting the inherent difficulty spread in CIFAR-10.}
    \label{fig:appendix_vgg16_Pacing_stagecosine}
\end{figure*}

\FloatBarrier
% \onecolumn
% \clearpage
\section{Data-Efficiency Results}

\renewcommand{\thefigure}{C\arabic{figure}}
\setcounter{figure}{0}
\renewcommand{\thetable}{C\arabic{table}}
\setcounter{table}{0}

\label{app:AppendixC}

\begin{figure*}[h]
    \centering
    \includegraphics[width=\textwidth]{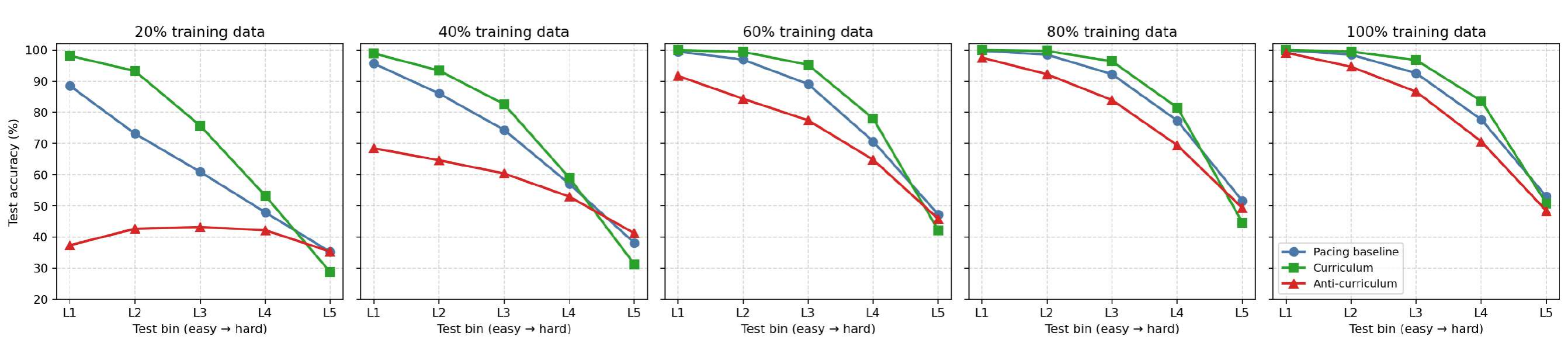}
    \caption{Data efficiency across difficulty stages $L_1$--$L_5$.}
    \label{fig:data_eff_2}
\end{figure*}

To evaluate training-stage data efficiency, we measure performance at progressive curriculum stages corresponding to increasing cumulative exposure to the ranked training set. Under the shared pacing schedule, the $20\%$, $40\%$, $60\%$, and $80\%$ stages represent matched points during training rather than independent low-data settings.

\begin{figure*}[h]
    \centering
    \begin{subfigure}[]{0.48\textwidth}
        \centering
        \includegraphics[width=\textwidth]{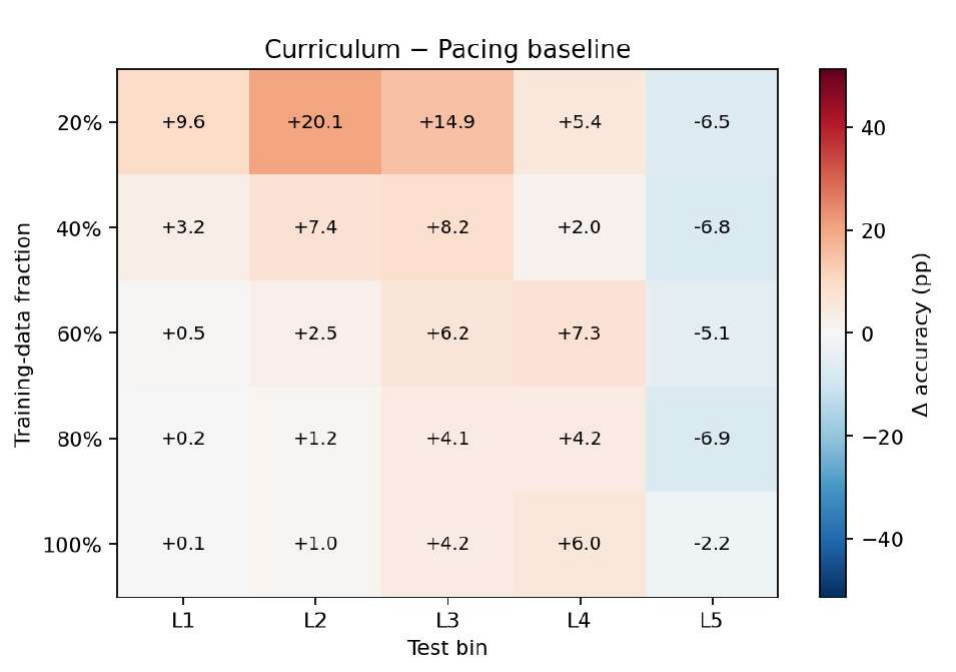}
        \caption{Data efficiency (extended view 1).}
        \label{fig:data_eff_3}
    \end{subfigure}
    \hfill
    \begin{subfigure}[]{0.48\textwidth}
        \centering
        \includegraphics[width=\textwidth]{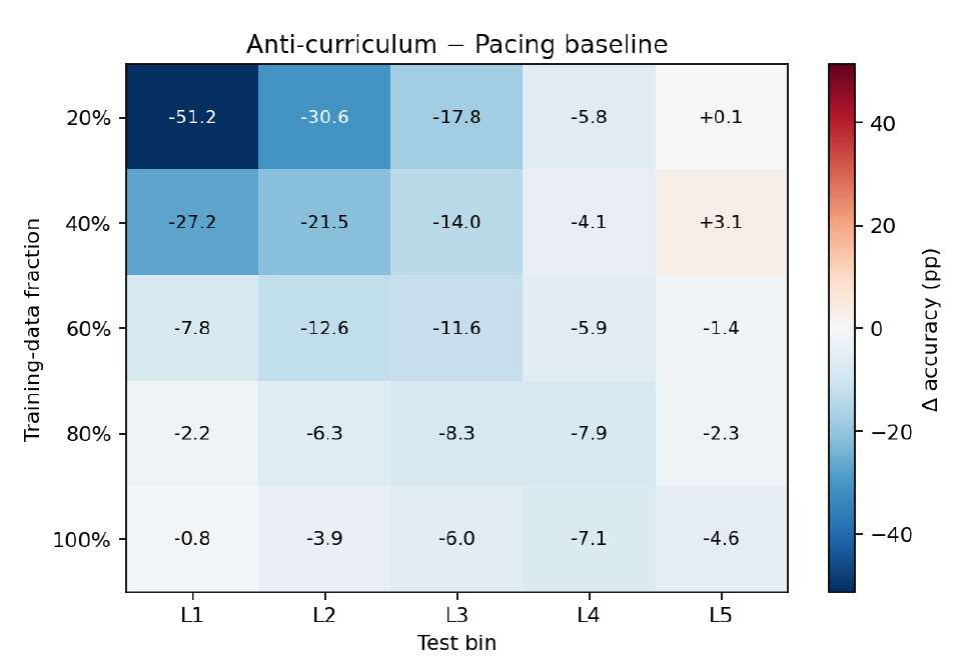}
        \caption{Data efficiency (extended view 2).}
        \label{fig:data_eff_4}
    \end{subfigure}
    \caption{Extended data efficiency analysis.}
    \label{fig:data_eff_appendix}
\end{figure*}

Table~\ref{tab:stgnn-comparison} and Figures~\ref{fig:data_eff_1}, \ref{fig:data_eff_appendix} summarize the results across the difficulty-aware test subsets $L_1$--$L_5$. Figure~\ref{fig:data_eff_1} visualizes how curriculum ordering affects performance as progressively harder samples are introduced during optimization.

Curriculum learning consistently outperforms the pacing-only baseline at matched training stages, particularly during early optimization. After the first stage ($20\%$ cumulative exposure), curriculum training achieves $69.81\%$ aggregate accuracy compared to $61.13\%$ for the pacing baseline, indicating that the scoring function provides meaningful structure beyond staged data exposure alone.

In contrast, anti-curriculum training performs substantially worse during early training stages. Introducing highly ambiguous samples first destabilizes optimization and leads to poor intermediate representations, particularly at the $20\%$ stage where aggregate accuracy drops to $40.09\%$. This suggests that early optimization dynamics are sensitive to sample ordering.

As training progresses toward full dataset exposure, the performance gap between curriculum and baseline narrows, consistent with prior observations that curriculum learning primarily improves optimization efficiency rather than the final converged solution.

Overall, these results suggest that the proposed framework improves data efficiency during staged training by enabling stronger intermediate performance earlier in optimization while maintaining a meaningful difficulty gradient across evaluation subsets.

\begin{table*}[h]

% \caption{Comparison of training strategies across test subsets at different training data percentages. Best results in \textbf{bold}, second best \underline{underlined}.}
\caption{Data efficiency of curriculum-based training strategies. We report test accuracy (\%) at progressive training subset sizes, where each $k\%$ subset contains the top-$k\%$ samples ranked by the scoring function (e.g., the $40\%$ subset extends the $20\%$ subset with the next-best $20\%$). Test subsets $L_{1}$--$L_{5}$ partition the test set by scoring-function value, from easiest ($L_{1}$) to hardest ($L_{5}$).}
\label{tab:stgnn-comparison}
\centering
\begin{tabular}{llccccccc}
\toprule
\multicolumn{1}{c}{\multirow{2}{*}{\makecell{Training Data\\ Subset}}} & \multirow{2}{*}{\makecell{Training \\ Strategy}} & \multicolumn{5}{c}{Test Subset} & \multicolumn{1}{c}{\multirow{2}{*}{\makecell{Aggregate \\ Test Accuracy}}} \\
\cmidrule(lr){3-7}
 & & $L_{1}$ & $L_{2}$ & $L_{3}$ & $L_{4}$ & $L_{5}$ & \\
\midrule
\multirow{4}{*}{20\%} 
 % & Baseline         & 99.10 & 94.70 & 87.45 & 73.25 & 48.60 & 80.62 \\
 & Pacing baseline  & 88.55 & 73.15 & 60.90 & 47.90 & 35.15 & 61.13 \\
 & Curriculum       & 98.15 & 93.25 & 75.75 & 53.25 & 28.65 & 69.81 \\
 & Anti-curriculum  & 37.30 & 42.60 & 43.15 & 42.15 & 35.25 & 40.09 \\
\midrule
\multirow{4}{*}{40\%} 
 % & Baseline         & 98.80 & 94.50 & 85.90 & 68.60 & 46.00 & 78.76 \\
 & Pacing baseline  & 95.65 & 86.05 & 74.35 & 57.05 & 38.05 & 70.23 \\
 & Curriculum       & 98.90 & 93.40 & 82.60 & 59.00 & 31.20 & 73.02 \\
 & Anti-curriculum  & 68.45 & 64.60 & 60.35 & 52.95 & 41.15 & 57.50 \\
\midrule
\multirow{4}{*}{60\%} 
 % & Baseline         & 99.75 & 97.45 & 92.90 & 81.95 & 56.15 & 85.64 \\
 & Pacing baseline  & 99.50 & 96.90 & 89.05 & 70.65 & 47.15 & 80.65 \\
 & Curriculum       & 99.95 & 99.40 & 95.25 & 77.95 & 42.05 & 82.92 \\
 & Anti-curriculum  & 91.70 & 84.35 & 77.40 & 64.75 & 45.75 & 72.79 \\
\midrule
\multirow{4}{*}{80\%} 
 % & Baseline         & 100.00 & 99.40 & 96.75 & 87.00 & 61.65 & 88.96 \\
 & Pacing baseline  & 99.75 & 98.55 & 92.20 & 77.30 & 51.55 & 83.87 \\
 & Curriculum       & 100.00 & 99.70 & 96.35 & 81.50 & 44.60 & 84.43 \\
 & Anti-curriculum  & 97.55 & 92.20 & 83.90 & 69.40 & 49.25 & 78.46 \\
\midrule
\multirow{4}{*}{100\%} 
 % & Baseline         & 100.00 & 99.25 & 97.10 & 87.20 & 61.55 & 89.02 \\
 & Pacing baseline  & 99.90 & 98.50 & 92.55 & 77.65 & 52.90 & 84.30 \\
 & Curriculum       & 100.00 & 99.50 & 96.75 & 83.65 & 50.65 & 86.11 \\
 & Anti-curriculum  & 99.15 & 94.60 & 86.60 & 70.60 & 48.30 & 79.85 \\
\bottomrule
\end{tabular}
\end{table*}

\FloatBarrier

\section{Sensitivity to Difficulty Ordering}
\renewcommand{\thefigure}{D\arabic{figure}}
\setcounter{figure}{0}
\renewcommand{\thetable}{D\arabic{table}}
\setcounter{table}{0}
\label{app:AppendixD}
To further evaluate the importance of ordering, we conduct a permutation study over difficulty bins $B_1$–$B_5$. These bins are constructed by partitioning the training set according to the proposed scoring function, from easiest ($B_1$) to hardest ($B_5$). This notation is distinct from the test subsets $L_1$–$L_5$, which are used for evaluation in Section~\ref{transfer_teacher_evluation}.

While the scoring function defines a ranking over samples, training dynamics depend on the order in which these bins are introduced during optimization.

We fix all hyperparameters across experiments (SGD optimizer with learning rate 0.1, momentum 0.9, weight decay $5 \times 10^{-4}$, and Stage-Cosine learning rate schedule over 100 epochs), and vary only the ordering of difficulty bins. This isolates the effect of ordering directionality independent of scoring and pacing design.

% \paragraph{Analysis.}
The results indicate that ordering has a measurable but architecture-dependent impact on performance. For VGG-16, several permutations that introduce lower-difficulty bins early tend to perform competitively, with the best result achieved by a partially aligned ordering rather than the strictly monotonic curriculum. This suggests that while a coarse easy-to-hard progression is beneficial, limited interleaving of difficulty levels may further improve generalization.

In contrast, ResNet-18 exhibits relatively small variation across the tested permutations, with no clear performance gain from curriculum-aligned ordering. The differences between orderings remain within a narrow range, indicating reduced sensitivity to bin sequencing.

Additionally, fully reversed ordering (hard-to-easy) leads to a noticeable drop in performance for both architectures, supporting the intuition that introducing highly ambiguous samples early can hinder optimization.

Overall, these findings suggest that the effectiveness of curriculum ordering depends not only on the quality of the difficulty signal, but also on the inductive biases of the underlying architecture.

\begin{figure*}[!t]
    \centering
    \includegraphics[width=0.7\textwidth]{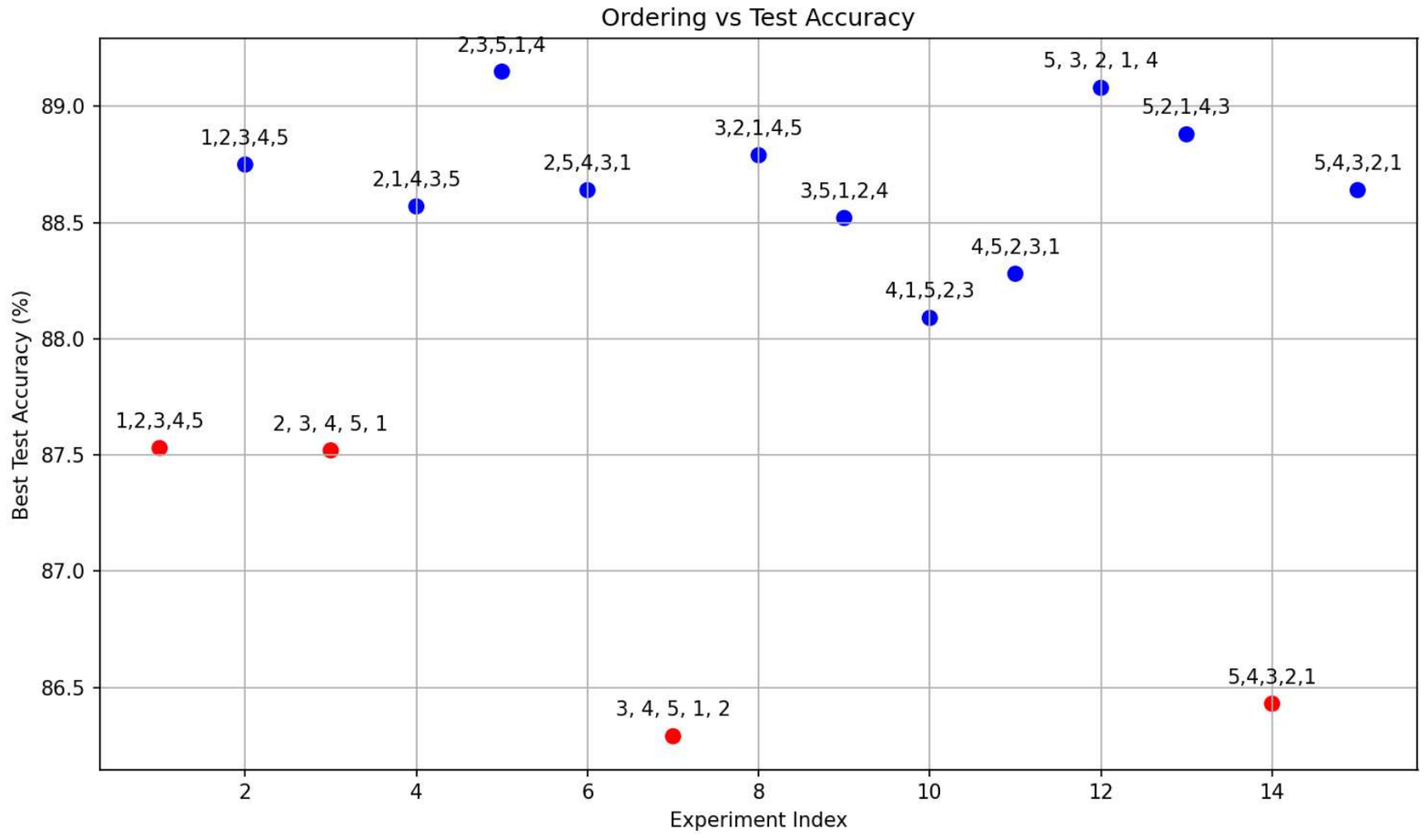}
    \caption{Test accuracy across all tested bin orderings for VGG-16 (blue) 
    and ResNet-18 (red). Each point corresponds to one permutation from 
    Table~\ref{tab:permutation_results}. VGG-16 shows higher sensitivity 
    to ordering with a wider spread, while ResNet-18 remains relatively 
    stable across permutations.}
    \label{fig:ordering_vs_accuracy}
\end{figure*}

\begin{figure*}[!t]
    \centering
    \begin{subfigure}[b]{0.32\textwidth}
        \centering
        \includegraphics[width=\textwidth]{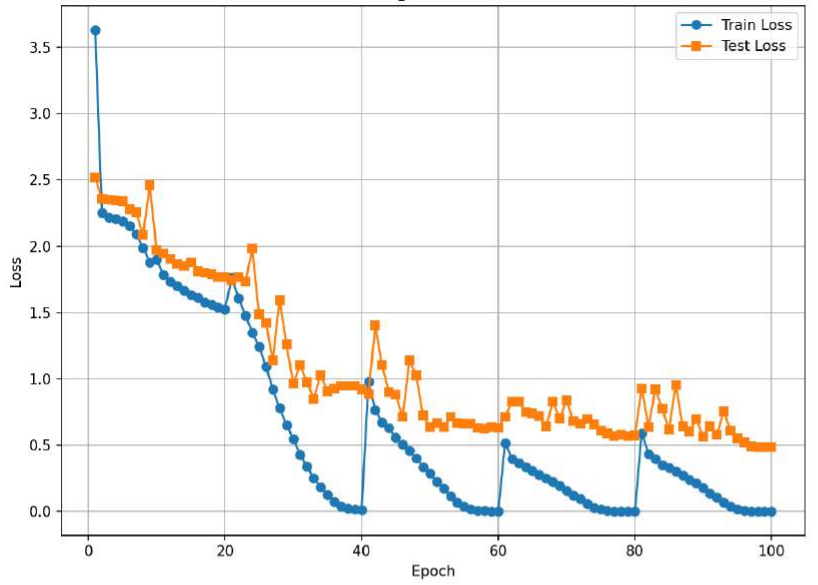}
        \caption{Training and test loss.}
        \label{fig:perm_best_loss}
    \end{subfigure}
    \hfill
    \begin{subfigure}[b]{0.32\textwidth}
        \centering
        \includegraphics[width=\textwidth]{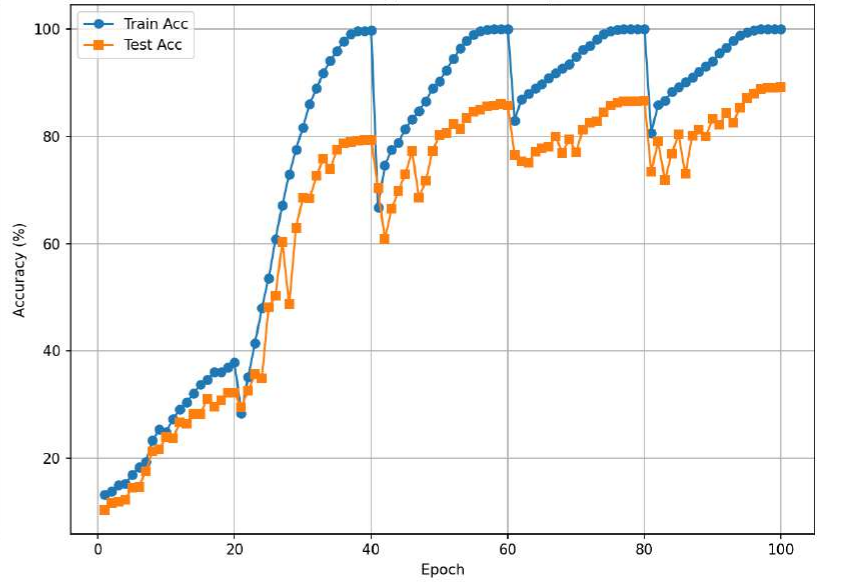}
        \caption{Training and test accuracy.}
        \label{fig:perm_best_acc}
    \end{subfigure}
    \hfill
    \begin{subfigure}[b]{0.32\textwidth}
        \centering
        \includegraphics[width=\textwidth]{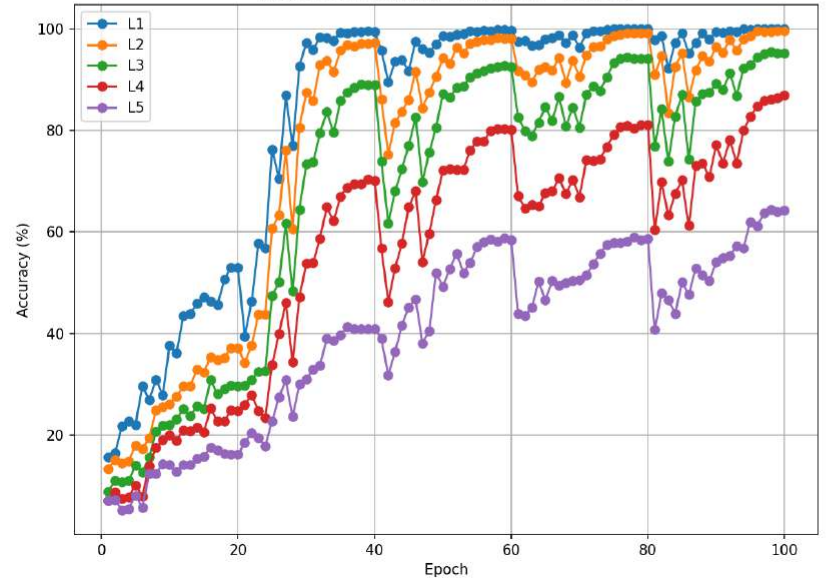}
        \caption{Stage-wise test accuracy.}
        \label{fig:perm_best_stages}
    \end{subfigure}
    \caption{Training curves for the best-performing permutation 
    $B_2 \rightarrow B_3 \rightarrow B_5 \rightarrow B_1 \rightarrow B_4$ 
    (VGG-16, 89.07\%). The stage-wise plot confirms a clear difficulty 
    gradient is maintained despite the non-monotonic bin ordering, 
    suggesting the model adapts effectively when early stages establish 
    sufficient representational structure.}
    \label{fig:perm_best_curves}
\end{figure*}

Table~\ref{tab:permutation_results} reports the final test accuracy for each permutation.

\begin{table*}[!t]
\caption{Effect of difficulty-bin ordering on test accuracy across architectures. Each ordering is a permutation of bins $B_1$–$B_5$ (easiest to hardest).}
\label{tab:permutation_results}
\centering
\begin{tabular}{llcc}
\toprule
\textbf{Architecture} & \textbf{Protocol} & \textbf{Ordering} & \textbf{Test Accuracy (\%)} \\
\midrule
\multirow{12}{*}{VGG-16}
  & Curriculum      & $B_1 \rightarrow B_2 \rightarrow B_3 \rightarrow B_4 \rightarrow B_5$ & 88.85 \\
\cmidrule(lr){2-4}
  & \multirow{10}{*}{Mixed}
                    & $B_2 \rightarrow B_1 \rightarrow B_4 \rightarrow B_3 \rightarrow B_5$ & 88.57 \\
  &                 & $B_3 \rightarrow B_2 \rightarrow B_1 \rightarrow B_4 \rightarrow B_5$ & 88.79 \\
  &                 & $B_2 \rightarrow B_3 \rightarrow B_5 \rightarrow B_1 \rightarrow B_4$ & 89.07 \\
  &                 & $B_2 \rightarrow B_5 \rightarrow B_4 \rightarrow B_3 \rightarrow B_1$ & 88.64 \\
  &                 & $B_3 \rightarrow B_5 \rightarrow B_1 \rightarrow B_2 \rightarrow B_4$ & 88.46 \\
  &                 & $B_4 \rightarrow B_1 \rightarrow B_5 \rightarrow B_2 \rightarrow B_3$ & 88.07 \\
  &                 & $B_4 \rightarrow B_5 \rightarrow B_2 \rightarrow B_3 \rightarrow B_1$ & 88.28 \\
  &                 & $B_3 \rightarrow B_4 \rightarrow B_5 \rightarrow B_1 \rightarrow B_2$ & 86.29 \\
  &                 & $B_5 \rightarrow B_2 \rightarrow B_1 \rightarrow B_4 \rightarrow B_3$ & 88.88 \\
  &                 & $B_5 \rightarrow B_3 \rightarrow B_2 \rightarrow B_1 \rightarrow B_4$ & \textbf{89.08} \\
\cmidrule(lr){2-4}
  & Anti-curriculum & $B_5 \rightarrow B_4 \rightarrow B_3 \rightarrow B_2 \rightarrow B_1$ & 88.64 \\
\midrule
\multirow{4}{*}{ResNet-18}
  & Curriculum      & $B_1 \rightarrow B_2 \rightarrow B_3 \rightarrow B_4 \rightarrow B_5$ & 87.53 \\
\cmidrule(lr){2-4}
  & \multirow{2}{*}{Mixed}
                    & $B_2 \rightarrow B_3 \rightarrow B_4 \rightarrow B_5 \rightarrow B_1$ & 87.52 \\
  &                 & $B_3 \rightarrow B_4 \rightarrow B_5 \rightarrow B_1 \rightarrow B_2$ & 86.29 \\
\cmidrule(lr){2-4}
  & Anti-curriculum & $B_5 \rightarrow B_4 \rightarrow B_3 \rightarrow B_2 \rightarrow B_1$ & 86.43 \\
\bottomrule
\end{tabular}
\end{table*}
%%%%%%%%%%%%%%%%%%%%%%%%%%%%%%%%%%%%%%%%%%%%%%%%%%%%%%%%%%%%%%%%%%%%%%%%%%%%%%%
%%%%%%%%%%%%%%%%%%%%%%%%%%%%%%%%%%%%%%%%%%%%%%%%%%%%%%%%%%%%%%%%%%%%%%%%%%%%%%%

\end{document}